\documentclass{article}

\usepackage{jmlr2e}
\usepackage[table,xcdraw]{xcolor}

\usepackage[utf8]{inputenc} 
\usepackage[T1]{fontenc}    
\usepackage{url}            

\usepackage{booktabs}       
\usepackage{amsfonts}       
\usepackage{nicefrac}       
\usepackage{microtype}      
\usepackage{lscape}
\usepackage[mathscr]{euscript}
\usepackage{latexsym}
\usepackage{epsfig}

\usepackage{comment}
\usepackage{subcaption}
\usepackage[framemethod=TikZ]{mdframed}
\mdfdefinestyle{MyFrame}{%
    linecolor=black,
    outerlinewidth=2pt,
    roundcorner=20pt,
    innertopmargin=\baselineskip,
    innerbottommargin=\baselineskip,
    innerrightmargin=20pt,
    innerleftmargin=20pt,
    backgroundcolor=gray!10!white}

\usepackage[mathscr]{euscript}
\usepackage{latexsym}
\usepackage{amsmath}
\usepackage{amssymb}
\usepackage{bm}
\usepackage{enumitem}
\usepackage{mathtools}
\DeclareSymbolFont{rsfs}{U}{rsfs}{m}{n}
\DeclareSymbolFontAlphabet{\mathscrsfs}{rsfs}
\DeclareMathOperator*{\argmin}{arg\,min}

\renewcommand{\P}{{\sf P}}

\usepackage{multirow}
    
\usepackage{hyperref}       
\usepackage{cleveref}

\newcommand{\trainP}{\P}
\newcommand{\testP}{\P'}

\title{\vspace{-.5em}Underspecification Presents Challenges for Credibility in Modern Machine Learning}
\ShortHeadings{Underspecification in Machine Learning}{A. D'Amour, K. Heller, D. Moldovan, et al}

\author{%
  \name Alexander D'Amour\thanks{These authors contributed equally to this work.} \email alexdamour@google.com\\
  \name Katherine Heller\footnotemark[\value{footnote}] \email kheller@google.com\\
  \name Dan Moldovan\footnotemark[\value{footnote}]  \email mdan@google.com\\
  \name Ben Adlam \email adlam@google.com\\
  \name Babak Alipanahi \email babaka@google.com\\
  \name Alex Beutel \email alexbeutel@google.com\\
  \name Christina Chen \email christinium@google.com\\
  \name Jonathan Deaton \email jdeaton@google.com\\
  \name Jacob Eisenstein \email jeisenstein@google.com\\
  \name Matthew D. Hoffman \email mhoffman@google.com\\
  \name Farhad Hormozdiari \email fhormoz@google.com\\
  \name Neil Houlsby \email neilhoulsby@google.com\\
  \name Shaobo Hou \email shaobohou@google.com\\
  \name Ghassen Jerfel \email ghassen@google.com\\
  \name Alan Karthikesalingam \email alankarthi@google.com\\
  \name Mario Lucic \email lucic@google.com\\
  \name Yian Ma \email yianma@ucsd.edu\\
  \name Cory McLean \email cym@google.com\\
  \name Diana Mincu \email dmincu@google.com\\
  \name Akinori Mitani \email amitani@google.com\\
  \name Andrea Montanari \email montanari@stanford.edu\\
  \name Zachary Nado \email znado@google.com\\
  \name Vivek Natarajan \email natviv@google.com\\
  \name Christopher Nielson\thanks{This paper represents the views of the authors, and not of the VA.} \email christopher.nielson@va.gov\\
  \name Thomas F. Osborne\footnotemark[\value{footnote}] \email thomas.osborne@va.gov\\
  \name Rajiv Raman \email drrrn@snmail.org\\
  \name Kim Ramasamy \email kim@aravind.org\\
  \name Rory Sayres \email sayres@google.com\\
  \name Jessica Schrouff \email schrouff@google.com\\
  \name Martin Seneviratne \email martsen@google.com\\
  \name Shannon Sequeira \email shnnn@google.com\\
  \name Harini Suresh \email hsuresh@mit.edu\\
  \name Victor Veitch \email victorveitch@google.com\\
  \name Max Vladymyrov \email mxv@google.com\\
  \name Xuezhi Wang \email xuezhiw@google.com\\
  \name Kellie Webster \email websterk@google.com\\
  \name Steve Yadlowsky \email yadlowsky@google.com\\
  \name Taedong Yun \email tedyun@google.com\\
  \name Xiaohua Zhai \email xzhai@google.com\\
  \name D. Sculley \email dsculley@google.com}  
\editor{}

\begin{document}

\maketitle

\begin{abstract}
ML models often exhibit unexpectedly poor behavior when they are deployed in real-world domains. We identify underspecification as a key reason for these failures. An ML pipeline is underspecified when it can return many predictors with equivalently strong held-out performance in the training domain. Underspecification is common in modern ML pipelines, such as those based on deep learning. Predictors returned by underspecified pipelines are often treated as equivalent based on their training domain performance, but we show here that such predictors can behave very differently in deployment domains. This ambiguity can lead to instability and poor model behavior in practice, and is a distinct failure mode from previously identified issues arising from structural mismatch between training and deployment domains. We show that this problem appears in a wide variety of practical ML pipelines, using examples from computer vision, medical imaging, natural language processing, clinical risk prediction based on electronic health records, and medical genomics. Our results show the need to explicitly account for underspecification in modeling pipelines that are intended for real-world deployment in any domain.
\end{abstract}

\section{Introduction}
In many applications of machine learning (ML), a trained model is required to not only predict well in the training domain, but also encode some essential structure of the underlying system.
In some domains, such as medical diagnostics, the required structure corresponds to causal phenomena that remain invariant under intervention.
In other domains, such as natural language processing, the required structure is determined by the details of the application (e.g., the requirements in question answering, where world knowledge is important, may be different from those in translation, where isolating semantic knowledge is desirable).
These requirements for encoded structure 
have practical consequences:
they determine whether the model will generalize as expected in deployment scenarios.
These requirements often determine whether a predictor is credible, that is, whether it can be trusted in practice.

Unfortunately, standard ML pipelines are poorly set up for satisfying these requirements. 
Standard ML pipelines are built around a training task that is characterized by a model specification, a training dataset, and an independent and identically distributed (iid) evaluation procedure; that is, a procedure that validates a predictor's expected predictive performance on data drawn from the training distribution.
Importantly, the evaluations in this pipeline are agnostic to the particular inductive biases encoded by the trained model.
While this paradigm has enabled transformational progress in a number of problem areas, its blind spots are now becoming more salient.
In particular, concerns regarding ``spurious correlations'' and ``shortcut learning'' in trained models are now widespread \citep[e.g.,][]{geirhos2020shortcut,arjovsky2019invariant}.

The purpose of this paper is to explore this gap, and how it can arise in practical ML pipelines.
A common explanation is simply that, in many situations, there is a fundamental conflict between iid performance and desirable behavior in deployment.
For example, this occurs when there are differences in causal structure between training and deployment domains, or when the data collection mechanism imposes a selection bias.
In such cases, the iid-optimal predictors must necessarily incorporate spurious associations \citep{Caruana2015,arjovsky2019invariant,ilyas2019adversarial}.
This is intuitive:
a predictor trained in a setting that is structurally misaligned with the application will reflect this mismatch.

However, this is not the whole story. 
Informally, in this structural-conflict view,
we would expect that two identically trained predictors would show the same defects in deployment.
The observation of this paper is that this structural-conflict view does not adequately capture the challenges of deploying ML models in practice.
Instead, predictors trained to the same level of iid generalization will often show widely divergent behavior when applied to real-world settings.

We identify the root cause of this behavior as underspecification in ML pipelines. 
In general, the solution to a problem is underspecified if there are many distinct solutions that solve the problem equivalently.
For example, the solution to an underdetermined system of linear equations (i.e., more unknowns than linearly independent equations) is underspecified, with an equivalence class of solutions given by a linear subspace of the variables.
In the context of ML, we say an ML pipeline is underspecified if there are many distinct ways (e.g., different weight configurations) for the model to achieve equivalent held-out performance on iid data, even if the model specification and training data are held constant.
Underspecification is well-documented in the ML literature, and is a core idea in deep ensembles, double descent, Bayesian deep learning, and loss landscape analysis \citep{Lakshminarayanan2017deepensembles,fort2019deep,belkin2018reconciling,Nakkiran2020Deep}.
However, its implications for the gap between iid and application-specific generalization are neglected.

Here, we make two main claims about the role of underspecification in modern machine learning.
The first claim is that underspecification in ML pipelines is a key obstacle to reliably training models that behave as expected in deployment.
Specifically, when a training pipeline must choose between many predictors that yield near-optimal iid performance, if the pipeline is only sensitive to iid performance, it will return an arbitrarily chosen predictor from this class.
Thus, even if there exists an iid-optimal predictor that encodes the right structure, we cannot guarantee that such a model will be returned when the pipeline is underspecified.
We demonstrate this issue in several examples that incorporate simple models: one simulated, one theoretical, and one a real empirical example from medical genomics.
In these examples, we show how, in practice, underspecification manifests as sensitivity to arbitrary choices that keep iid performance fixed, but can have substantial effects on performance in a new domain, such as in model deployment.

The second claim is that underspecification is ubiquitous in modern applications of ML, and has substantial practical implications.
We support this claim with an empirical study, in which we apply a simple experimental protocol across production-grade deep learning pipelines in computer vision, medical imaging, natural language processing (NLP), and electronic health record (EHR) based prediction.
The protocol is designed to detect underspecification by showing that a predictor's performance on \emph{stress tests}---empirical evaluations that probe the model's inductive biases on practically relevant dimensions---is sensitive to arbitrary, iid-performance-preserving choices, such as the choice of random seed.
A key point is that the stress tests induce variation between predictors' behavior, not simply a uniform degradation of performance. 
This variation distinguishes underspecification-induced failure from the more familiar case of structural-change induced failure.
We find evidence of underspecification in all applications, with downstream effects on robustness, fairness, and causal grounding.

Together, our findings indicate that underspecification can, and does, degrade the credibility of ML predictors in applications, even in settings where the prediction problem is well-aligned with the goals of an application.
The direct implication of our findings is that substantive real-world behavior of ML predictors can be determined in unpredictable ways by choices that are made for convenience, such as initialization schemes or step size schedules chosen for trainability---even when these choices do not affect iid performance. 
More broadly, our results suggest a need to explicitly test models for required behaviors in all cases where these requirements are not directly guaranteed by iid evaluations.
Finally, these results suggest a need for training and evaluation techniques tailored to address underspecification, such as flexible methods to constrain ML pipelines toward the credible inductive biases for each specific application. 
Interestingly, our findings suggest that enforcing these credible inductive biases need not compromise iid performance.

\paragraph{Organization}
The paper is organized as follows.
We present some core concepts and review relevant literature in Section~\ref{sec:setup}.
We present a set of examples of underspecification in simple, analytically tractable models as a warm-up in Section~\ref{sec:warmup}.
We then present a set of four deep learning case studies in Sections~\ref{sec:vision}--\ref{sec:ehr}.
We close with a discussion in Section~\ref{sec:discussion}.

Overall, our strategy in this paper is to provide a broad range of examples of underspecification in a variety of modeling pipelines.
Readers may not find it necessary to peruse every example to appreciate our argument, but different readers may find different domains to be more familiar.
As such, the paper is organized such that readers can take away most of the argument from understanding one example from Section~\ref{sec:warmup} and one case study from Sections~\ref{sec:vision}--\ref{sec:ehr}.
However, we believe there is benefit to presenting all of these examples under the single banner of underspecification, so we include them all in the main text.

\section{Preliminaries and Related Work}
\label{sec:setup}

\subsection{Underspecification}
We consider a supervised learning setting, where the goal is to obtain a predictor $f: \mathcal X \mapsto \mathcal Y$ that maps inputs $x$ (e.g., images, text) to labels $y$.
We say a \emph{model} is specified by a function class $\mathcal F$ from which a predictor $f(x)$ will be chosen.
An \emph{ML pipeline} takes in training data $\mathcal D$ drawn from a training distribution
$\trainP$ and produces a trained model, or \emph{predictor}, $f(x)$ from $\mathcal F$.
Usually, the pipeline selects $f \in \mathcal F$ by approximately minimizing the predictive risk  on the training distribution $\mathcal R_{ \trainP{}}(f) := \mathbb E_{(X, Y) \sim \trainP{}} [\ell(f(X), Y)]$.
Regardless of the method used to obtain a predictor $f$, we assume that the pipeline validates that $f$ achieves low expected risk on the training distribution $\trainP{}$ by evaluating its predictions on an independent and identically distributed test set $D'$, e.g., a hold-out set selected completely at random.
This validation translates to a behavioral guarantee, or \emph{contract} \citep{jacovi2020formalizing}, about the model's aggregate performance on future data drawn from $\trainP{}$.

We say that an ML pipeline is \emph{underspecified} if there are many predictors $f$ that a pipeline could return with similar predictive risk.
We denote this set of risk-equivalent near-optimal predictors $\mathcal F^* \subset \mathcal F$.
However, underspecification creates difficulties when the predictors in $\mathcal F^*$ encode substantially different inductive biases that result in different generalization behavior on distributions that differ from $\trainP$.
When this is true, even when $\mathcal{F}^*$ contains a predictor with credible inductive biases, a pipeline may return a different predictor because it cannot distinguish between them.

The ML literature has studied various notions of underspecification before.
In the deep learning literature specifically, much of the discussion has focused on the shape of the loss landscape $\mathbb E_{(X, Y) \sim \trainP{}} [\ell(f(X), Y)]$, and of the geometry of non-unique risk minimizers, including discussions of wide or narrow optima \citep[see, e.g.][]{chaudhari2019entropy}, and connectivity between global modes in the context of model averaging~\citep{izmailov2018averaging,fort2019deep,wilson2020bayesian} and network pruning~\citep{frankle2020linear}.
Underspecification also plays a role in recent analyses of overparametrization in theoretical and real deep learning models \citep{belkin2018reconciling,mei2019generalization,Nakkiran2020Deep}.
Here, underspecification is a direct consequence of having more degrees of freedom than datapoints.
Our work here complements these efforts in two ways: first, our goal is to understand how underspecification relates to inductive biases that could enable generalization beyond the training distribution $\trainP{}$; and secondly, the primary object that we study is practical ML \emph{pipelines} rather than the loss landscape itself.
This latter distinction is important for our empirical investigation, where the pipelines that we analyze incorporate a number of standard tricks, such as early stopping, which are ubiquitous in ML as it is applied to real problems, but difficult to fully incorporate into theoretical analysis.
However, we note that these questions are clearly connected, and in Section~\ref{sec:warmup}, we motivate underspecification using a similar approach to this previous literature.

Our treatment of underspecification is more closely related to work on ``Rashomon sets'' \citep{fisher2019all,semenova2019study}, ``predictive multiplicity'' \citep{marx2019predictive}, and methods that seek our risk-equivalent predictors that are ``right for the right reasons'' \citep{ross2017right}.
These lines of work similarly note that a single learning problem specification can admit many near-optimal solutions, and that these solutions may have very different properties along axes such as interpretability or fairness.
Our work here is complementary: we provide concrete examples of how such equivalence classes manifest empirically in common machine learning practice.

\subsection{Shortcuts, Spurious Correlations, and Structural vs Underspecified Failure Modes}
Many explorations of the failures of ML pipelines that optimize for iid generalization focus on cases where there is an explicit tension between iid generalization and encoding credible inductive biases.
We call these \emph{structural failure modes}, because they are often diagnosed as a misalignment between the predictor learned by empirical risk minimization and the causal structure of the desired predictor \citep{scholkopf2019causality,arjovsky2019invariant}.
In these scenarios, a predictor with credible inductive biases cannot achieve optimal iid generalization in the training distribution, because there are so-called ``spurious'' features in that are strongly associated with the label in the training data, but are not associated with the label in some practically important settings.

Some well-known examples of this case have been reported in medical applications of ML, where the training inputs often include markers of a doctor's diagnostic judgment \citep{oakden2020hidden}.
For example, \citet{winkler2019association} report on a CNN model used to diagnose skin lesions, which exhibited strong reliance on surgical ink markings around skin lesions that doctors had deemed to be cancerous.
Because the judgment that went into the ink markings may have used information not available in the image itself, an iid-optimal predictor would need to incorporate this feature, but these markings would not be expected to be present in deployment, where the predictor would itself be part of the workflow for making a diagnostic judgment.
In this context, \citet{peters2016causal,heinze2018invariant,arjovsky2019invariant,Magliacane++_NeurIPS_18} propose approaches to overcome this structural bias, often by using data collected in multiple environments to identify causal invariances.

While structural failure modes are important when they arise, they do not cover all cases where predictors trained to minimize predictive risk encode poor inductive biases.
In many settings where ML excels, the structural issues identified above are not present. For example, it's known in many perception problems that sufficient information exists in the relevant features of the input alone to recover the label with high certainty.
Instead, we argue that it is often the case that there is simply not enough information in the training distribution to distinguish between these inductive biases and spurious relationships: making the connection to causal reasoning, this underspecified failure mode corresponds to a lack of positivity, not a structural defect in the learning problem.
\citet{geirhos2020shortcut} connects this idea to the notion of ``shortcut learning''.
They point out that there may be many predictors that generalize well in iid settings, but only some that align with the intended solution tot he prediction problem.
In addition, they also note (as we do) that some seemingly arbitrary aspects of ML pipelines, such as the optimization procedure, can make certain inductive biases easier for a pipeline to represent, and note the need for future investigation in this area.
We agree with these points, and we offer additional empirical support to this argument. Furthermore, we show that even pipelines that are identical up to their random seed can produce predictors that encode distinct shortcuts, emphasizing the relevance of underspecification. We also emphasize that these problems are far-reaching across ML applications.

\subsection{Stress Tests and Credibility}
\label{sec:stress tests}
Our core claims revolve around how underspecification creates ambiguity in the encoded structure of a predictor, which, in turn, affect the predictor's credibility. 
In particular, we are interested in behavior that is \emph{not} tested by iid evaluations, but has observable implications in practically important situations.
To this end, we follow the framework presented in \citet{jacovi2020formalizing}, and focus on inductive biases that can be expressed in terms of a \emph{contract}, or an explicit statement of expected predictor behavior, that can be falsified concretely by \emph{stress tests}, or evaluations that probe a predictor by observing its outputs on specifically designed inputs.

Importantly, stress tests probe a broader set of contracts than iid evaluations.
Stress tests are becoming a key part of standards of evidence in a number of  applied domains, including medicine \citep{collins2015transparent,liu2020reporting,rivera2020guidelines}, economics \citep{mullainathan2017machine,athey2017beyond}, public policy \citep{kleinberg2015prediction}, and epidemiology \citep{hoffmann2019guidelines}.
In many settings where stress tests have been proposed in the ML literature, they have often uncovered cases where models fail to generalize as required for direct real-world application.
Our aim is to show that underspecification can play a role in these failures.

Here, we review three types of stress tests that we consider in this paper, and make connections to existing literature where they have been applied.

\paragraph{Stratified Performance Evaluations}
Stratified evaluations (i.e., subgroup analyses) test whether a predictor $f$ encodes inductive biases that yield similar performance across different strata of a dataset.
We choose a particular feature $A$ and stratify a standard test dataset $\mathcal D'$ into strata $\mathcal D'_a = \{(x_i, y_i) : A_i = a\}$.
A performance metric can then be calculated and compared across different values of $a$.

Stratified evaluations have been presented in the literature on fairness in machine learning, where examples are stratified by socially salient characteristics like skin type \citep{buolamwini2018gender}; the ML for healthcare literature \citep{Obermeyer2019,oakden2020hidden}, where examples are stratified by subpopulations; and the natural language processing and computer vision literatures where examples are stratified by topic or notions of difficulty \citep{hendrycks2019natural,zellers2018swag}.

\paragraph{Shifted Performance Evaluations}
Shifted performance evaluations test whether the average performance of a predictor $f$ generalizes when the test distribution differs in a specific way from the training distribution.
Specifically, these tests define a new data distribution $\testP \neq \trainP$ from which to draw the test dataset $\mathcal D'$, then evaluate a performance metric with respect to this shifted dataset.

There are several strategies for generating $\testP{}$, which test different properties of $f$.
For example, to test whether $f$ exhibits \emph{invariance} to a particular transformation $T(x)$ of the input, one can define $\testP{}$ to be the distribution of the variables $(T(x), y)$ when $(x, y)$ are drawn from the training distribution $P_{\mathcal D}$ (e.g., noising of images in ImageNet-C \citep{hendrycks2018benchmarking}).
One can also define $\mathcal P_{\mathcal D'}$ less formally, for example by changing the data scraping protocol used to collect the test dataset (e.g., ObjectNet \citep{barbu2019objectnet}), or changing the instrument used to collect data.

Shifted performance evaluations form the backbone of empirical evaluations in the literature on robust machine learning and task adaptation \citep[e.g.,][]{hendrycks2018benchmarking,wang2019learning,djolonga2020robustness,taori2020measuring}.
Shifted evaluations are also required in some reporting standards, including those for medical applications of AI~\citep{collins2015transparent,liu2020reporting,rivera2020guidelines}.

\paragraph{Contrastive Evaluations}
Shifted evaluations that measure aggregate performance can be useful for diagnosing the existence of poor inductive biases, but the aggregation involved can obscure more fine-grained patterns.
Contrastive evaluations can support localized analysis of particular inductive biases. 
Specifically, contrastive evaluations are performed on the example, rather than distribution level, and check whether a particular modification of the input $x$ causes the output of the model to change in unexpected ways.
Formally, a contrastive evaluation makes use of a dataset of matched sets $\mathcal C = \{z_i\}_{i=1}^{|\mathcal C|}$, where each matched set $z_i$ consists of a base input $x_i$ that is modified by a set of transformations $\mathcal T$, $z_i = (T_j(x_i))_{T_j \in \mathcal T}$.
In contrastive evaluations, metrics are computed with respect to matched sets, and can include, for example, measures of similarity or ordering among the examples in the matched set.
For instance, if it is assumed that each transformation in $\mathcal T$ should be label-preserving, then a measurement of disagreement within the matched sets can reveal a poor inductive bias.

Contrastive evaluations are common in the ML fairness literature, e.g., to assess counterfactual notions of fairness \citep{garg2019counterfactual,kusner2017counterfactual}.
They are also increasingly common as robustness or debugging checks in the natural language processing literature \citep{ribeiro2020beyond,Kaushik2020Learning}.

\section{Warm-Up: Underspecification in Simple Models}
\label{sec:warmup}
To build intuition for how underspecification manifests in practice, we demonstrate its consequences in three relatively simple models before moving on to study production-scale deep neural networks.
In particular, we examine three underspecified models in three different settings: (1) a simple parametric model for an epidemic in a simulated setting; (2) a shallow random feature model in the theoretical infinitely wide limit; and (3) a linear model in a real-world medical genomics setting, where such models are currently state-of-the-art.
In each case, we show that underspecification is an obstacle to learning a predictor with the required inductive bias.

\subsection{Underspecification in a Simple Epidemiological Model}

One core task in infectious disease epidemiology is forecasting the trajectory of an epidemic.
Dynamical models are often used for this task. 
Here, we consider a simple simulated setting where the data is generated exactly from this model; thus, unlike a real setting where model misspecification is a primary concern, the only challenge here is to recover the true parameters of the generating process, which would enable an accurate forecast.
We show that even in this simplified setting, underspecification can derail the forecasting task.

Specifically, we consider the simple Susceptible-Infected-Recovered (SIR) model that is often used as the basis of epidemic forecasting models in infectious disease epidemiology.
This model is specified in terms of the rates at which the number of susceptible ($S$), infected ($I$), and recovered ($R$) individuals in a population of size $N$, change over time:
\begin{align*}
  \frac{dS}{dt} = -\beta\left(\frac{I}{N}\right) S, \qquad
  \frac{dI}{dt} = - \frac{I}{D} + \beta\left(\frac{I}{N}\right) S,  \qquad
  \frac{dR}{dt} = \frac{I}{D}.
\end{align*}
In this model, the parameter $\beta$ represents the transmission rate of the disease from the infected to susceptible populations, and the parameter $D$ represents the average duration that an infected individual remains infectious.

To simulate the forecasting task, we generate a full trajectory from this model for a full time-course $T$, but learn the parameters $(\beta, D)$ from data of observed infections up to some time $T_{\text{obs}} < T$ by minimizing squared-error loss on predicted infections at each timepoint using gradient descent (susceptible and recovered are usually not observed).
Importantly, during the early stages of an epidemic, when $T_{\text{obs}}$ is small, the parameters of the model are underspecified by this training task. 
This is because, at this stage, the number of susceptible
is approximately constant at the total population size ($N$),
and the number of infections grows approximately exponentially
at rate
$\beta - 1/D$.
The data only determine this rate.
Thus, there are many pairs of parameter values $(\beta, D)$ that describe the exponentially growing timeseries of infections equivalently.

\begin{figure}[t!]
    \centering
    \includegraphics[width=0.3\textwidth]{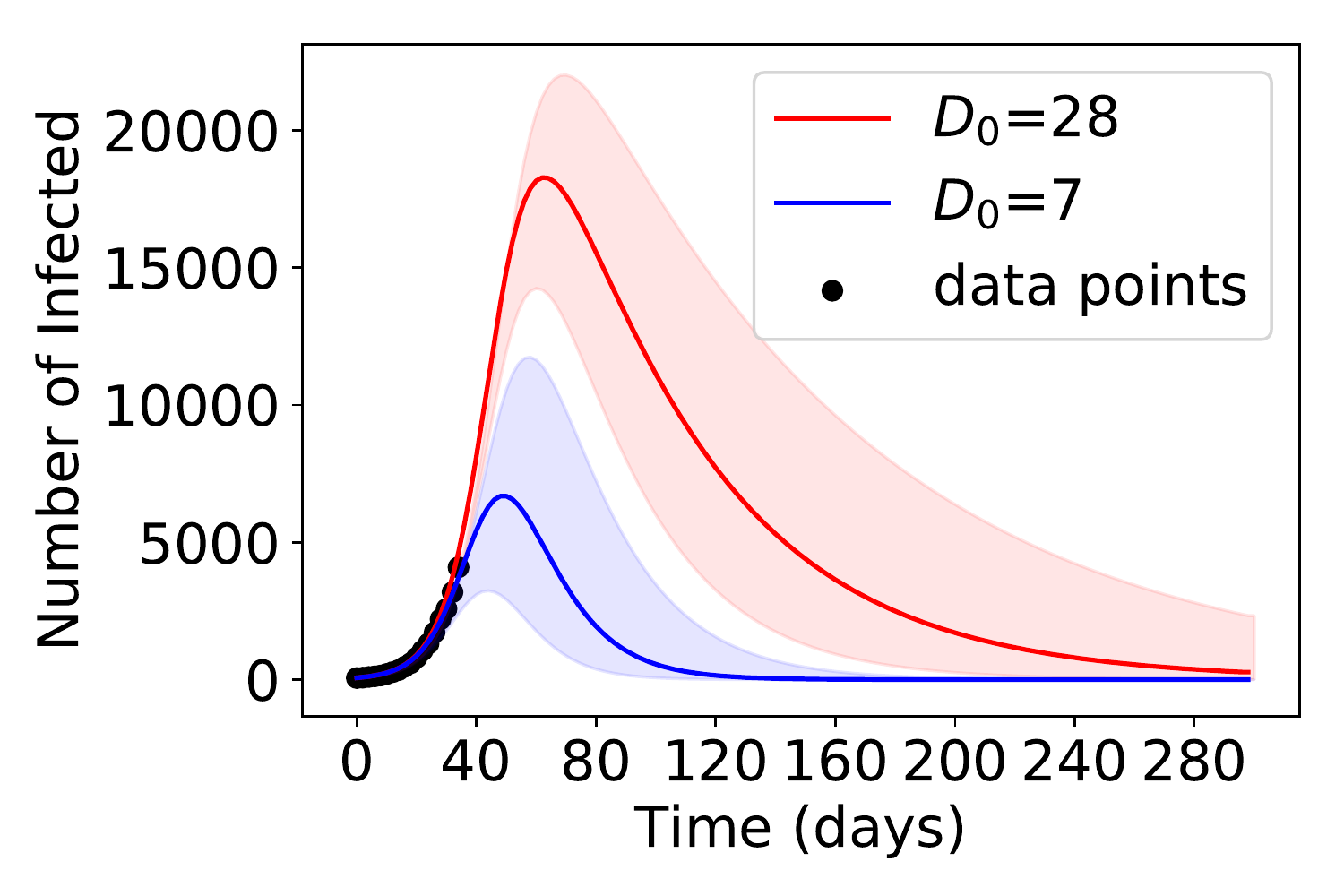}
    \includegraphics[width=0.3\textwidth]{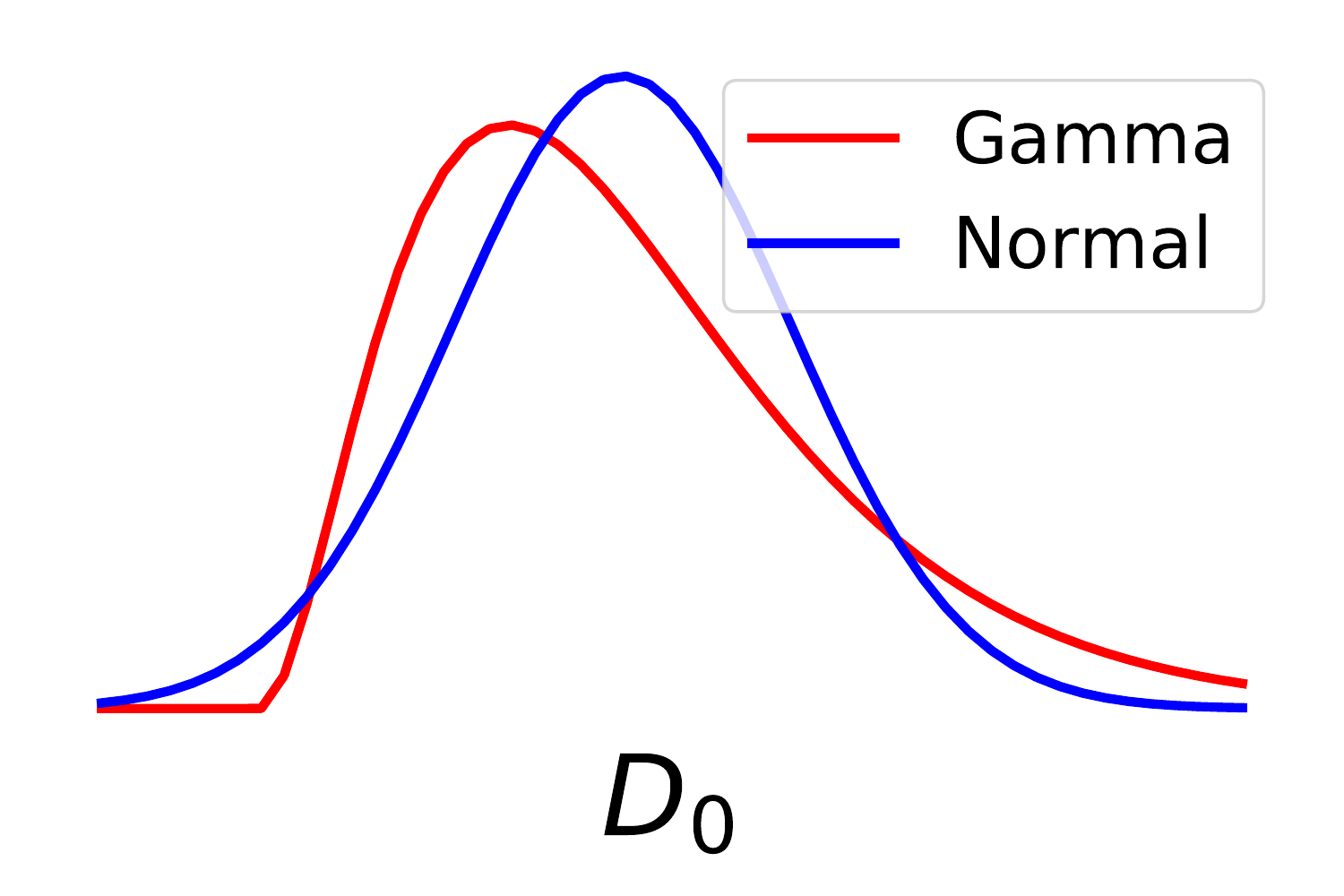}
    \includegraphics[width=0.3\textwidth]{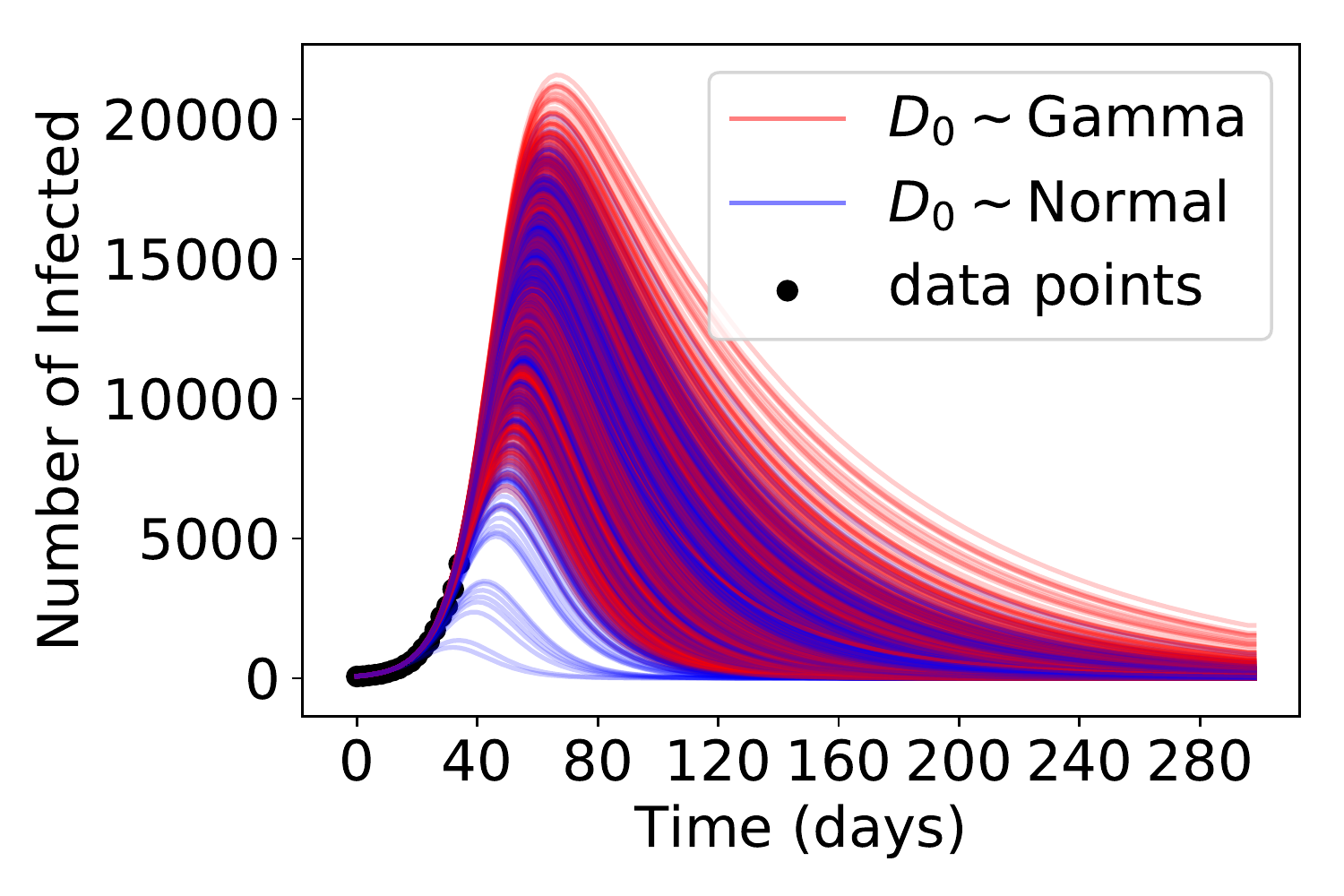}
    \caption{\textbf{Underspecification in a simple epidemiological model.}
        A training pipeline that only minimizes predictive risk on early stages of the epidemic leaves key parameters underspecified, making key behaviors of the model sensitive to arbitrary training choices. Because many parameter values are equivalently compatible with fitting data from early in the epidemic, the trajectory returned by a given training run depends on where it was initialized, and different initialization distributions result in different distributions of predicted trajectories.}
    \label{fig:epi_fig}
\end{figure}

However, when used to forecast the trajectory of the epidemic past $T_{\text{obs}}$, these parameters yield very different predictions.
In Figure~\ref{fig:epi_fig}(a), we show two predicted trajectories of infections corresponding to two parameter sets $(\beta, D)$.
Despite fitting the observed data identically, these models predict peak infection numbers, for example, that are orders of magnitude apart.

Because the training objective cannot distinguish between parameter sets $(\beta, D)$ that yield equivalent growth rates $\beta - 1/D$, arbitrary choices in the learning process determine which set of observation-equivalent parameters are returned by the learning algorithm.
In Figure~\ref{fig:epi_fig}(c), we show that by changing the point $D_0$ at which the parameter $D$ is initialized in the least-squares minimization procedure, we obtain a wide variety of predicted trajectories from the model.
In addition, the particular distribution used to draw $D_0$ (Figure~\ref{fig:epi_fig}(b)) has a substantial influence on the distribution of predicted trajectories.

In realistic epidemiological models that have been used to inform policy, underspecification is dealt with by testing models in forecasting scenarios (i.e., stress testing), and constraining the problem with domain knowledge and external data, for example about viral dynamics in patients (informing $D$) and contact patterns in the population (informing $\beta$) \citep[see, e.g.][]{flaxman2020estimating}.

\def\hyp{{\tau}}
\def\bzero{{\boldsymbol 0}}
\def\bs{{\boldsymbol s}}
\def\bx{{\boldsymbol x}}
\def\bw{{\boldsymbol w}}
\def\bW{{\boldsymbol W}}
\def\btheta{{\boldsymbol \theta}}
\def\bbeta{{\boldsymbol \beta}}
\def\hbtheta{\hat{\boldsymbol \theta}}
\def\bphi{{\boldsymbol \phi}}
\def\<{\langle}
\def\>{\rangle}
\def\sT{{\sf T}}
\def\normal{{\sf N}}
\def\eps{{\varepsilon}}

\def\reals{{\mathbb R}}
\def\strain{\mbox{\rm\tiny train}}
\def\stest{\mbox{\rm\tiny test}}

\def\sshift{\mbox{\rm\tiny shift}}
\def\E{{\sf E}}
\def\P{{\sf P}}
\def\Q{{\sf Q}}
\def\cN{{\mathcal N}}
\def\normal{{\sf N}}

\subsection{Theoretical Analysis of Underspecification in a Random Feature Model}

Underspecification is also a natural consequence of overparameterization, which is a key property of many modern neural network models: when there are more parameters than datapoints, the learning problem is inherently underspecified.
Much recent work has shown that this underspecification has interesting regularizing effects on iid generalization, but there has been little focus on its impact on how models behave on other distributions.
Here, we show that we can recover the effect of underspecification on out-of-distribution generalization in an asymptotic analysis of a simple random feature model, which is often used as a model system for neural networks in the infinitely wide regime. 

We consider for simplicity a regression problem: we are given
data $\{(\bx_i,y_i)\}_{i\le n}$, with $\bx_i\in\reals^d$  
vector of covariates and $y_i\in\reals$ a response.
As a tractable and yet mathematically rich setting, 
we use the random features model of
\citet{neal1996priors} and \citet{rahimi2008random}.
This is a one-hidden-layer neural network with
random first layer weights $\bW$ and learned second layer weights $\btheta$.
We learn a predictor $f_{\bW}:\reals^d\to\reals$ of the form 
$$f_{\bW}(\bx) =\btheta^{\sT}\sigma(\bW\bx).$$
Here, $\bW\in\reals^{N\times d}$
is a random matrix with rows $\bw_{i}\in\reals^d$, $1\le i\le N$
that are not optimized and define the featurization
map $\bx\mapsto\sigma(\bW\bx)$.
We take $(\bw_{i})_{i\le N}$
to be iid and uniformly random with $\|\bw_{i}\|_2=1$. 
We consider data  $(\bx_i,y_i)$, where $\bx_i$ are 
uniformly random with $\|\bx_i\|_2=\sqrt{d}$ and a linear 
target $y_i = f_*(\bx_i)  =\bbeta_0^\sT\bx_i$.

We analyze this model in a setting where both the number of datapoints $n$ and the neurons $N$ both tend toward infinity with a fixed overparameterization ratio $N/n$.
For $N/n<1$, we learn the second layer weights using least squares. For $N/n\ge 1$ there exists choices of the parameters
$\btheta$ that perfectly interpolate the data $f_{\tau}(\bx_i)=y_i$
for all $i\le n$. We choose the minimum $\ell_2$-norm interpolant
(which is the model selected by GD when $\btheta$ is initialized at 0): 
\begin{align*}
    \text{minimize } & \|\btheta\| \\
    \text{subject to } & f_{\tau}(\bx_i)=y_i \text{ for all }i . 
\end{align*}

We analyze the predictive risk of the predictor $f_{\bW}$ on two test distributions, $\trainP{}$, which matches the training distribution, and $\P_{\Delta}$, which is perturbed in a specific way that we describe below.
For a given distribution $\Q$, we define the prediction risk as the mean squared error for the random feature model derived from $\bW$ and for a test point sampled from $\Q$:
$$R(\bW, \Q) = \mathbb{E}_{(X,Y)\sim \Q} ( Y - \hat{\btheta}(\bW) \sigma( \bW X ) )^2.$$
\noindent This risk depends implicitly on the training data through $\hat{\btheta}$, but we suppress this dependence.

Building on the work of \cite{mei2019generalization} we 
can determine the precise asymptotics of
the risk under certain distribution shifts 
in the limit $n,N,d\to\infty$ with fixed ratios $n/d$, $N/n$.
We provide detailed derivations in Appendix~\ref{sec:theory appendix}, as well as characterizations of other quantities such as the sensitivity of the prediction function $f_{\bW}$ to the choice of $\bW$.

In this limit, any two independent random choices $\bW_1$ and $\bW_2$ induce trained predictors $f_{\bW_1}$ and $f_{\bW_2}$ that have indistinguishable in-distribution error $R(\bW_i, \trainP)$.
However, given this value of the risk, the prediction function $f_{\bW_1}(\bx)$ and
$f_{\bW_2}(\bx)$ are nearly as orthogonal as they can be, and this leads to very different test errors on certain shifted distributions $\P_{\Delta}$.

\begin{figure}[t]
  \begin{center}
    \includegraphics[width=0.75\linewidth]{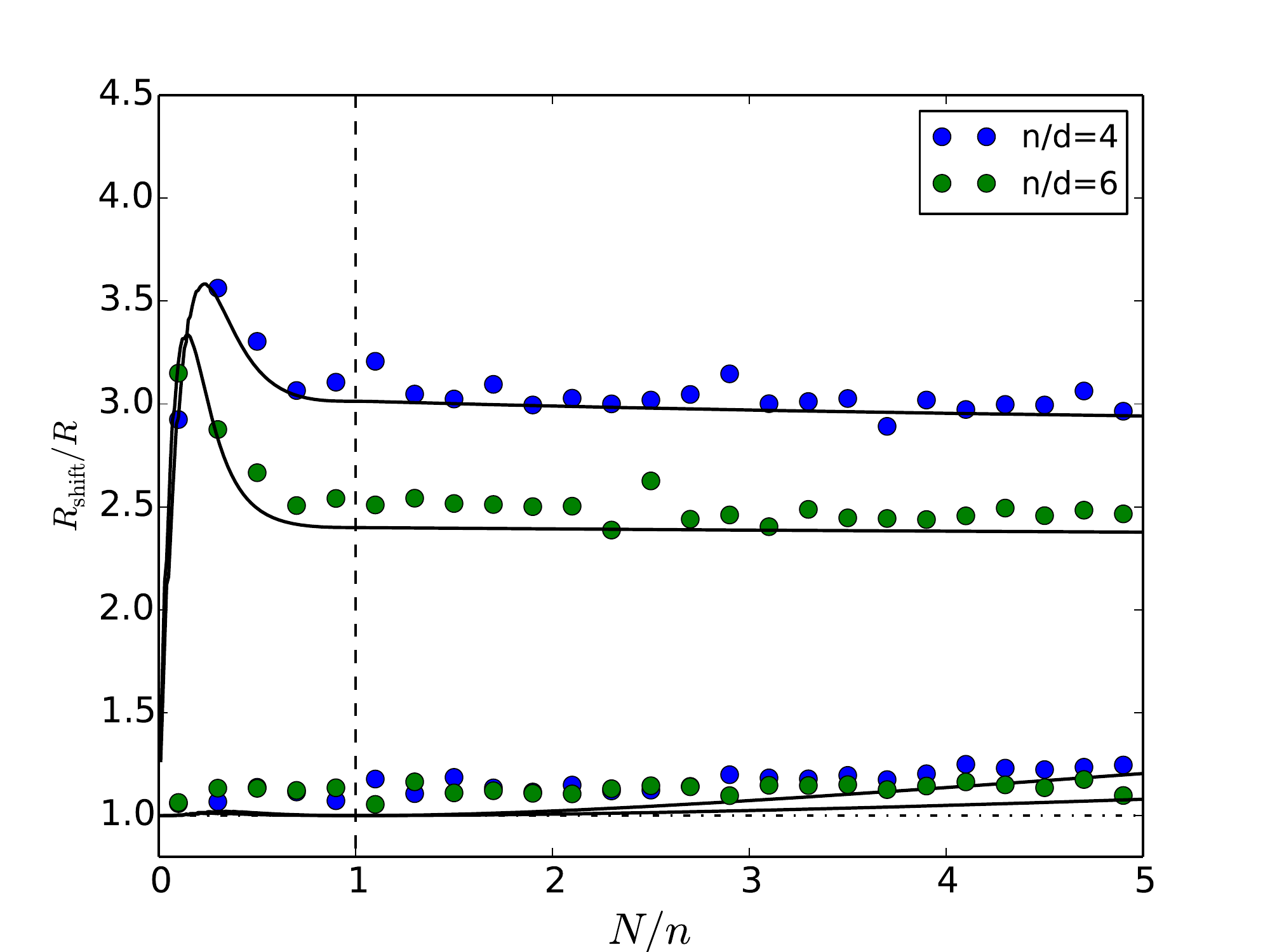}
    \end{center}
    \caption{
    \textbf{Random feature models with identical in-distribution risk show distinct risks under mean shift.}
    Expected risk (averaging over random features $\bW_0$, $\bW$) of predictors $f_{\bW_0}$, $f_{\bW}$ under a $\bW_0$-adversarial mean-shift at different levels of overparameterization $(N / n)$ and sample size-to-parameter ratio $(n / d)$.
      Upper curves: Normalized risk ${\E_{\bW_0} R(\bW_0;\P_{\bW_0,\Delta})}/{\E_{\bW}R(\bW;\P)}$ of the adversarially targeted predictor $f_{\bW_0}$.
      Lower curves: Normalized risk ${\E_{\bW,\bW_0} R(\bW;\P_{\bW_0,\Delta})}/{\E_{\bW}R(\bW;\P)}$ of a predictor $f_{\bW}$ defined with independently draw random weights $\bW$.
      Here the input dimension is $d=80$, $N$ is the number of neurons, and $n$ the number of samples.
      We use ReLU activations; the ground truth is linear with  $\|\bbeta_0\|_2=1$. 
    Circles are empirical results obtained by averaging over $50$  realizations. 
    Continuous lines correspond to the analytical predictions detailed in the supplement.}
    \label{fig:RF-Shift-Theory}
\end{figure}

Specifically, we define $\P_{\Delta}$ in terms of an adversarial mean shift.
We consider test inputs $\bx_\text{test} = \bx_0 + \bx$, where $\bx$ is an independent sample from the training distribution, but $\bx_0$ is a constant mean-shift defined with respect to a fixed set of random feature weights $W_0$.
We denote this shifted distribution with $\P_{\Delta,\bW_0}$.
For a given $\bW_0$, a shift $\bx_0$ can be chosen such that (1) it has small norm ($||\bx_0||<\Delta\ll || \bx ||$), (2) it leaves the risk of an independently sampled $\bW$ mostly unchanged ($R(\bW,\P_{\Delta,\bW_0})\approx R(\bW,\P_\text{train})$), but (3) it drastically increases the risk of $\bW_0$ ($R(\bW_0,\P_{\Delta,\bW_0})> R(\bW_0,\P_\text{train})$).
In Figure~\ref{fig:RF-Shift-Theory} we plot the risks $R(\bW,\P_{\Delta,\bW_0})$ and $R(\bW_0,\P_{\Delta,\bW_0})$ normalized by the iid test risk $R(\bW,\P_\text{train})$ as a function of the overparameterization ratio for two different data dimensionalities.
The upper curves correspond to the risk for the model against which the shift was chosen adversarially, producing a 3-fold increase in risk.
Lower curves correspond to the risk for the same distributional shift for the independent model, resulting in very little risk inflation.

These results show that any predictor selected by min-norm interpolation is vulnerable to shifts along a certain direction, while many other models with equivalent risk are not vulnerable to the same shift.
The particular shift itself depends on a random set of choices made during model training.
Here, we argue that similar dynamics are at play in many modern ML pipelines, under distribution shfits that reveal practically important model properties.
\subsection{Underspecification in a Linear Polygenic Risk Score Model}
\label{sec:genomics}

Polygenic risk scores (PRS) in medical genomics leverage patient genetic information (genotype) to predict clinically relevant characteristics (phenotype). Typically, they are linear models built on categorical features that represent genetic variants.
PRS have shown great success in some settings \citep{Khera2018-prs-diseases}, but face difficulties when applied to new patient populations \citep{Martin2017-prs-ancestry, Duncan2019-prs-ancestry, Berg2019-prs-ancestry}.

We show that underspecification plays a role in this difficulty with generalization.
Specifically, we show that there is a non-trivial set of predictors $\mathcal F^*$ that have near-optimal performance in the training domain, but transfer very differently to a new population.
Thus, a modeling pipeline based on iid performance alone cannot reliably return a predictor that transfers well. 

To construct distinct, near-optimal predictors, we exploit
a core ambiguity in PRS, namely, that many genetic variants that are used as features are nearly collinear.
This collinearity makes it difficult to distinguish causal and correlated-but-noncausal variants \citep{Slatkin-LD}.
A common approach to this problem is to partition variants into clusters of highly-correlated variants and to only include one representative of each cluster in the PRS \citep[e.g., ][]{ISC2009-prs, CAD2013-prs}.
Usually, standard heuristics are applied to choose clusters and cluster representatives as a pre-processing step \citep[e.g., ``LD clumping'', ][]{Purcell2007-plink}.

Importantly, because of the high correlation of features within clusters, the choice of cluster representative leaves the iid risk of the predictor largely unchanged.
Thus, distinct PRS predictors that incorporate different cluster representatives can be treated as members of the risk-minimizing set $\mathcal F^*$.
However, this choice has strong consequences for model generalization.

\begin{figure*}[t]
    \centering
    \includegraphics[width=0.40\linewidth]{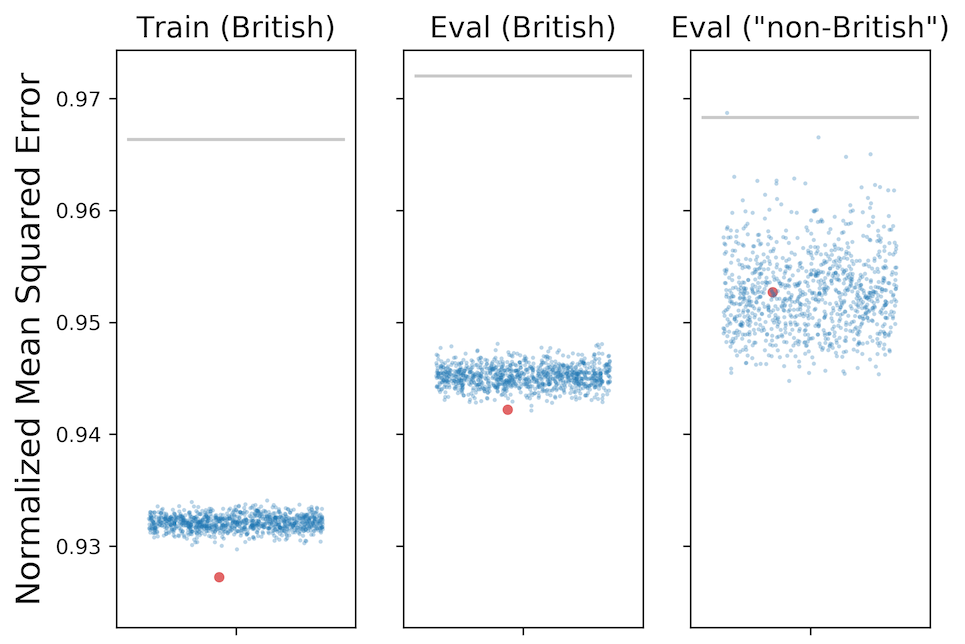}
    \includegraphics[width=0.28\linewidth]{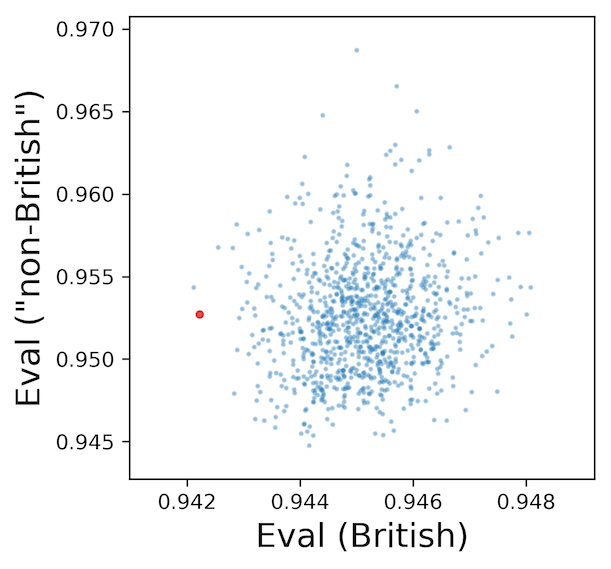}
    \caption{
        \textbf{Underspecification in linear models in medical genomics.}
        \textbf{(Left)} Performance of a PRS model using genetic features in the British training set, the British evaluation set, and the ``non-British'' evaluation set, as measured by the normalized mean squared error (MSE divided by the true variance, lower is better). Each dot represents a PRS predictor (using both genomic and demographic features); large red dots are PRS predictors using the ``index'' variants of the clusters of correlated features selected by PLINK. Gray lines represent the baseline models using only demographic information.
        \textbf{(Right)} Comparison of model performance (NMSE) in British and ``non-British'' eval sets, given the same set of genomic features  (Spearman $\rho=0.135; 95\%\text{ CI }0.070\text{-}0.20$).
    }
    \label{fig:gen_iop_ml4h}
\end{figure*}
To demonstrate this effect, we examine how feature selection influences behavior in a stress test that simulates transfer of PRS across populations.
Using data from the UK Biobank \citep{Sudlow2015-ukb}, we examine how a PRS predicting a particular continuous phenotype called the \emph{intraocular pressure} (IOP) transfers from a predominantly British training population to ``non-British'' test population (see Appendix~\ref{sec:apd_genomics} for definitions).
We construct an ensemble of 1000 PRS predictors that sample different representatives from each feature cluster, including one that applies a standard heuristic from the popular tool PLINK \citep{Purcell2007-plink}.

The three plots on the left side of Figure~\ref{fig:gen_iop_ml4h} confirm that each predictor with distinct features attains comparable performance in the training set and iid test set, with the standard heuristic (red dots) slightly outperforming random representative selection.
However, on the shifted ``non-British'' test data, we see far wider variation in performance, and the standard heuristic fares no better than the rest of the ensemble.
More generally, performance on the British test set is only weakly associated with performance on the ``non-British'' set (Spearman $\rho=0.135; 95\%\text{ CI }0.070\text{-}0.20$; Figure~\ref{fig:gen_iop_ml4h}, right).

Thus, because the model is underspecified, this PRS training pipeline cannot reliably return a predictor that transfers as required between populations, despite some models in $\mathcal F^*$ having acceptable transfer performance.
For full details of this experiment and additional background information, see Appendix~\ref{sec:apd_genomics}.

\section{Underspecification in Deep Learning Models}

Underspecification is present in a wide range of modern deep learning pipelines, and poses an obstacle to reliably learning predictors that encode credible inductive biases.
We show this empirically in three domains: computer vision (including both basic research and medical imaging), natural language processing, and clinical risk prediction using electronic health records.
In each case, we use a simple experimental protocol to show that these modeling pipelines admit a non-trivial set $\mathcal F^*$ of near-optimal predictors, and that different models in $\mathcal F^*$ encode different inductive biases that result in different generalization behavior.

Similarly to our approach in Section~\ref{sec:warmup}, our protocol approaches underspecification constructively by instantiating a set of predictors from the near-optimal set $\mathcal F^*$, and then probing them to show that they encode different inductive biases.
However, for deep models, it is difficult to specify predictors in this set analytically.
Instead, we construct an ensemble of predictors from a given model by perturbing small parts of the ML pipeline (e.g., the random seed used in training, or the recurrent unit in an RNN), and retraining the model several times. 
When there is a non-trivial set $\mathcal F^*$, such small perturbations are often enough to push the pipeline to return a different choice $f \in \mathcal F^*$.
This strategy does not yield an exhaustive exploration of $\mathcal F^*$; rather, it is a conservative indicator of which predictor properties are well-constrained and which are underspecified by the modeling pipeline.

Once we obtain an ensemble, we make several measurements.
First, we empirically confirm that the models in the ensemble have near-equivalent iid performance, and can thus be considered to be members of $\mathcal F^*$.
Secondly, we evaluate the ensemble on one or more application-specific stress tests that probe whether the predictors encode appropriate inductive biases for the application (see Section~\ref{sec:stress tests}).
Variability in stress test performance provides evidence that the modeling pipeline is underspecified along a practically important dimension.

The experimental protocol we use to probe underspecification is closely related to uncertainty quantification approaches based on deep ensembles
\citep[e.g.,][]{Lakshminarayanan2017deepensembles,Dusenberry2020}.
In particular, by averaging across many randomly perturbed predictors from a single modeling pipline, deep ensembles have been shown to be effective tools for detecting out-of-distribution inputs, and correspondingly for tamping down the confidence of predictions for such inputs \citep{snoek2019can}.
Our experimental strategy and the deep ensembles approach can be framed as probing a notion of model stability resulting from perturbations to the model, even when the data are held constant \citep{yu2013stability}.

To establish that observed variability in stress test performance is a genuine indicator of underspecification, we evaluate three properties.
\begin{itemize}
\item First, we consider the \emph{magnitude} of the variation, either relative to iid performance (when they are on the same scale), or relative to external benchmarks, such as comparisons between ML pipelines with featuring different model architectures.
\item Secondly, when sample size permits, we consider \emph{unpredictability} of the variation from iid performance.
Even if the observed differences in iid performance in our ensemble is small, if stress test performance tracks closely with iid performance, this would suggest that our characterization of $\mathcal F^*$ is too permissive.
We assess this with the Spearman rank correlation between the iid validation metric and the stress test metric.
\item Finally, we establish that the variation in stress tests indicates \emph{systematic differences} between the predictors in the ensemble.
Often, the magnitude of variation in stress test performance alone will be enough to establish systematicness.
However, in some cases we supplement with a mixture of quantitative and qualitative analyses of stress test outputs to illustrate that the differences between models does align with important dimensions of the application.
\end{itemize}

In all cases that we consider, we find evidence that important inductive biases are underspecified.
In some cases the evidence is obvious, while in others it is more subtle, owing in part to the conservative nature of our exploration of $\mathcal F^*$.
Our results interact with a number of research areas in each of the fields that we consider, so we close each case study with a short application-specific discussion.

\section{Case Studies in Computer Vision}
\label{sec:vision}

Computer vision is one of the flagship application areas in which deep learning on large-scale training sets has advanced the state of the art.
Here, we focus on an image classification task, specifically on the ImageNet validation set \citep{imagenet_cvpr09}.
We examine two models: the ResNet-50 model \citep{he2016deep} trained in ImageNet, and a ResNet-101x3 Big Transfer (BiT) model \citep{kolesnikov2019large} pre-trained on the JFT-300M dataset \citep{sun2017revisiting} and fine-tuned on ImageNet.
The former is a standard baseline in image classification.
The latter is scaled-up ResNet designed for transfer learning, which attains state-of-the-art, or near state-of-the-art, on many image classification benchmarks, including ImageNet.

A key challenge in computer vision is robustness under distribution shift.
It has been well-documented that many deep computer vision models suffer from brittleness under distribution shifts that humans do not find challenging \citep{goodfellow2016deep,hendrycks2018benchmarking,barbu2019objectnet}.
This brittleness has raised questions about deployments open-world high-stakes applications.
This has given rise to an active literature on robustness in image classification \citep[see, e.g.,][]{taori2020measuring,djolonga2020robustness}.
Recent work has connected lack of robustness to computer vision models' encoding counterintuitive inductive biases \citep{ilyas2019adversarial,geirhos2018imagenettrained,yin2019fourier,wang2020high}.

Here, we show concretely that the models we study are underspecified in ways that are important for robustness to distribution shift.
We apply our experimental protocol to show that there is substantial ambiguity in how image classification models will perform under distribution shift, even when their iid performance is held fixed.
Specifically, we construct ensembles of the ResNet-50 and BiT models to stress test:
we train 50 ResNet-50 models on ImageNet using identical pipelines that differ only in their random seed, 30 BiT models that are initialized at the same JFT-300M-trained checkpoint, and differ only in their fine-tuning seed and initialization distributions (10 runs each of zero, uniform, and Gaussian initializations).
On the ImageNet validation set, the ResNet-50 predictors achieve a $75.9\% \pm 0.11$ top-1 accuracy, while the BiT models achieve a $86.2\% \pm 0.09$ top-1 accuracy.

We evaluate these predictor ensembles on two stress tests that have been proposed in the image classification robustness literature: ImageNet-C \citep{hendrycks2018benchmarking} and ObjectNet \citep{barbu2019objectnet}.
ImageNet-C is a benchmark dataset that replicates the ImageNet validation set, but applies synthetic but realistic corruptions to the images, such as pixelation or simulated snow, at varying levels of intensity.
ObjectNet is a crowdsourced benchmark dataset designed to cover a set of classes included in the ImageNet validation set, but to vary the settings and configurations in which these objects are observed.
Both stress tests have been used as prime examples of the lack of human-like robustness in deep image classification models.

\subsection{ImageNet-C}
\begin{figure}
    \centering
    \includegraphics[origin=c,width=\textwidth]{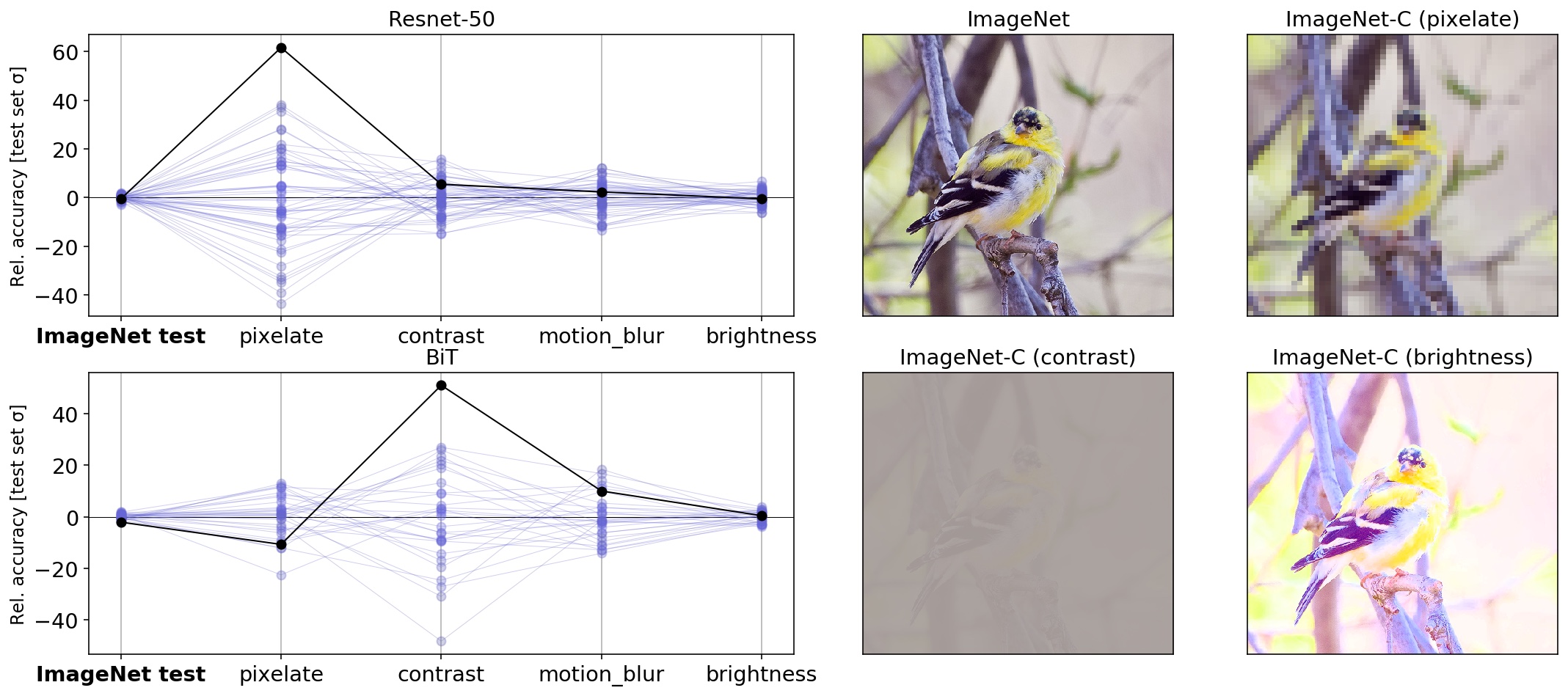}
    \caption{\textbf{Image classification model performance on stress tests is sensitive to random initialization in ways that are not apparent in iid evaluation.} \textbf{(Top Left)} Parallel axis plot showing variation in accuracy between identical, randomly initialized ResNet 50 models on several ImageNet-C tasks at corruption strength 5.
    Each line corresponds to a particular model in the ensemble; each each parallel axis shows deviation from the ensemble mean in accuracy, scaled by the standard deviation of accuracies on the ``clean'' ImageNet test set.
    On some tasks, variation in performance is orders of magnitude larger than on the standard test set.
    \textbf{(Right)} Example image from the standard ImageNet test set, with corrupted versions from the ImageNet-C benchmark.
    }
    \label{fig:image_results}
\end{figure}

\begin{figure}
    \centering
    \includegraphics[width=0.7\textwidth]{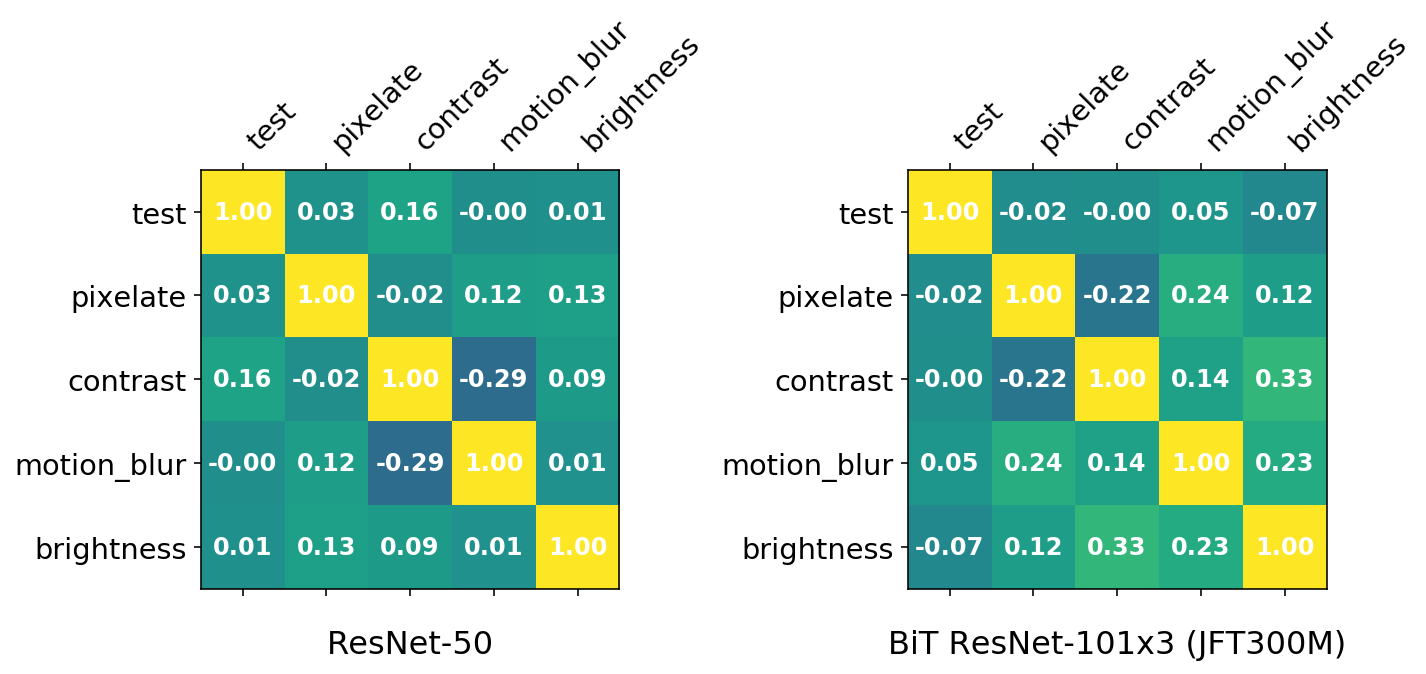}
    \caption{\textbf{Performance on ImageNet-C stress tests is unpredictable from standard test performance.}
    Spearman rank correlations of predictor performance, calculated from random initialization predictor ensembles.
    (Left) Correlations from 50 retrainings of a ResNet-50 model on ImageNet.
    (Right) Correlations from 30 ImageNet fine-tunings of a ResNet-101x3 model pre-trained on the JFT300M dataset.
    }
    \label{fig:vision correlations}
\end{figure}

We show results from the evaluation on several ImageNet-C tasks in Figure~\ref{fig:image_results}.
The tasks we show here incorporate corruptions at their highest intensity levels (level 5 in the benchmark). 
In the figure, we highlight variability in the accuracy across predictors in the ensemble, relative to the variability in accuracy on the standard iid test set.
For both the ResNet-50 and BiT models, variation on some ImageNet-C tasks is an order of magnitude larger than variation in iid performance.
Furthermore, within this ensemble, there is weak sample correlation between performance on the iid test set and performance on each benchmark stress test, and performance between tasks (all 95\% CI's for Pearson correlation using $n=50$ and $n=30$ contain zero, see Figure~\ref{fig:vision correlations}).
We report full results on model accuracies and ensemble standard deviations in Table~\ref{tab:imagenetc table}.

\subsection{ObjectNet}
We also evaluate these ensembles along more ``natural'' shifts in the ObjectNet test set.
Here, we compare the variability in model performance on the ObjectNet test set to a subset of the standard ImageNet test set with the 113 classes that appear in ObjectNet.
The results of this evaluation are in Table~\ref{tab:imagenetc table}.
The relative variability in accuracy on the ObjectNet stress test is larger that the variability seen in the standard test set (standard deviation is 2x for ResNet-50 and 5x for BiT), although the difference in magnitude is not as striking as in the ImageNet-C case.
There is also a slightly stronger relationship between standard test accuracy and test accuracy on ObjectNet (Spearman $\rho$ 0.22 $(-0.06, 0.47)$ for ResNet-50, 0.47 $(0.13, 71)$ for BiT).

Nonetheless, the variability in accuracy suggests that some predictors in the ensembles are systematically better or worse at making predictions on the ObjectNet test set.
We quantify this with p-values from a one-sided permutation test, which we interpret as descriptive statistics.
Specifically, we compare the variability in model performance on the ObjectNet test set with variability that would be expected if prediction errors were randomly distributed between predictors.
The variability of predictor accuracies on ObjectNet is large compared to this baseline ($p=0.002$ for ResNet-50 and $p = 0.000$ for BiT).
On the other hand, the variability between predictor accuracies on the standard ImageNet test set are more typical of what would be observed if errors were randomly distributed ($p=0.203$ for ResNet-50 and $p=0.474$ for BiT).
In addition, the predictors in our ensembles disagree far more often on the ObjectNet test set than they do in the ImageNet test set, whether or not we consider the subset of the ImageNet test set examples that have classes that appear in ObjectNet (Table~\ref{tab:objectnet table}).

\subsection{Conclusions}
These results indicate that the inductive biases that are relevant to making predictions in the presence of these corruptions are so weakly differentiated by the iid prediction task that changing random seeds in training can cause the pipeline to return predictors with substantially different stress test performance.
The fact that underspecification persists in the BiT models is particularly notable, because simultaneously scaling up data and model size has been shown to improve performance across a wide range of robustness stress tests, aligning closely with how much these models improve performance on iid evaluations \citep{djolonga2020robustness,taori2020measuring,hendrycks2020many}.
Our results here suggest that underspecification remains an issue even for these models; potentially, as models are scaled up, underspecified dimensions may account for a larger proportion of the ``headroom'' available for improving out-of-distribution model performance. 

\begin{table}
    \centering
    \footnotesize
  \begin{tabular}{lrrrrrr}
    \toprule
    Dataset &  ImageNet &  pixelate &  contrast &  motion blur&  brightness & ObjectNet\\
    \midrule
    ResNet-50 &                 0.759 (0.001) &            0.197 (0.024) &            0.091 (0.008) &               0.100 (0.007) &              0.607 (0.003) &          0.259 (0.002)\\
    BiT &                 0.862 (0.001) &            0.555 (0.008) &            0.462 (0.019) &               0.515 (0.008) &              0.723 (0.002) &                0.520 (0.005) \\
    \bottomrule
    \end{tabular}
 
    \caption{\textbf{Accuracies of ensemble members on stress tests.} Ensemble mean (standard deviations) of accuracy proportions on ResNet-50 and BiT models.}
    \label{tab:imagenetc table}
\end{table}

\begin{table}
    \centering
    \begin{tabular}{lrrr}
    \toprule
    Dataset &       ImageNet & ImageNet (subset) &      ObjectNet \\
    \midrule
    ResNet-50 &  0.160 (0.001) &     0.245 (0.005) &  0.509 (0.003) \\
    BiT       &  0.064 (0.004) &     0.094 (0.006) &  0.253 (0.012) \\
    \bottomrule
    \end{tabular}
    \caption{\textbf{Ensemble disagreement proportions for ImageNet vs ObjectNet models.} Average disagreement between pairs of predictors in the ResNet and BiT ensembles. The ``subset'' test set only includes classes that also appear in the ObjectNet test set. Models show substantially more disagreement on the ObjectNet test set.}
    \label{tab:objectnet table}
\end{table}

\section{Case Studies in Medical Imaging}

Medical imaging is one of the primary high-stakes domains where deep image classification models are directly applicable. 
In this section, we examine underspecification in two medical imaging models designed for real-world deployment.
The first classifies images of patient retinas, while the second classifies clinical images of patient skin. 
We show that these models are underspecified along dimensions that are practically important for deployment. These results confirm the need for explicitly testing and monitoring ML models in settings that accurately represent the deployment domain, as codified in recent best practices \citep{collins2015transparent,kelly2019key,rivera2020guidelines,liu2020reporting}.

\subsection{Ophthalmological Imaging}
\label{sec:retina}

\begin{figure*}[!t]
    \centering
    \includegraphics[width=\linewidth]{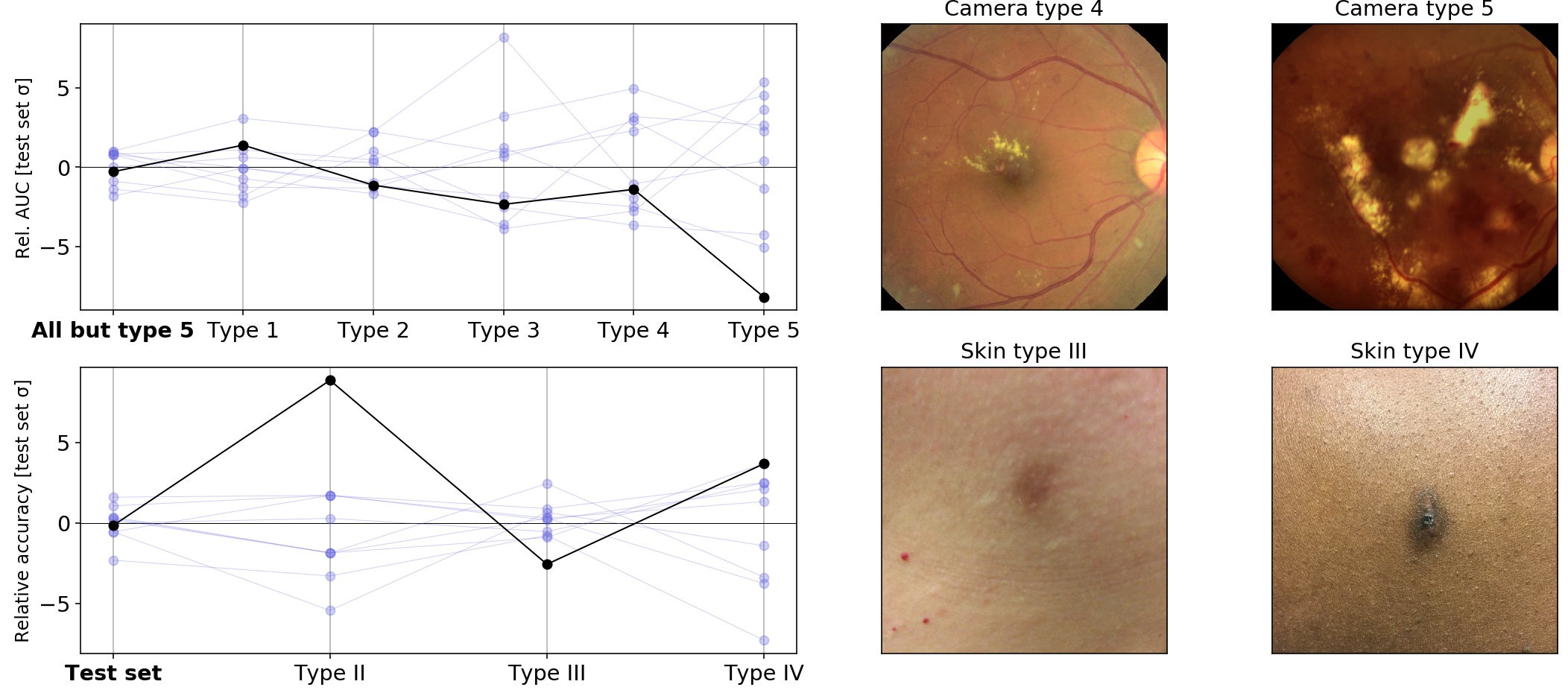}
    \caption{\textbf{Stress test performance varies across identically trained medical imaging models.} 
    Points connected by lines represent metrics from the same model, evaluated on an iid test set (bold) and stress tests.
    Each axis shows deviations from the ensemble mean, divided by the standard deviation for that metric in the standard iid test set. 
    These models differ only in random initialization at the fine-tuning stage.
    \textbf{(Top Left)} 
    Variation in AUC between identical diabetic retinopathy classification models when evaluated on images from different camera types.
    Camera type 5 is a camera type that was not encountered during training.
    \textbf{(Bottom Left)} Variation in accuracy between identical skin condition classification models when evaluated on different skin types.
    \textbf{(Right)} Example images from the original test set (left) and the stress test set (right). Some images are cropped to match the aspect ratio.
    }
    \label{fig:med_image_results}
\end{figure*}

\begin{figure*}
\centering
\includegraphics[width=1.0\linewidth]{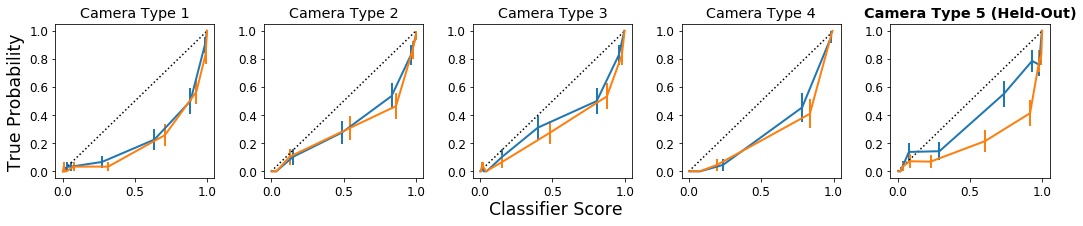}
\caption{\textbf{Identically trained retinal imaging models show systematically different behavior on stress tests.} Calibration plots for two diabetic retinopathy classifiers (orange and blue) that differ only in random seed at fine-tuning. Calibration characteristics of the models are nearly identical for each in-distribution camera type 1--4, but are qualitatively different for the held-out camera type 5.
Error bars are $\pm 2$ standard errors.
}
\label{fig:calibration}
\end{figure*}
Deep learning models have shown great promise in the ophthalmological domain \citep{gulshan2016development, ting2017development}.
Here, we consider one such model trained to predict diabetic retinopathy (DR) and referable diabetic macular edema (DME) from retinal fundus images.
The model employs an Inception-V4 backbone \citep{szegedy2017inception} pre-trained on ImageNet, and fine-tuned using de-identified retrospective fundus images from EyePACS in the United States and from eye hospitals in India.
Dataset and model architecture details are similar to those in \citep{krause2018grader}.

A key use case for these models is to augment human clinical expertise in underserved settings, where doctor capacity may be stretched thin.
As such, generalization to images taken by a range of cameras, including those deployed at different locations and clinical settings, is essential for system usability
\citep{beede2020human}.

Here, we show that the performance of predictors produced by this model is sensitive to underspecification.
Specifically, we construct an ensemble of 10 models that differ only in random initialization at the fine-tuning stage.
We evaluate these models on stress tests predicting DR using camera type images not encountered during training.

The results are shown in Figure~\ref{fig:med_image_results}.
Measuring accuracy in terms of AUC, variability in AUC on the held-out camera type is larger than that in the standard test set, both in aggregate, and compared to most strata of camera types in the training set.
To establish that this larger variability is not easily explained away by differences in sample size, we conduct a two-sample z-test comparing the AUC standard deviation in the held-out camera test set ($n=287$) against the AUC standard deviation in the standard test set ($n=3712$) using jackknife standard errors, obtaining a z-value of $2.47$ and a one-sided p-value of $0.007$.
In addition, models in the ensemble differ systematically in ways that are not revealed by performance in the standard test set.
For example, in Figure~\ref{fig:calibration}, we show calibration plots of two models from the ensemble computed across camera types.  
The models have similar calibration curves for the cameras encountered during training, but have markedly different calibration curves for the held-out camera type.
This suggests that these predictors process images in systematically different ways that only become apparent when evaluated on the held-out camera type.

\subsection{Dermatological Imaging}
Deep learning based image classification models have also been explored for applications in dermatology \citep{esteva2017dermatologist}.
Here, we examine a model proposed in \citet{liu2020deep} that is trained to classify skin conditions from clinical skin images.
As in Section~\ref{sec:retina}, this model incorporates an ImageNet--pre-trained Inception-V4 backbone followed by fine-tuning.

In this setting, one key concern is that the model may have variable performance across skin types, especially when these skin types are differently represented in the training data.
Given the social salience of skin type,
this concern is aligned with broader concerns about ensuring that machine learning does not amplify existing healthcare disparities \citep{adamson2018machine}.
In dermatology in particular, differences between the presentation of skin conditions across skin types has been linked to disparities in care \citep{adelekun2020skin}.

Here, we show that model performance across skin types is sensitive to underspecification.
Specifically, we construct an ensemble of 10 models with randomly initialized fine-tuning layer weights.
We then evaluate the models on a stress test that stratifies the test set by skin type on the Fitzpatrick scale \citep{fitzpatrick1975sun} and measures Top-1 accuracy within each slice.

The results are shown at the bottom of Figure~\ref{fig:med_image_results}.
Compared to overall test accuracy, there is larger variation in test accuracy within skin type strata across models, particularly in skin types II and IV, which form substantial portions ($n = 437$, or $10.7\%$,  and $n = 798$, or $19.6\%$, respectively) of the test data.
Based on this test set, some models in this ensemble would be judged to have higher discrepancies across skin types than others, even though they were all produced by an identical training pipeline. 

Because the sample sizes in each skin type stratum differ substantially, we use a permutation test to explore the extent to which the larger variation in some subgroups can be accounted for by sampling noise.
In particular, the larger variation within some strata could be explained by either sampling noise driven by smaller sample sizes, or by systematic differences between predictors that are revealed when they are evaluated on inputs whose distribution departs from the overall iid test set.
This test shuffles the skin type indicators across examples in the test set, then calculates the variance of the accuracy across these random strata. 
We compute one-sided p-values with respect to this null distribution and interpret them as exploratory descriptive statistics.
The key question is whether the larger variability in some strata, particularly skin types II and IV, can be explained away by sampling noise alone. (Our expectation is that skin type III is both large enough and similar enough to the iid test set that its accuracy variance should be similar to the overall variance, and the sample size for skin type V is so small that a reliable characterization would be difficult.)
Here, we find that the variation in accuracy in skin types III and V are easily explained by sampling noise, as expected ($p=0.54, n=2619$; $p=0.42, n=109$).
Meanwhile the variation in skin type II is largely consistent with sampling noise ($p=0.29, n=437$), but the variation in skin type IV seems to be more systematic ($p=0.03, n=798$).
These results are exploratory, but they suggest a need to pay special attention to this dimension of underspecification in ML models for dermatology.

\subsection{Conclusions}
Overall, the vignettes in this section demonstrate that underspecification can introduce complications for deploying ML, even in application areas where it has the potential to highly beneficial.
In particular, these results suggest that one cannot expect ML models to automatically generalize to new clinical settings or populations, because the inductive biases that would enable such generalization are underspecified.
This confirms the need to tailor and test models for the clinical settings and population in which they will be deployed. 
While current strategies exist to mitigate these concerns, addressing underspecification, and generalization issues more generally, could reduce a number of points of friction at the point of care \citep{beede2020human}.
\section{Case Study in Natural Language Processing}

Deep learning models play a major role in modern natural language processing (NLP).
In particular, large-scale Transformer models~\citep{vaswani2017attention} trained on massive unlabeled text corpora have become a core component of many NLP pipelines~\citep{devlin2019bert}.
For many applications, a successful recipe is to ``pretrain'' by applying a masked language modeling objective to a large generic unlabeled corpus, and then fine-tune using labeled data from a task of interest, sometimes no more than a few hundred examples~\citep[e.g.,][]{howard2018universal,peters2018deep}
This workflows has yielded strong results across a wide range of tasks in natural language processing, including machine translation, question answering, summarization, sequence labeling, and more.
As a result, a number of NLP products are built on top of publicly released pretrained checkpoints of language models such as BERT~\citep{devlin2019bert}. 

However, recent work has shown that NLP systems built with this pattern often rely on ``shortcuts''~\citep{geirhos2020shortcut}, which may be based on spurious phenomena in the training data~\citep{mccoy2019right}.
Shortcut learning presents a number of difficulties in natural language processing: failure to satisfy intuitive invariances, such as invariance to typographical errors or seemingly irrelevant word substitutions~\citep{ribeiro2020beyond}; ambiguity in measuring progress in language understanding~\citep{zellers2019hellaswag}; and reliance on stereotypical associations with race and gender~\citep{caliskan2017semantics,rudinger2018gender,zhao2018gender,de2019bias}.

In this section, we show that underspecification plays a role in shortcut learning in the pretrain/fine-tune approach to NLP, in both stages.
In particular, we show that reliance on specific shortcuts can vary substantially between predictors that differ only in their random seed at fine-tuning or pretraining time. Following our experimental protocol, we perform this case study with an ensemble of predictors obtained from identical training pipelines that differ only in the specific random seed used at pretraining and/or fine-tuning time.
Specifically, we train 5 instances of the BERT ``large-cased'' language model~\citet{devlin2019bert}, using the same Wikipedia and BookCorpus data that was used to train the public checkpoints.
This model has 340 million parameters, and is the largest BERT model with publicly released pretraining checkpoints. 
For tasks that require fine-tuning, we fine-tune each of the five checkpoints 20 times using different random seeds.

In each case, we evaluate the ensemble of models on stress tests designed to probe for specific shortcuts, focusing on shortcuts based on stereotypical correlations, and find evidence of underspecification along this dimension in both pretraining and fine-tuning.
As in the other cases we study here, these results suggest that shortcut learning is not enforced by model architectures, but can be a symptom of ambiguity in model specification.

Underspecification has a wider range of implications in NLP.
In the supplement, we connect our results to instability that has previously been reported on stress tests designed to diagnose ``cheating'' on Natural Language Inference tasks \citep{mccoy2019right,naik2018stress}. 
Using the same protocol, we replicate the results \citep{mccoy2019berts,dodge2020fine,zhou2020curse}, and extend them to show sensitivity to the pretraining random seed.
We also explore how underspecification affects inductive biases in static word embeddings.

\subsection{Gendered Correlations in Downstream Tasks}

We begin by examining gender-based shortcuts on two previously proposed benchmarks: a semantic textual similarity (STS) task and a pronoun resolution task. 

\subsubsection{Semantic textual similarity (STS)}
In the STS task, a predictor takes in two sentences as input and scores their similarity.
We obtain predictors for this task by fine-tuning BERT checkpoints on the STS-B benchmark~\citep{cer-etal-2017-semeval}, which is part of the GLUE suite of benchmarks for representation learning in NLP~\citep{wang2018glue}.
Our ensemble of predictors achieves consistent accuracy, measured in terms of correlation with human-provided similarity scores, ranging from $0.87$ to $0.90$. This matches reported results from \cite{devlin2019bert}, although better correlations have subsequently been obtained by pretraining on larger datasets~\citep{liu2019roberta,lan2019albert,yang2019xlnet}.

To measure reliance on gendered correlations in the STS task, we use a set of challenge templates proposed by \citet{webster2020measuring}: we create a set of triples in which the noun phrase in a given sentence is replaced by a profession, ``a man'', or ``a woman'', e.g., ``a doctor/woman/man is walking.'' 
The model's gender association for each profession is quantified by the \emph{similarity delta} between pairs from this triple, e.g., 
\begin{equation*}
    \text{sim}(\text{``a woman is walking''}, \text{``a doctor is walking''})
    -
    \text{sim}(\text{``a man is walking''}, \text{``a doctor is walking''}).
\end{equation*}
A model that does not learn a gendered correlation for a given profession will have an expected similarity delta of zero. We are particularly interested in the extent to which the similarity delta for each profession correlates with the percentage of women actually employed in that profession, as measured by U.S. Bureau of Labor Statistics~\citep[BLS;][]{rudinger2018gender}.

\subsubsection{Pronoun resolution}
In the pronoun resolution task, the input is a sentence with a pronoun that could refer to one of two possible antecedents, and the predictor must determine which of the antecedents is the correct one.
We obtain predictors for this task by fine-tuning BERT checkpoints on the OntoNotes dataset~\citep{hovy-etal-2006-ontonotes}. Our ensemble of predictors achieves accuracy ranging from $0.960$ to $0.965$.

To measure gendered correlations on the pronoun resolution task, we use the challenge templates proposed by~\citet{rudinger2018gender}.
In these templates, there is a gendered pronoun with two possible antecedents, one of which is a profession.
The linguistic cues in the template are sufficient to indicate the correct antecedent, but models may instead learn to rely on the correlation between gender and profession.
In this case, the similarity delta is the difference in predictive probability for the profession depending on the gender of the pronoun.

\subsubsection{Gender correlations and underspecification} 
\begin{figure}
    \centering
    \includegraphics[width=\textwidth]{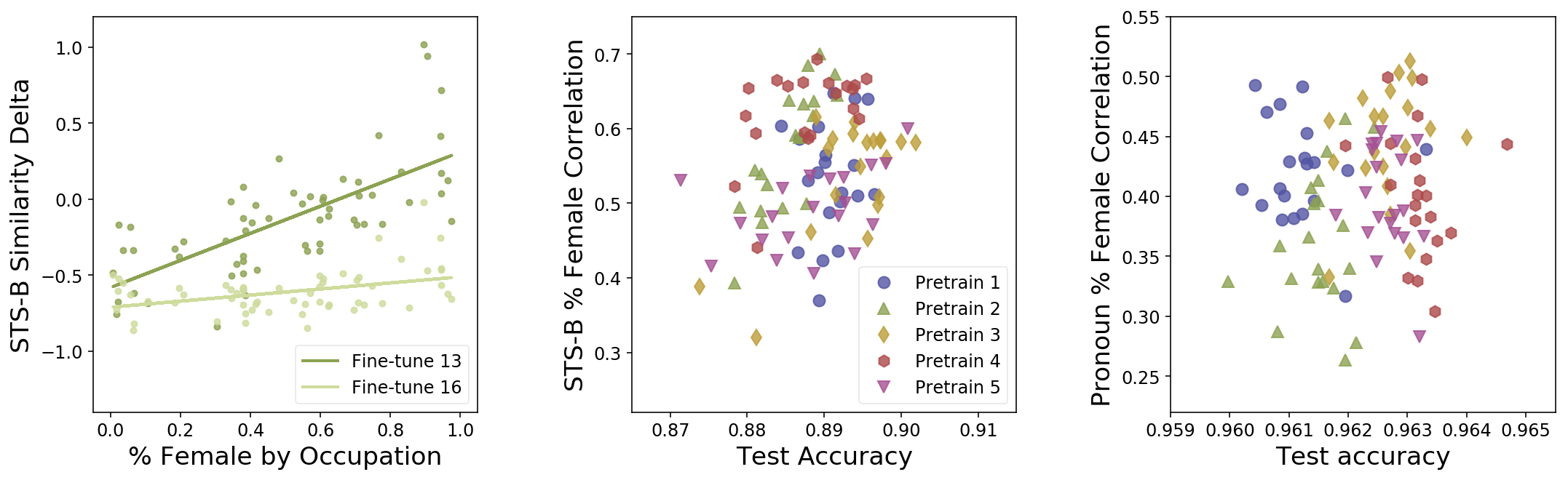}
    \caption{
    \textbf{Reliance on gendered correlations is affected by random initialization.}
    \textbf{(Left)} The gap in similarity for female and male template sentences is correlated with the gender statistics of the occupation, shown in two randomly-initialized fine-tunes. \textbf{(Right)} Pretraining initialization significantly affects the distribution of gender biases encoded at the fine-tuning stage.
    }
    \label{fig:cheating_NLP_EHR}
\end{figure}

We find significant variation in the extent to which the models in our ensemble incorporate gendered correlations.
For example, in \autoref{fig:cheating_NLP_EHR} (Left), we contrast the behavior of two predictors (which differ only in pretraining and fine-tuning seed) on the STS task.  
Here, the slope of the line is a proxy for the predictor's reliance on gender. One fine-tuning run shows strong correlation with BLS statistics about gender and occupations in the United States, while another shows a much weaker relationship.
For an aggregate view, Figures~\ref{fig:cheating_NLP_EHR} (Center) and (Right) show these correlations in the STS and coreference tasks across all predictors in our ensemble, with predictors produced from different pretrainings indicated by different markers.
These plots show three important patterns:
\begin{enumerate}
\item There is a large spread in correlation with BLS statistics: on the STS task, correlations range from $0.3$ to $0.7$; on the pronoun resolution task, the range is $0.26$ to $0.51$. 
As a point of comparison, prior work on gender shortcuts in pronoun resolution found correlations ranging between $0.31$ and $0.55$ for different types of models~\citep{rudinger2018gender}.
\item There is a weak relationship between test accuracy performance and gendered correlation (STS-B: Spearman $\rho=0.21$; $95\%$ CI $=(0.00, 0.39)$, Pronoun resolution: Spearman $\rho=0.08$; $95\%$ CI $=(-0.13, 0.29)$). 
This indicates that learning accurate predictors does \emph{not} require learning strong gendered correlations.
\item Third, the encoding of spurious correlations is sensitive to the random seed at pretraining, and not just fine-tuning.
Especially in the pronoun resolution task, (\autoref{fig:cheating_NLP_EHR}(Right)) predictors produced by different pretraining seeds cluster together, tending to show substantially weaker or stronger gender correlations.
\end{enumerate}

In \autoref{tab:sts_coref_tab}, we numerically summarize the variance with respect to pretraining and fine-tuning using an $F$ statistic --- the ratio of between-pretraining to within-pretraining variance.
The pretraining seed has an effect on both the main fine-tuning task and the stress test, but the small correlation between the fine-tuning tasks and stress test metrics suggests that this random seed affects these metrics independently.

\begin{table}[]
    \centering
\begin{tabular}{lrr}
\toprule
                      & $F$ ($p$-value)   &  Spearman $\rho$ (95\% CI)\\
\midrule
\textbf{Semantic text similarity (STS)}\\ 
Test Accuracy           &  5.66 (4e-04) &  --- \\
Gender Correlation       &  9.66 (1e-06) & 0.21 (-0.00,  0.40) \\[1ex]
\textbf{Pronoun resolution} \\
Test Accuracy               & 48.98 (3e-22) &  --- \\
Gender Correlation    &  7.91 (2e-05) & 0.08 (-0.13,  0.28) \\
\bottomrule
\end{tabular}
    \caption{\textbf{Summary statistics for structure of variation on gendered shortcut stress tests.}
    For each dataset, we measure the accuracy of 100 predictors, corresponding to 20 randomly initialized fine-tunings from 5 randomly initialized pretrained BERT checkpoints.
    Models are fine-tuned on the STS-B and OntoNotes training sets, respectively.
    The $F$ statistic quantifies how systematic differences are between pretrainings using the ratio of within-pretraining variance to between-pretraining variance in the accuracy statistics.
    $p$-values are reported to give a sense of scale, but not for inferential purposes; it is unlikely that assumptions for a valid $F$-test are met.
    $F$-values of this magnitude are consistent with systematic between-group variation. 
    The Spearman $\rho$ statistic quantifies how ranked performance on the fine-tuning task correlates with the stress test metric of gender correlation.
    }
    \label{tab:sts_coref_tab}
\end{table}


\begin{figure}
    \centering
    \includegraphics[width=\textwidth]{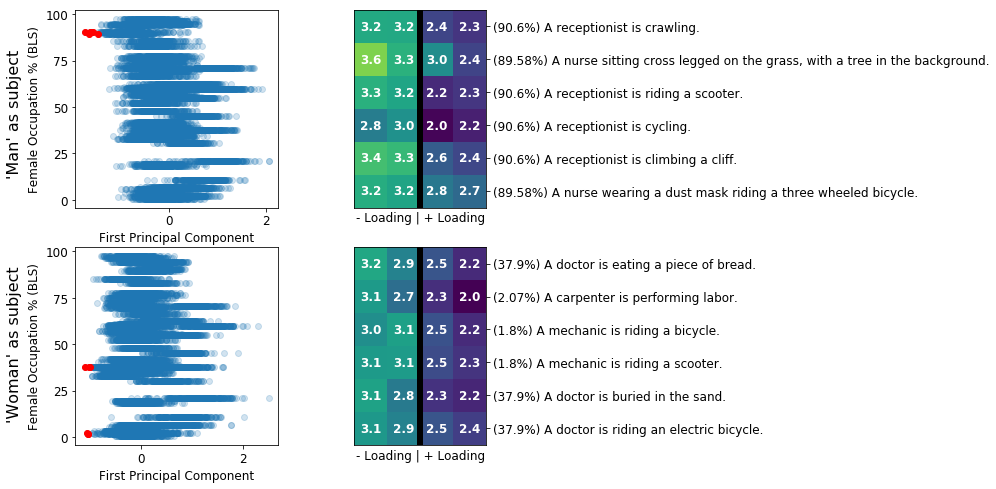}
\caption{\textbf{The first principal axis of model disagreement predicts differences in handling stereotypes.} The first principal component of BERT models fine-tuned for STS-B, against the \% female participation of a profession in the BLS data. The top panel shows examples with a male subject (e.g., ``a man'') and the bottom panel shows examples with a female subject. The region to the far left (below $-1$) shows that the second principal component encodes apparent gender contradictions: `man' partnered with a female-dominated profession (top) or `woman' partnered with a male-dominated profession (bottom). On the right, examples marked with red points in the left panels are shown, along with their BLS percentages in parentheses,
and predicted similarities from the predictors with the most negative and positive loadings in the first principal component.}
    \label{fig:bls_pca}
\end{figure}

To better understand the differences between
predictors in our ensemble,
we analyze the structure in how similarity scores produced by the predictors in our ensemble deviate from the ensemble mean.
Here, we find that the main axis of variation aligns, at least at its extremes, with differences in how predictors represent sterotypical associations between profession and gender.  
Specifically, we perform principal components analysis (PCA) over similarity score produced by 20 fine-tunings of a single BERT checkpoint.
We plot the first principal components, which contains 22\% of the variation in score deviations, against BLS female participation percentages in Figure~\ref{fig:bls_pca}.
Notably, examples in the region where the first principal component values are strongly negative include some of the strongest gender imbalances.
The right side of Figure~\ref{fig:bls_pca} shows some of these examples (marked in red on the scatterplots), along with the predicted similarities from models that have strongly negative or strongly positive loadings on this principal axis.
The similarity scores between these models are clearly divergent, with the positive-loading models encoding a sterotypical contradiction between gender and profession---that is, a contradiction between `man' and `receptionist' or `nurse'; or a contradiction between `woman' and `mechanic', `carpenter', and `doctor'---that the negative-loading models do not.

\begin{figure}
    \centering
    \includegraphics[width=.9\textwidth]{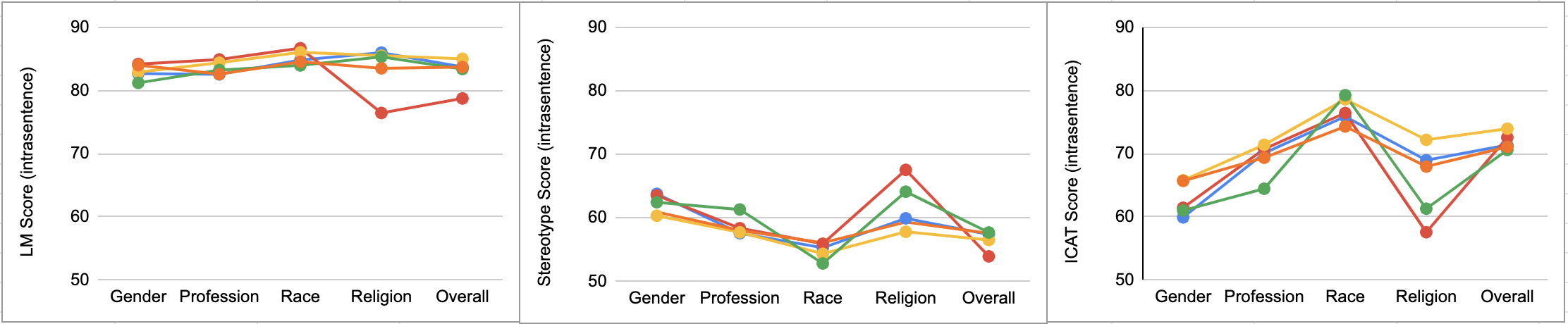}
    \caption{\textbf{Different pretraining seeds produce different steretypical associations.}
    Results across five identically trained BERT Large (Cased) pretraining checkpoints on StereoSet~\citep{nadeem2020stereoset}.
    The ICAT score combines a language model (LM) score measuring ``sensibility'' and a stereotype score measuring correlations of language model predictions with known stereotypes. A leaderboard featuring canonical pretrainings is available at \url{https://stereoset.mit.edu/}.}
    \label{fig:nlp-stereoset}
\end{figure}

\subsection{Stereotypical Associations in Pretrained Language Models}

Underspecification in supervised NLP systems can occur at both the fine-tuning and pretraining stages.
In the previous section, we gave suggestive evidence that underspecification allows identically pretrained BERT checkpoints to encode substantively different inductive biases.
Here, we examine pretraining underspecification more directly, considering again its impact on reliance on stereotypical shortcuts.
Specifically, we examine the performance of our ensemble of five BERT checkpoints on the StereoSet benchmark~\citep{nadeem2020stereoset}.

StereoSet is a set of stress tests designed to directly assess how the predictions of pretrained language models correlate with well-known social stereotypes.
Specifically, the test inputs are spans of text with sentences or words masked out, and the task is to score a set of choices for the missing piece of text.
The choice set contains one non-sensical option, and two sensical options, one of which conforms to a stereotype, and the other of which does not.
The benchmark probes stereotypes along the axes of gender, profession, race, and religion.
Models are scored based on both whether they are able to exclude the non-sensical option (LM Score) and whether they consistently choose the option that conforms with the stereotype (Stereotype Score).
These scores are averaged together to produce an Idealized Context Association Test (ICAT) score, which can be applied to any language model.

In \autoref{fig:nlp-stereoset}, we show the results of evaluating our five BERT checkpoints, which differ only in random seed, across all StereoSet metrics.
The variation across checkpoints is large. The range of overall ICAT score between the our identical checkpoints is 3.35. For context, this range is larger than the gap between the top six models on the public leaderboard,\footnote{\url{https://stereoset.mit.edu} retrieved October 28, 2020.}, which differ in size, architecture, and training data (GPT-2 (small), XLNet (large), GPT-2 (medium), BERT (base), GPT-2 (large), BERT (large)). 
On the disaggregated metrics, the score range between checkpoints is narrower on the LM score (sensible vs. non-sensible sentence completions) than on the Stereotype score (consistent vs. inconsistent with social stereotypes).
This is consistent with underspecification, as the LM score is more closely aligned to the training task.
Interestingly, score ranges are also lower on overall metrics compared to by-demographic metrics, suggesting that even when model performance looks stable in aggregate, checkpoints can encode different social stereotypes.

\subsection{Spurious Correlations in Natural Language Inference}
Underspecification also affects more general inductive biases that align with some notions of ``semantic understanding'' in NLP systems.
One task that probes such notions is natural language inference (NLI).
The NLI task is to classify sentence pairs (called the \textbf{premise} and \textbf{hypothesis}) into one of the following semantic relations: entailment (the hypothesis is true whenever the premise is), contradiction (the hypothesis is false when the premise is true), and neutral~\citep{bowman-etal-2015-large}.
Typically, language models are fine-tuned for this task on labeled datasets such as the MultiNLI training set~\citep{williams2018broad}.
While test set performance on benchmark NLI datasets approaches human agreement~\citep{wang2018glue}, it has been shown that there are shortcuts to achieving high performance on many NLI datasets~\citep{mccoy2019right,zellers2018swag,zellers2019hellaswag}.
In particular, on stress tests that are designed to probe semantic inductive biases more directly these models are still far below human performance.

Notably, previous work has shown that performance on these stronger stress tests has also been shown to be unstable with respect to the fine-tuning seed~\citep{zhou2020curse,mccoy2019berts,dodge2020fine}.
We interpret this to be a symptom of underspecification. 
Here, we replicate and extend this prior work by assessing sensitivity to both fine-tuning and, for the first time, pretraining.
Here we use the same five pre-trained BERT Large cased checkpoints, and fine-tune each on the MultiNLI training set~\citep{williams2018broad} 20 times.
Across all pre-trainings and fine-tunings, accuracy on the standard MNLI matched and unmatched test sets are in tightly constrained ranges of $(83.4\%-84.4\%)$ and $(83.8\%-84.7\%)$, respectively.\footnote{The ``matched'' and ``unmatched'' conditions refer to whether the test data is drawn from the same genre of text as the training set.}

We evaluate our ensemble of predictors on the HANS stress test \citep{mccoy2019right} and the StressTest suite from \citet{naik2018stress}.
The HANS Stress Tests are constructed by identifying spurious correlations in the training data --- for example, that entailed pairs tend to have high lexical overlap --- and then generating a test set such that the spurious correlations no longer hold. The \citet{naik2018stress} stress tests are constructed by perturbing examples, for example by introducing spelling errors or meaningless expressions (``and true is true''). 

We again find strong evidence that the extent to which a trained model relies on shortcuts is underspecified, as demonstrated by sensitivity to the choice of random seed at both fine-tuning and pre-training time. Here, we report several broad trends of variation on these stress tests: first, the magnitude of the variation is large; second, the variation is also sensitive to the fine-tuning seed, replicating \citet{zhou2020curse};
third, the variation is also sensitive to the pre-training seed; 
fourth, the variation is difficult to predict based on performance on the standard MNLI validation sets; and finally, the variation on different stress tests tends to be weakly correlated. 

\begin{figure}
    \centering
    \includegraphics[width=0.95\textwidth]{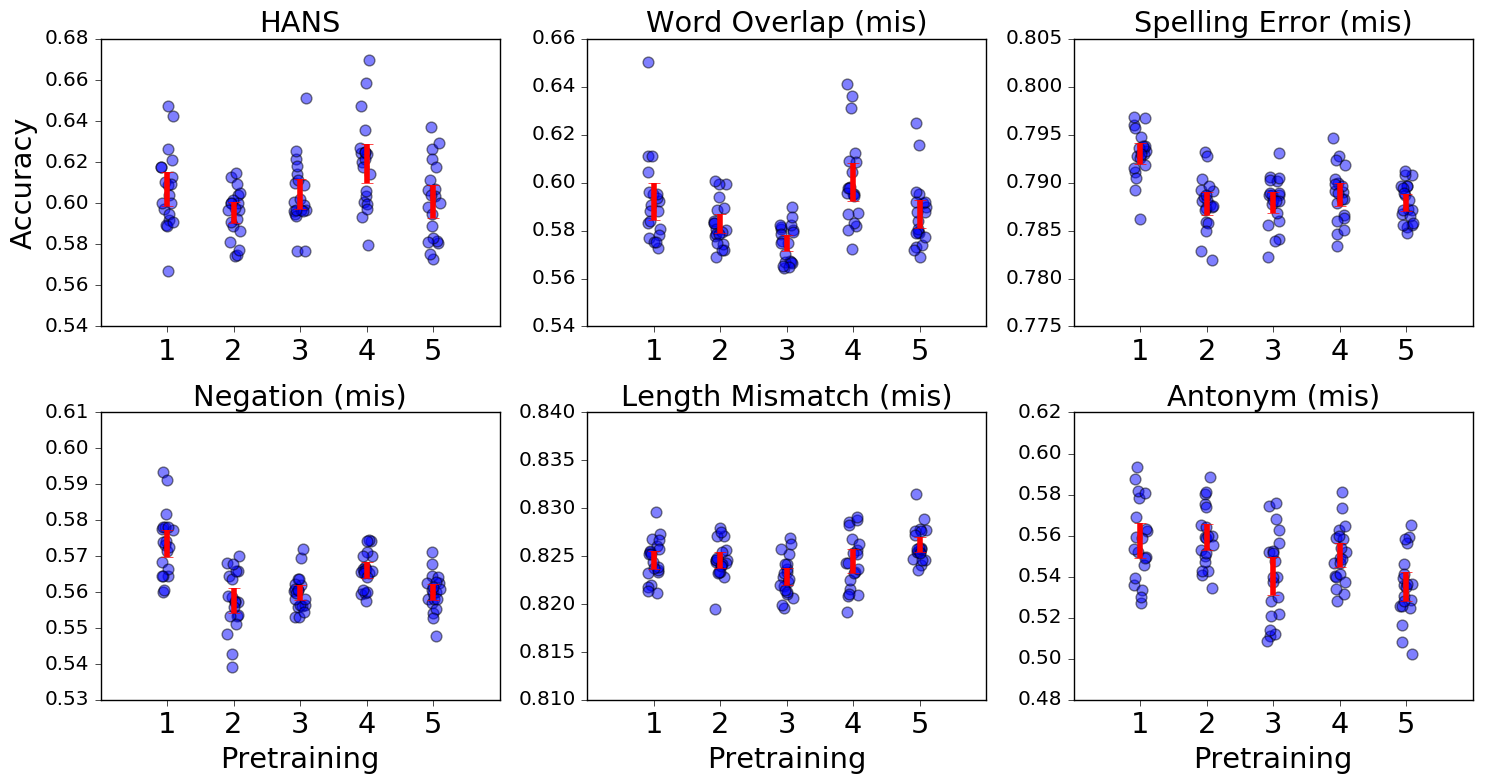}\\
    \vspace{10pt}
    \includegraphics[width=0.95\textwidth]{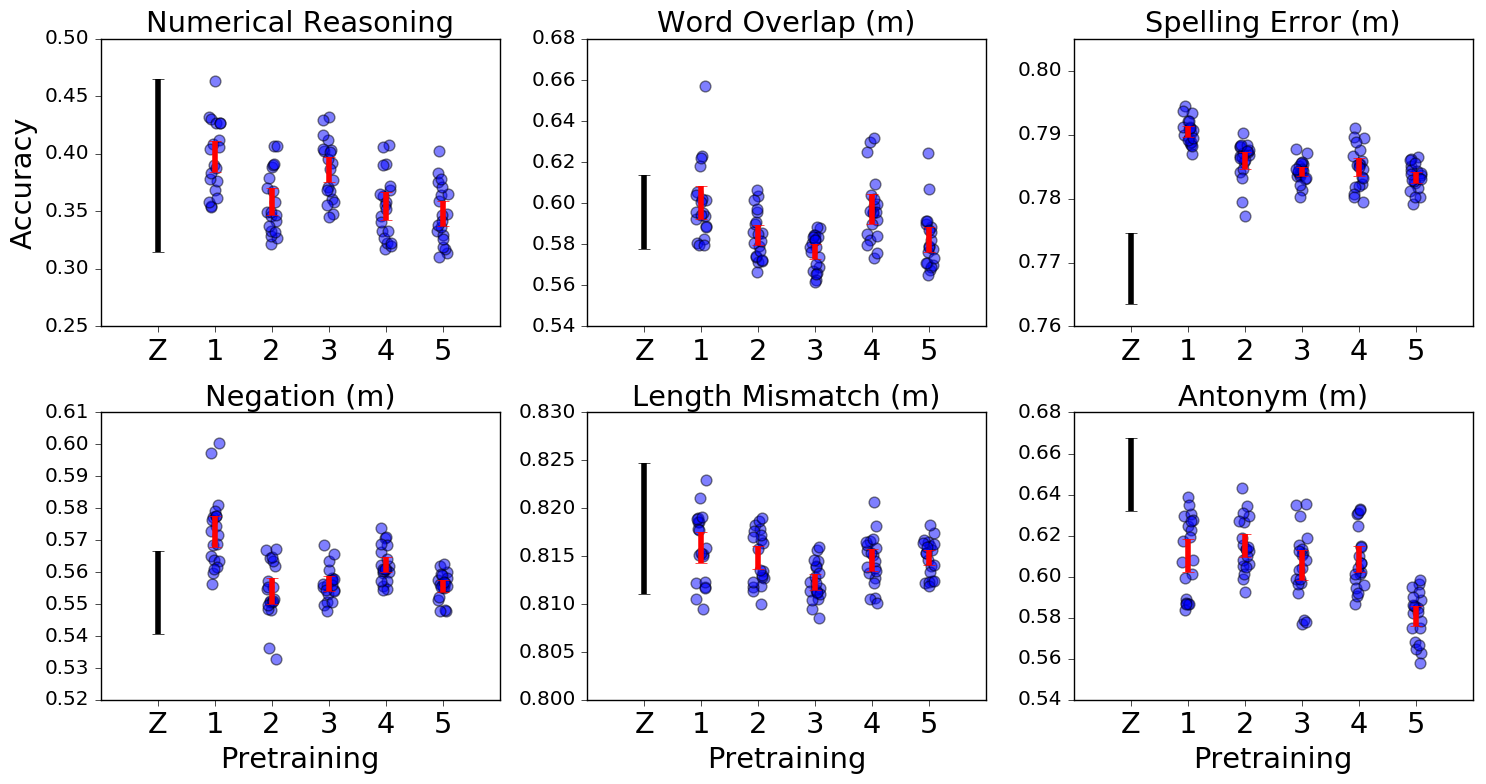}
    
    \caption{\textbf{Predictor performance on NLI stress tests varies both within and between pre-training checkpoints.} 
    Each point corresponds to a fine-tuning of a pre-trained BERT checkpoint on the MNLI training set, with pre-training distinguished on the $x$-axis.
    All pre-trainings and fine-tunings differ only in random seed at their respective training stages.
    Performance on HANS \citep{mccoy2019right} is shown in the top left; remaining results are from the StressTest suite \citep{naik2018stress}.
    Red bars show a 95\% CI around for the mean accuracy within each pre-training.
    The tests in the bottom group of panels were also explored in \citet{zhou2020curse} across fine-tunings from the public BERT large cased checkpoint \citep{devlin2019bert}; for these, we also plot the mean +/- 1.96 standard deviations interval, using values reported in \citet{zhou2020curse}.
    The magnitude of variation is substantially larger on most stress tests than the MNLI test sets ($< 1\%$ on both MNLI matched and unmatched). 
    There is also substantial variation between some pretrained checkpoints, even after fine-tuning.}
    \label{fig:nlp-nli-stress}
\end{figure}

\begin{figure}
    \centering
    \includegraphics[width=0.7\textwidth]{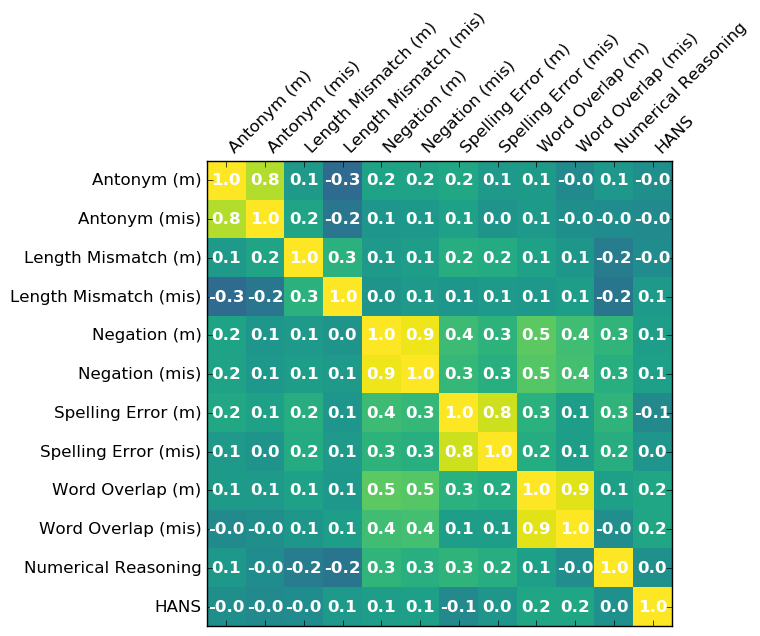}
    \caption{\textbf{Predictor performance across stress tests are typically weakly correlated.}
    Spearman correlation coefficients of 100 predictor accuracies from 20 fine-tunings of five pretrained BERT checkpoints.}
    \label{fig:nlp_nli_stress_cor}
\end{figure}

\autoref{fig:nlp-nli-stress} shows our full set of results, broken down by pre-training seed.
These plots show evidence of the influence of the pre-training seed; for many tests, there appear to be systematic differences in performance from fine-tunings based on checkpoints that were pre-trained with different seeds.
We report one numerical measurement of these differences with $F$ statistics in \autoref{tab:nlp-nli-stress}, where the ratio of between-group variance to within-group variance is generally quite large.
\autoref{tab:nlp-nli-stress} also reports Spearman rank correlations between stress test accuracies and accuracy on the MNLI matched validation set.
The rank correlation is typically small, suggesting that the variation in stress test accuracy is largely orthogonal to validation set accuracy enforced by the training pipeline.
Finally, in \autoref{fig:nlp_nli_stress_cor}, we show that the correlation between stress tests performance is also typically small (with the exception of some pairs of stress tests meant to test the same inductive bias), suggesting that the space of underspecified inductive biases spans many dimensions.

\begin{table}[]
    \centering
\begin{tabular}{lrr}
\toprule
Dataset                      & $F$ (p-value)   & Spearman $\rho$ (95\% CI)\\
\midrule
MNLI, matched                &  1.71 (2E-01) &   --- \\
MNLI, mismatched             & 20.18 (5E-12) &  0.11 (-0.10, 0.31) \\
\midrule
\textbf{\citet{naik2018stress} stress tests}\\
Antonym, matched             & 15.46 (9E-10) &  0.05 (-0.16, 0.26) \\
Antonym, mismatched          &  7.32 (4E-05) &  0.01 (-0.20, 0.21) \\
Length Mismatch, matched     &  4.83 (1E-03) &  0.33 ( 0.13, 0.50) \\
Length Mismatch, mismatched  &  5.61 (4E-04) & -0.03 (-0.24, 0.18) \\
Negation, matched            & 19.62 (8E-12) &  0.17 (-0.04, 0.36) \\
Negation, mismatched         & 18.21 (4E-11) &  0.09 (-0.12, 0.29) \\
Spelling Error, matched      & 25.11 (3E-14) &  0.40 ( 0.21, 0.56) \\
Spelling Error, mismatched   & 14.65 (2E-09) &  0.43 ( 0.24, 0.58) \\
Word Overlap, matched        &  9.99 (9E-07) &  0.08 (-0.13, 0.28) \\
Word Overlap, mismatched     &  9.13 (3E-06) & -0.07 (-0.27, 0.14) \\
Numerical Reasoning          & 12.02 (6E-08) &  0.18 (-0.03, 0.38) \\
\midrule
HANS~\citep{mccoy2019right}                         &  4.95 (1E-03) &  0.07 (-0.14, 0.27) \\
\bottomrule
\end{tabular}
    \caption{\textbf{Summary statistics for structure of variation in predictor accuracy across NLI stress tests.}
    For each dataset, we measure the accuracy of 100 predictors, corresponding to 20 randomly initialized fine-tunings from 5 randomly initialized pretrained BERT checkpoints.
    All models are fine-tuned on the MNLI training set, and validated on the MNLI matched test set~\citep{williams2018broad}.
    The $F$ statistic quantifies how systematic differences are between pretrainings.
    Specifically, it is the ratio of within-pretraining variance to between-pretraining variance in the accuracy statistics.
    $p$-values are reported to give a sense of scale, but not for inferential purposes; it is unlikely that assumptions for a valid $F$-test are met.
    The Spearman $\rho$ statistic quantifies how ranked performance on the MNLI matched test set correlates with ranked performance on each stress test.
    For most stress tests, there is only a weak relationship, such that choosing models based on test performance alone would not yield the best models on stress test performance.}
    \label{tab:nlp-nli-stress}
\end{table}

\subsection{Conclusions}
There is increasing concern about whether natural language processing systems are learning general linguistic principles, or whether they are simply learning to use surface-level shortcuts~\citep[e.g.,][]{bender-koller-2020-climbing,linzen-2020-accelerate}. Particularly worrying are shortcuts that reinforce societal biases around protected attributes such as gender~\citep[e.g.,][]{webster2020measuring}. The results in this section replicate prior findings that highly-parametrized NLP models do learn spurious correlations and shortcuts. However, this reliance is underspecified by the model architecture, learning algorithm, and training data: merely changing the random seed can induce large variation in the extent to which spurious correlations are learned. Furthermore, this variation is demonstrated in both pretraining and fine-tuning, indicating that pretraining alone can ``bake in'' more or less robustness. This implies that individual stress test results should be viewed as statements about individual model checkpoints, and not about architectures or learning algorithms. More general comparisons require the evaluation of multiple random seeds.
\section{Case Study in Clinical Predictions from Electronic Health Records}
\label{sec:ehr}

The rise of Electronic Health Record (EHR) systems has created an opportunity for building predictive ML models for diagnosis and prognosis (e.g. \cite{Ambrosino1995, Brisimi2019, Feng2019}).
In this section, we focus on one such model that uses a Recurrent Neural Network (RNN) architecture with EHR data to predict acute kidney injury (AKI) during hospital admissions \citep{Tomasev2019}. AKI is a common complication in hospitalized patients and is associated with increased morbidity, mortality, and healthcare costs \citep{Khwaja2012}.
Early intervention can improve outcomes in AKI~\citep{nice2019}, which has driven efforts to predict it in advance using machine learning. \cite{Tomasev2019} achieve state-of-the-art performance, detecting the onset of AKI up to 48 hours in advance with an accuracy of 55.8\% across all episodes and 90.2\% for episodes associated with dialysis administration. 

Despite this strong discriminative performance, there have been questions raised about 
the associations being learned by this model and whether they conform with our understanding of physiology\citep{kellum2019artificial}.
Specifically, for some applications, it is desirable to disentangle physiological signals from operational factors related to the delivery of healthcare, both of which appear in EHR data.
As an example, the value of a lab test may be considered a physiological signal; however the timing of that same test may be considered an operational one (e.g. due to staffing constraints during the night or timing of ward rounds).
Given the fact that operational signals may be institution-specific and are likely to change over time, understanding to what extent a model relies on different signals can help practitioners determine whether the model meets their specific generalization requirements \citep{futoma2020myth}. 

Here, we show that underspecification makes the answer to this question ambiguous. 
Specifically, we apply our experimental protocol to the \citet{Tomasev2019} AKI model which predicts the continuous risk (every 6 hours) of AKI in a 48h lookahead time window (see Supplement for details).

\subsection{Data, Predictor Ensemble, and Metrics}

The pipeline and data used in this study are described in detail in \cite{Tomasev2019}. Briefly, the data consists of de-identified EHRs from 703,782 patients across multiple sites in the United States collected at the US Department of Veterans Affairs\footnote{Disclaimer: Please note that the views presented in this manuscript are that of the authors and not that of the Department of the Veterans Affairs.} between 2011 and 2015. Records include structured data elements such as medications, labs, vital signs, diagnosis codes etc, aggregated in six hour time buckets (time of day 1: 12am-6am, 2: 6am-12pm, 3: 12pm-6pm, 4: 6pm-12am).
In addition, precautions beyond standard de-identification have been taken to safeguard patient privacy: free text notes and rare diagnoses have been excluded; many feature names have been obfuscated; feature values have been jittered; and all patient records are time-shifted, respecting relative temporal relationships for individual patients. Therefore, this dataset is only intended for methodological exploration.

The model consists of embedding layers followed by a 3 layer-stacked RNN before a final dense layer for prediction of AKI across multiple time horizons.
Our analyses focus on predictions with a 48h lookahead horizon, which have been showcased in the original work for their clinical actionability. 
To examine underspecification, we construct a model ensemble by training the model from 5 random seeds for each of three RNN cell types: Simple Recursive Units (SRU, \cite{Lei2018}), Long Short-Term Memory (LSTM, \cite{Hochreiter1997}) or Update Gate RNN (UGRNN, \cite{Collins2017}).
This yields an ensemble of 15 model instances in total.

The primary metric that we use to evaluate predictive performance is normalized area under the precision-recall curve (PRAUC) \citep{Boyd2012}, evaluated across all patient-timepoints where the model makes a prediction.
This is a PRAUC metric that is normalized for prevalence of the positive label (in this case, AKI events).
Our ensemble of predictors achieves tightly constrained normalized PRAUC values between 34.59 and 36.61.

\subsection{Reliance on Operational Signals}
We evaluate these predictors on stress tests designed to probe the sensitivity to specific operational signals in the data: the timing and number of labs recorded in the EHR\footnote{Neither of these factors are purely operational---there is known variation in kidney function across the day and the values of accompanying lab tests carry valuable information about patient physiology. However, we use these here as approximations for an operational perturbation.}.
In this dataset, the prevalence of AKI is largely the same across different times of day (see Table 1 of Supplement).
However, AKI is diagnosed based on lab tests,
\footnote{specifically a comparison of past and current values of creatinine \citep{Khwaja2012}}, and there are clear temporal patterns in how tests are ordered.
For most patients, creatinine is measured in the morning as part of a `routine', comprehensive panel of lab tests.
Meanwhile, patients requiring closer monitoring may have creatinine samples taken at additional times, often ordered as part of an `acute', limited panel (usually, the basic metabolic panel\footnote{This panel samples Creatinine, Sodium, Potassium, Urea Nitrogen, CO2, Chloride and Glucose.}).
Thus, both the time of day that a test is ordered, and the panel of tests that accompany a given measurement may be considered primarily as operational factors correlated with AKI risk.

We test for reliance on this signal by applying two interventions to the test data that modify (1) the time of day of all features (aggregated in 6h buckets) and (2) the selection of lab tests.
The first intervention shifts the patient timeseries by a fixed offset, while the second intervention additionally removes all blood tests that are not directly relevant to the diagnosis of AKI.
We hypothesize that if the predictor encodes physiological signals rather than these operational cues, the predictions would be invariant to these interventions.
More importantly, if the model's reliance on these operational signals is underspecified, we would expect the behavior of the predictors in our ensemble to respond differently to these modified inputs.

\begin{figure}[!ht]
    \centering
    \includegraphics{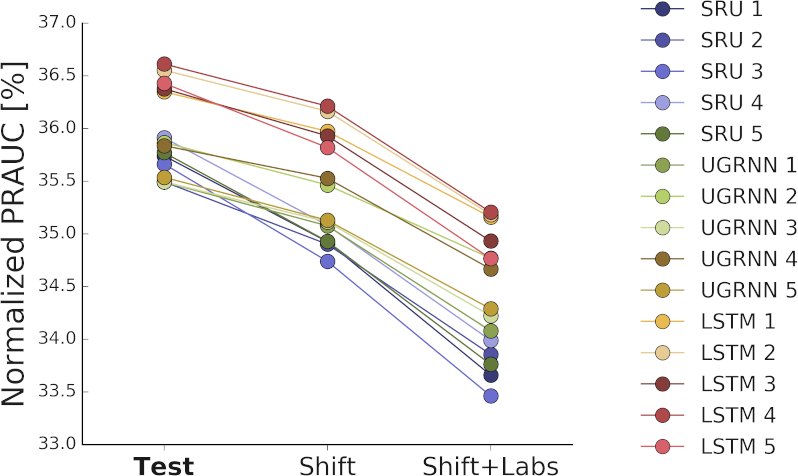}
    \caption{\textbf{Variability in performance from ensemble of RNN models processing electronic health records (EHR)}. Model sensitivity to time of day and lab perturbations.
    The x-axis denotes the evaluation set: ``Test'' is the original test set; ``Shift'' is the test set with time shifts applied; ``Shift + Labs'' applies the time shift and subsets lab orders to only include the basic metabolic panel CHEM-7.
    The y-axis represents the normalized PRAUC, and each set of dots joined by a line represents a model instance.}
    \label{fig:ehr_shift_pop}
\end{figure}

We begin by examining overall performance on this shifted test set across our ensemble.
In Figure~\ref{fig:ehr_shift_pop}, we show that performance on the intervened data is both worse and more widely dispersed than in the standard test set, especially when both interventions are applied.
This shows that the model incorporates time of day and lab content signals, and that the extent to which it relies on these signals is sensitive to both the recurrent unit and random initialization.

\begin{figure}[t]
    \centering
    \includegraphics[width=\textwidth]{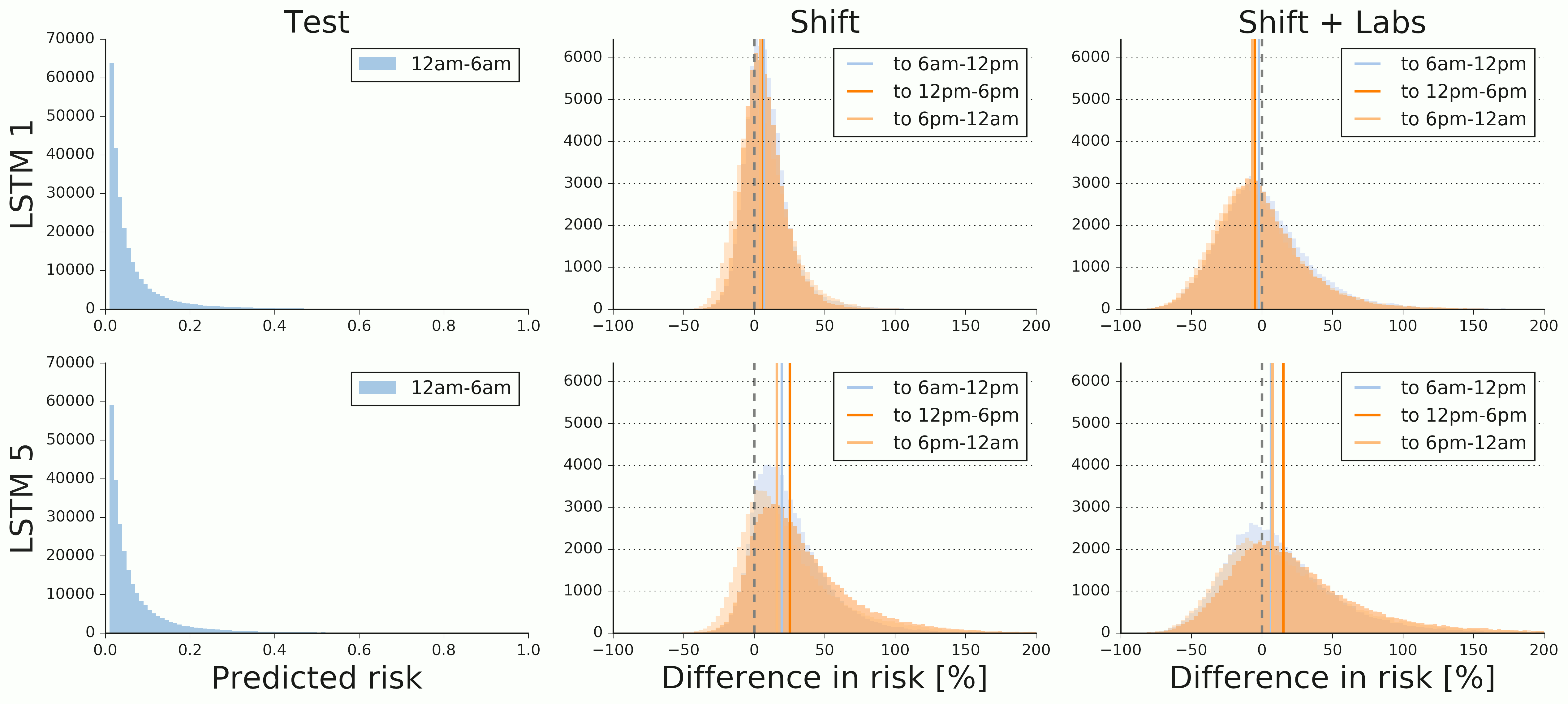}
    \caption{\textbf{Variability in AKI risk predictions between two LSTM models processing electronic health records (EHR)}. Histograms showing showing risk predictions from two models, and changes induced by time of day and lab perturbations.
    Histograms show counts of patient-timepoints where creatinine measurements were taken in the early morning (12am-6am). LSTM 1 and 5 differ only in random seed.
    ``Test'' shows histogram of risk predicted in original test data.
    ``Shift'' and ``Shift + Labs'' show histograms of proportional changes (in \%) $\frac{\text{Perturbed} - \text{Baseline}}{\text{Baseline}}$ induced by the time-shift perturbation and the combined time-shift and lab perturbation, respectively.}
    
    \label{fig:ehr_shift_hist}
\end{figure}

The variation in performance reflects systematically different inductive biases encoded by the predictors in the ensemble. 
We examine this directly by measuring how individual model predictions change under the timeshift and lab intervensions.
Here, we focus on two trained LSTM models that differ only in their random seeds, and examine patient-timepoints at which creatinine measurements were taken.
In Figure~\ref{fig:ehr_shift_hist}\textbf{(Right)}, we show distributions of predicted risk on the original patient-timepoints observed in the ``early morning'' (12am-6am) time range, and proportional changes to these risks when the timeshift and lab interventions were applied.
Both predictors exhibit substantial changes in predicted risk under both interventions, but the second predictor is far more sensitive to these changes than the first, with the predicted risks taking on substantially different distributions depending on the time range to which the observation is shifted.

\begin{figure}[t]
    \centering
    \includegraphics[width=0.85\textwidth]{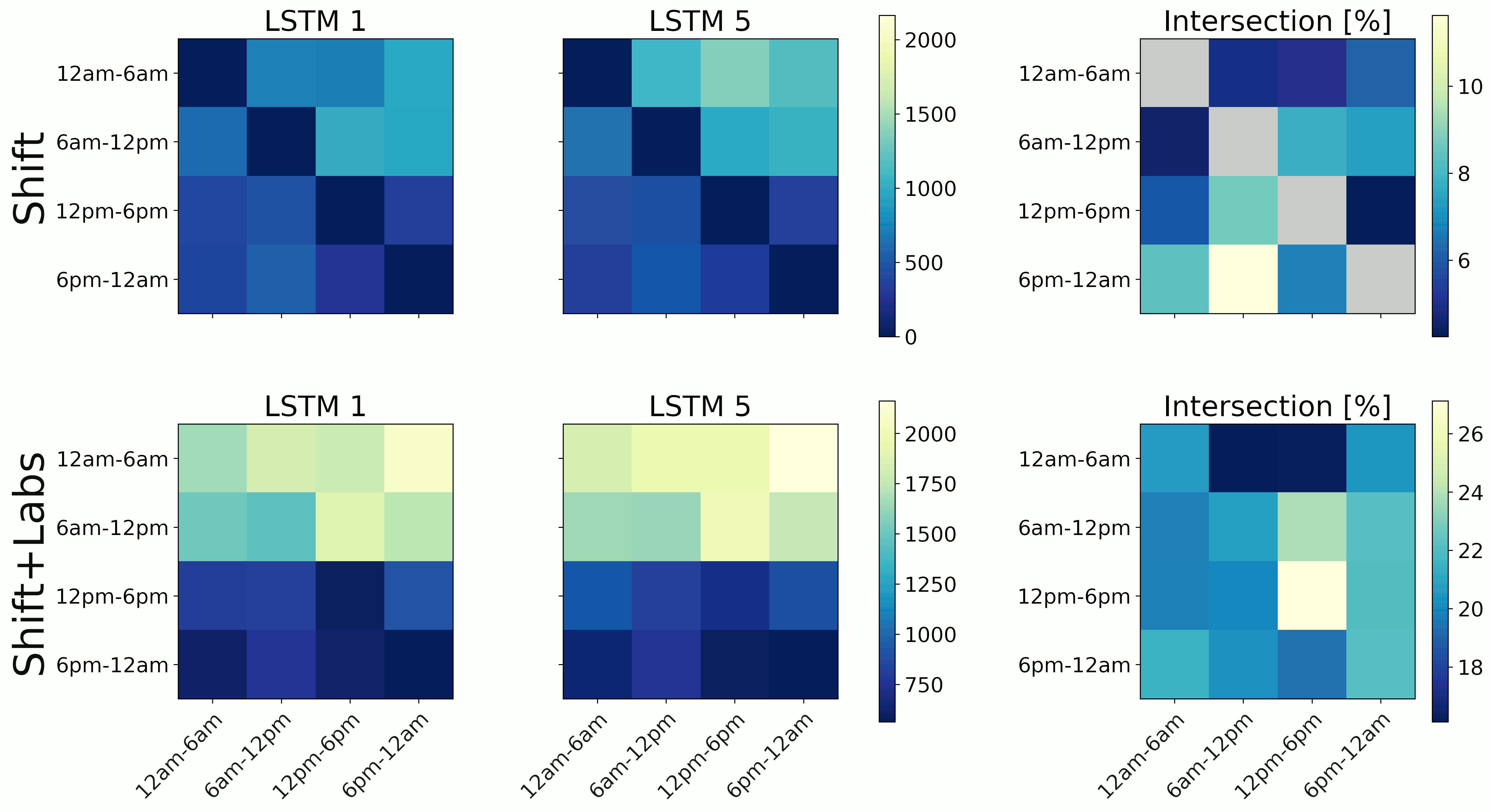}
    \caption{\textbf{Variability in AKI predictions between two LSTM models processing electronic health records (EHR)}.
    Counts (color-coded) of decisions being flipped due to the stress tests, from the LSTM 1 and LSTM 5 models, as well as the proportions of those flipped decision intersecting between the two models (in \%). Rows represent the time of day in the original test set, while columns represent the time of day these samples were shifted to. LSTM 1 and 5 differ only in random seed. ``Shift'' represents the flipped decisions (both positive to negative and negative to positive) between the predictions on the test set and the predictions after time-shift perturbation. ``Shift + Labs'' represents the same information for the combined time-shift and labs perturbation.}
    
    \label{fig:ehr_shift_flipped}
\end{figure}

These shifts in risk are consequential for decision-making and can result in AKI episodes being predicted tardily or missed.
In Figure~\ref{fig:ehr_shift_flipped}, we illustrate the number of patient-timepoints where the changed risk score crosses each model's calibrated decision threshold.
In addition to substantial differences in the number of flipped decisions, we also show that most of these flipped decisions occur at different patient-timepoints across models.

\subsection{Conclusions}
Our results here suggest that predictors produced by this model tend to rely on the pattern of lab orders in a substantial way, but the extent of this reliance is underspecified. 
Depending on how stable this signal is in the deployment context, this may or may not present challenges.
However, this result also shows that the reliance on this signal is not \emph{enforced} by the model specification of training data, suggesting that the reliance on lab ordering patterns could be modulated by simply adding constraints to the training procedure, and without sacrificing iid performance.
In the Supplement, we show one such preliminary result, where a model trained with the timestamp feature completely ablated was able to achieve identical iid predictive performance.
This is compatible with previous findings that inputting medical/domain relational knowledge has led to better out of domain behaviour \cite{Nestor2019}, performance \cite{Popescu2011, Choi2017, Tomasev2019, Tomasev2019a} and interpretability \cite{Panigutti2020} of ML models.

\section{Discussion: Implications for ML Practice}
\label{sec:discussion}

Our results show that underspecification is a key failure mode for machine leaning models to encode generalizable inductive biases.
We have used between-predictor variation in stress test performance as an observable signature of underspecification.
This failure mode is distinct from generalization failures due to structural mismatch between training and deployment domains.
We have seen that underspecification is ubiquitious in practical machine learning pipelines across many domains.
Indeed, thanks to underspecification, substantively important aspects of the decisions are determined by arbitrary choices such as the random seed used for parameter initialization.
We close with a discussion of some of the implications of the study, which broadly suggest a need to find better interfaces for domain knowledge in ML pipelines.

First, we note that the methodology in this study underestimates the impact of underspecification: our goal was to detect rather than fully characterize underspecification, and in most examples, we only explored underspecification through the subtle variation that can result from modifying random seeds in training.
However, modern deep learning pipelines incorporate a wide variety of \emph{ad hoc} practices, each of which may carry its own ``implicit regularization'', which in turn can translate into substantive inductive biases about how different features contribute to the behavior of predictors.
These include the particular scheme used for initialization; conventions for parameterization; choice of optimization algorithm; conventions for representing data; and choices of batch size, learning rate, and other hyperparameters, all of which may interact with the infrastructure available for training and serving models \citep{hooker2020hardware}. 
We conjecture that many combinations of these choices would reveal a far larger risk-preserving set of predictors $\mathcal F^*$, a conjecture that has been partially borne out by concurrent work~\citep{wenzel2020hyperparameter}.
However, we believe that there would be value in more systematically mapping out the set of iid-equivalent predictors that a pipeline could return as a true measurement of the uncertainty entailed by underspecification.
Current efforts to design more effective methods for exploring loss landscapes \citep{fort2019deep,garipov2018loss} could play an important role here, and there are opportunities to import ideas from the sensitivity analysis and partial identification subfields in causal inference and inverse problems.

Second, our findings underscore the need to thoroughly test models on application-specific tasks, and in particular to check that the performance on these tasks is stable.
The extreme complexity of modern ML models ensures that some aspect of the model will almost certainly be underspecified; thus, the challenge is to ensure that this underspecification does not jeopardize the inductive biases that are required by an application.
In this vein, designing stress tests that are well matched to applied requirements, and that provide good ``coverage" of potential failure modes is a major challenge that requires incorporating domain knowledge.
This can be particularly challenging, given our results show that there is often low correlation between performance on distinct stress tests when iid performance is held constant, and the fact that many applications will have fine-grained requirements that require more customized stress testing.
For example, within the medical risk prediction domain, the dimensions that a model is required to generalize across (e.g., temporal, demographic, operational, etc.)  will depend on the details of the deployment  and the goals of the practitioners \citep{futoma2020myth}.
For this reason, developing best practices for \emph{building} stress tests that crisply represent requirements, rather than standardizing on specific benchmarks, may be an effective approach.
This approach has gained traction in the NLP subfiled, where several papers now discuss the process by which stress tests datasets should iterate continuously \citep{zellers2019hellaswag}, and new systems for developing customized stress tests have been proposed \citep{ribeiro2020beyond,Kaushik2020Learning}.

Third, our results suggest some ways forward for training models with credible inductive biases.
By definition, underspecification can be resolved by specifying additional criteria for selecting predictors from the equivalence class of near-optimal predictors $\mathcal F^*$. 
Importantly, this suggests a departure from a popular strategy of improving iid performance of models by marginalizing across $\mathcal F^*$ \citep{wilson2020bayesian}.
Here, because it is known that some predictors in $\mathcal F^*$ generalize poorly in new domains, simply averaging them together is not guaranteed to produce better results on stress tests than carefully choosing a specific predictor from the equivalence class (see Appendix~\ref{sec:vision appendix} for some examples).
Of course, these approaches can be reconciled if the marginalization is restricted to models that satisfy required constraints.
The challenge of specifying selection criteria or constraints on $\mathcal F^*$, however, remains an active area of research.
Because they are meant to enforce application-specific requirements, such criteria or constraints must also be application-specific, presenting a challenge for the development of general methods.
Although some general-purpose heuristics have been proposed \citep{bengio2017consciousness}, proposals for expressing application-specific requirements with flexible but unified frameworks may be a promising middle ground solution.
Causal DAGs \citep{scholkopf2019causality} and explanations \citep{ross2017right} are both promising candidates for such frameworks.

Finally, when the main hurdle is underspecification, adding constraints should not result in a tradeoff between learning better inductive biases and generalizing well in iid settings.
Some recent work in several places comports with this conjecture: in the robustness literature \citet{raghunathan2020understanding} show that robustness / accuracy tradeoffs need not be fundamental; in the NLP literature, \citet{webster2020measuring} show that reliance on gendered correlations can be reduced in BERT-derived models little-to-no tradeoff \citep{webster2020measuring}; and in Appendix~\ref{sec:ehr appendix}, we show a similar preliminary result from our EHR example.
These results suggest that designing application-specific regularization schemes (e.g., that bias the pipeline toward predictors that approximately respect causal structure) may be a promising direction for incorporating domain expertise without compromising the powerful prediction abilities of modern ML models.

\section*{Acknowledgements}

This research has been conducted using the UK Biobank Resource under Application Number 17643.
We would also like to thank our partners - EyePACS in the United States and Aravind Eye Hospital and Sankara Nethralaya in India for providing the datasets used to train the models for predicting diabetic retinopathy from fundus images.
We also appreciate the advice of our DeepMind collaborator Dr. Nenad Tomasev, Prof. Finale Doshi-Velez and the wider Google Health Research UK team led by Dr. Alan Karthikesalingam.

\bibliographystyle{alpha}
\bibliography{refs,theory_refs,supp_sections/supp_refs}

\begin{thebibliography}{139}
\providecommand{\natexlab}[1]{#1}
\providecommand{\url}[1]{\texttt{#1}}
\expandafter\ifx\csname urlstyle\endcsname\relax
  \providecommand{\doi}[1]{doi: #1}\else
  \providecommand{\doi}{doi: \begingroup \urlstyle{rm}\Url}\fi

\bibitem[Adamson and Smith(2018)]{adamson2018machine}
Adewole~S Adamson and Avery Smith.
\newblock Machine learning and health care disparities in dermatology.
\newblock \emph{JAMA dermatology}, 154\penalty0 (11):\penalty0 1247--1248,
  2018.

\bibitem[Adelekun et~al.(2020)Adelekun, Onyekaba, and Lipoff]{adelekun2020skin}
Ademide Adelekun, Ginikanwa Onyekaba, and Jules~B Lipoff.
\newblock Skin color in dermatology textbooks: An updated evaluation and
  analysis.
\newblock \emph{Journal of the American Academy of Dermatology}, 2020.

\bibitem[Ambrosino et~al.(1995)Ambrosino, Buchanan, Cooper, and
  Fine]{Ambrosino1995}
R~Ambrosino, B~G Buchanan, G~F Cooper, and M~J Fine.
\newblock {The use of misclassification costs to learn rule-based decision
  support models for cost-effective hospital admission strategies.}
\newblock \emph{Proceedings. Symposium on Computer Applications in Medical
  Care}, pages 304--8, 1995.
\newblock ISSN 0195-4210.
\newblock URL \url{http://www.ncbi.nlm.nih.gov/pubmed/8563290
  http://www.pubmedcentral.nih.gov/articlerender.fcgi?artid=PMC2579104}.

\bibitem[Arjovsky et~al.(2019)Arjovsky, Bottou, Gulrajani, and
  Lopez-Paz]{arjovsky2019invariant}
Martin Arjovsky, L{\'e}on Bottou, Ishaan Gulrajani, and David Lopez-Paz.
\newblock Invariant risk minimization.
\newblock \emph{arXiv preprint arXiv:1907.02893}, 2019.

\bibitem[Athey(2017)]{athey2017beyond}
Susan Athey.
\newblock Beyond prediction: Using big data for policy problems.
\newblock \emph{Science}, 355\penalty0 (6324):\penalty0 483--485, 2017.

\bibitem[Babaeianjelodar et~al.(2020)Babaeianjelodar, Lorenz, Gordon, Matthews,
  and Freitag]{babaeianjelodar2020quantifying}
Marzieh Babaeianjelodar, Stephen Lorenz, Josh Gordon, Jeanna Matthews, and Evan
  Freitag.
\newblock Quantifying gender bias in different corpora.
\newblock In \emph{Companion Proceedings of the Web Conference 2020}, pages
  752--759, 2020.

\bibitem[Barbu et~al.(2019)Barbu, Mayo, Alverio, Luo, Wang, Gutfreund,
  Tenenbaum, and Katz]{barbu2019objectnet}
Andrei Barbu, David Mayo, Julian Alverio, William Luo, Christopher Wang, Dan
  Gutfreund, Josh Tenenbaum, and Boris Katz.
\newblock Objectnet: A large-scale bias-controlled dataset for pushing the
  limits of object recognition models.
\newblock In \emph{Advances in Neural Information Processing Systems}, pages
  9448--9458, 2019.

\bibitem[Beede et~al.(2020)Beede, Baylor, Hersch, Iurchenko, Wilcox,
  Ruamviboonsuk, and Vardoulakis]{beede2020human}
Emma Beede, Elizabeth Baylor, Fred Hersch, Anna Iurchenko, Lauren Wilcox,
  Paisan Ruamviboonsuk, and Laura~M Vardoulakis.
\newblock A human-centered evaluation of a deep learning system deployed in
  clinics for the detection of diabetic retinopathy.
\newblock In \emph{Proceedings of the 2020 CHI Conference on Human Factors in
  Computing Systems}, pages 1--12, 2020.

\bibitem[Belkin et~al.(2018)Belkin, Hsu, Ma, and Mandal]{belkin2018reconciling}
Mikhail Belkin, Daniel Hsu, Siyuan Ma, and Soumik Mandal.
\newblock Reconciling modern machine learning and the bias-variance trade-off.
\newblock \emph{arXiv preprint arXiv:1812.11118}, 2018.

\bibitem[Bender and Koller(2020)]{bender-koller-2020-climbing}
Emily~M. Bender and Alexander Koller.
\newblock Climbing towards {NLU}: {On} meaning, form, and understanding in the
  age of data.
\newblock In \emph{Proceedings of the 58th Annual Meeting of the Association
  for Computational Linguistics}, pages 5185--5198, Online, July 2020.
  Association for Computational Linguistics.
\newblock \doi{10.18653/v1/2020.acl-main.463}.
\newblock URL \url{https://www.aclweb.org/anthology/2020.acl-main.463}.

\bibitem[Bengio(2017)]{bengio2017consciousness}
Yoshua Bengio.
\newblock The consciousness prior.
\newblock \emph{arXiv preprint arXiv:1709.08568}, 2017.

\bibitem[Berg et~al.(2019)Berg, Harpak, Sinnott-Armstrong, Joergensen,
  Mostafavi, Field, Boyle, Zhang, Racimo, Pritchard, and
  Coop]{Berg2019-prs-ancestry}
Jeremy~J Berg, Arbel Harpak, Nasa Sinnott-Armstrong, Anja~Moltke Joergensen,
  Hakhamanesh Mostafavi, Yair Field, Evan~August Boyle, Xinjun Zhang, Fernando
  Racimo, Jonathan~K Pritchard, and Graham Coop.
\newblock Reduced signal for polygenic adaptation of height in {UK} biobank.
\newblock \emph{Elife}, 8, March 2019.

\bibitem[Bolukbasi et~al.(2016)Bolukbasi, Chang, Zou, Saligrama, and
  Kalai]{bolukbasi2016man}
Tolga Bolukbasi, Kai-Wei Chang, James~Y Zou, Venkatesh Saligrama, and Adam~T
  Kalai.
\newblock Man is to computer programmer as woman is to homemaker? debiasing
  word embeddings.
\newblock In \emph{Advances in neural information processing systems}, pages
  4349--4357, 2016.

\bibitem[Bowman et~al.(2015)Bowman, Angeli, Potts, and
  Manning]{bowman-etal-2015-large}
Samuel~R. Bowman, Gabor Angeli, Christopher Potts, and Christopher~D. Manning.
\newblock A large annotated corpus for learning natural language inference.
\newblock In \emph{Proceedings of the 2015 Conference on Empirical Methods in
  Natural Language Processing}, pages 632--642, Lisbon, Portugal, September
  2015. Association for Computational Linguistics.
\newblock \doi{10.18653/v1/D15-1075}.
\newblock URL \url{https://www.aclweb.org/anthology/D15-1075}.

\bibitem[Boyd et~al.(2012)Boyd, Costa, Davis, and Page]{Boyd2012}
Kendrick Boyd, V{\'{i}}tor~Santos Costa, Jesse Davis, and C.~David Page.
\newblock {Unachievable region in precision-recall space and its effect on
  empirical evaluation}.
\newblock In \emph{Proceedings of the 29th International Conference on Machine
  Learning, ICML 2012}, volume~1, pages 639--646, 2012.
\newblock ISBN 9781450312851.

\bibitem[Brisimi et~al.(2019)Brisimi, Xu, Wang, Dai, and
  Paschalidis]{Brisimi2019}
Theodora~S. Brisimi, Tingting Xu, Taiyao Wang, Wuyang Dai, and Ioannis~Ch
  Paschalidis.
\newblock {Predicting diabetes-related hospitalizations based on electronic
  health records}.
\newblock \emph{Statistical Methods in Medical Research}, 28\penalty0
  (12):\penalty0 3667--3682, dec 2019.
\newblock ISSN 14770334.
\newblock \doi{10.1177/0962280218810911}.

\bibitem[Buolamwini and Gebru(2018)]{buolamwini2018gender}
Joy Buolamwini and Timnit Gebru.
\newblock Gender shades: Intersectional accuracy disparities in commercial
  gender classification.
\newblock In \emph{Conference on fairness, accountability and transparency},
  pages 77--91, 2018.

\bibitem[Caliskan et~al.(2017)Caliskan, Bryson, and
  Narayanan]{caliskan2017semantics}
Aylin Caliskan, Joanna~J Bryson, and Arvind Narayanan.
\newblock Semantics derived automatically from language corpora contain
  human-like biases.
\newblock \emph{Science}, 356\penalty0 (6334):\penalty0 183--186, 2017.

\bibitem[{CARDIoGRAMplusC4D Consortium} et~al.(2013){CARDIoGRAMplusC4D
  Consortium}, Deloukas, Kanoni, Willenborg, Farrall, Assimes, Thompson,
  Ingelsson, Saleheen, Erdmann, Goldstein, Stirrups, K{\"o}nig, Cazier,
  Johansson, Hall, Lee, Willer, Chambers, Esko, Folkersen, Goel, Grundberg,
  Havulinna, Ho, Hopewell, Eriksson, Kleber, Kristiansson, Lundmark,
  Lyytik{\"a}inen, Rafelt, Shungin, Strawbridge, Thorleifsson, Tikkanen,
  Van~Zuydam, Voight, Waite, Zhang, Ziegler, Absher, Altshuler, Balmforth,
  Barroso, Braund, Burgdorf, Claudi-Boehm, Cox, Dimitriou, Do, {DIAGRAM
  Consortium}, {CARDIOGENICS Consortium}, Doney, El~Mokhtari, Eriksson,
  Fischer, Fontanillas, Franco-Cereceda, Gigante, Groop, Gustafsson, Hager,
  Hallmans, Han, Hunt, Kang, Illig, Kessler, Knowles, Kolovou, Kuusisto,
  Langenberg, Langford, Leander, Lokki, Lundmark, McCarthy, Meisinger,
  Melander, Mihailov, Maouche, Morris, M{\"u}ller-Nurasyid, {MuTHER
  Consortium}, Nikus, Peden, Rayner, Rasheed, Rosinger, Rubin, Rumpf,
  Sch{\"a}fer, Sivananthan, Song, Stewart, Tan, Thorgeirsson, van~der Schoot,
  Wagner, {Wellcome Trust Case Control Consortium}, Wells, Wild, Yang, Amouyel,
  Arveiler, Basart, Boehnke, Boerwinkle, Brambilla, Cambien, Cupples, de~Faire,
  Dehghan, Diemert, Epstein, Evans, Ferrario, Ferri{\`e}res, Gauguier, Go,
  Goodall, Gudnason, Hazen, Holm, Iribarren, Jang, K{\"a}h{\"o}nen, Kee, Kim,
  Klopp, Koenig, Kratzer, Kuulasmaa, Laakso, Laaksonen, Lee, Lind, Ouwehand,
  Parish, Park, Pedersen, Peters, Quertermous, Rader, Salomaa, Schadt, Shah,
  Sinisalo, Stark, Stefansson, Tr{\'e}gou{\"e}t, Virtamo, Wallentin, Wareham,
  Zimmermann, Nieminen, Hengstenberg, Sandhu, Pastinen, Syv{\"a}nen, Hovingh,
  Dedoussis, Franks, Lehtim{\"a}ki, Metspalu, Zalloua, Siegbahn, Schreiber,
  Ripatti, Blankenberg, Perola, Clarke, Boehm, O'Donnell, Reilly, M{\"a}rz,
  Collins, Kathiresan, Hamsten, Kooner, Thorsteinsdottir, Danesh, Palmer,
  Roberts, Watkins, Schunkert, and Samani]{CAD2013-prs}
{CARDIoGRAMplusC4D Consortium}, Panos Deloukas, Stavroula Kanoni, Christina
  Willenborg, Martin Farrall, Themistocles~L Assimes, John~R Thompson, Erik
  Ingelsson, Danish Saleheen, Jeanette Erdmann, Benjamin~A Goldstein, Kathleen
  Stirrups, Inke~R K{\"o}nig, Jean-Baptiste Cazier, Asa Johansson, Alistair~S
  Hall, Jong-Young Lee, Cristen~J Willer, John~C Chambers, T{\~o}nu Esko, Lasse
  Folkersen, Anuj Goel, Elin Grundberg, Aki~S Havulinna, Weang~K Ho, Jemma~C
  Hopewell, Niclas Eriksson, Marcus~E Kleber, Kati Kristiansson, Per Lundmark,
  Leo-Pekka Lyytik{\"a}inen, Suzanne Rafelt, Dmitry Shungin, Rona~J
  Strawbridge, Gudmar Thorleifsson, Emmi Tikkanen, Natalie Van~Zuydam,
  Benjamin~F Voight, Lindsay~L Waite, Weihua Zhang, Andreas Ziegler, Devin
  Absher, David Altshuler, Anthony~J Balmforth, In{\^e}s Barroso, Peter~S
  Braund, Christof Burgdorf, Simone Claudi-Boehm, David Cox, Maria Dimitriou,
  Ron Do, {DIAGRAM Consortium}, {CARDIOGENICS Consortium}, Alex S~F Doney,
  Noureddine El~Mokhtari, Per Eriksson, Krista Fischer, Pierre Fontanillas,
  Anders Franco-Cereceda, Bruna Gigante, Leif Groop, Stefan Gustafsson,
  J{\"o}rg Hager, G{\"o}ran Hallmans, Bok-Ghee Han, Sarah~E Hunt, Hyun~M Kang,
  Thomas Illig, Thorsten Kessler, Joshua~W Knowles, Genovefa Kolovou, Johanna
  Kuusisto, Claudia Langenberg, Cordelia Langford, Karin Leander, Marja-Liisa
  Lokki, Anders Lundmark, Mark~I McCarthy, Christa Meisinger, Olle Melander,
  Evelin Mihailov, Seraya Maouche, Andrew~D Morris, Martina
  M{\"u}ller-Nurasyid, {MuTHER Consortium}, Kjell Nikus, John~F Peden,
  N~William Rayner, Asif Rasheed, Silke Rosinger, Diana Rubin, Moritz~P Rumpf,
  Arne Sch{\"a}fer, Mohan Sivananthan, Ci~Song, Alexandre F~R Stewart,
  Sian-Tsung Tan, Gudmundur Thorgeirsson, C~Ellen van~der Schoot, Peter~J
  Wagner, {Wellcome Trust Case Control Consortium}, George~A Wells, Philipp~S
  Wild, Tsun-Po Yang, Philippe Amouyel, Dominique Arveiler, Hanneke Basart,
  Michael Boehnke, Eric Boerwinkle, Paolo Brambilla, Francois Cambien,
  Adrienne~L Cupples, Ulf de~Faire, Abbas Dehghan, Patrick Diemert, Stephen~E
  Epstein, Alun Evans, Marco~M Ferrario, Jean Ferri{\`e}res, Dominique
  Gauguier, Alan~S Go, Alison~H Goodall, Villi Gudnason, Stanley~L Hazen, Hilma
  Holm, Carlos Iribarren, Yangsoo Jang, Mika K{\"a}h{\"o}nen, Frank Kee,
  Hyo-Soo Kim, Norman Klopp, Wolfgang Koenig, Wolfgang Kratzer, Kari Kuulasmaa,
  Markku Laakso, Reijo Laaksonen, Ji-Young Lee, Lars Lind, Willem~H Ouwehand,
  Sarah Parish, Jeong~E Park, Nancy~L Pedersen, Annette Peters, Thomas
  Quertermous, Daniel~J Rader, Veikko Salomaa, Eric Schadt, Svati~H Shah, Juha
  Sinisalo, Klaus Stark, Kari Stefansson, David-Alexandre Tr{\'e}gou{\"e}t,
  Jarmo Virtamo, Lars Wallentin, Nicholas Wareham, Martina~E Zimmermann,
  Markku~S Nieminen, Christian Hengstenberg, Manjinder~S Sandhu, Tomi Pastinen,
  Ann-Christine Syv{\"a}nen, G~Kees Hovingh, George Dedoussis, Paul~W Franks,
  Terho Lehtim{\"a}ki, Andres Metspalu, Pierre~A Zalloua, Agneta Siegbahn,
  Stefan Schreiber, Samuli Ripatti, Stefan~S Blankenberg, Markus Perola, Robert
  Clarke, Bernhard~O Boehm, Christopher O'Donnell, Muredach~P Reilly, Winfried
  M{\"a}rz, Rory Collins, Sekar Kathiresan, Anders Hamsten, Jaspal~S Kooner,
  Unnur Thorsteinsdottir, John Danesh, Colin N~A Palmer, Robert Roberts, Hugh
  Watkins, Heribert Schunkert, and Nilesh~J Samani.
\newblock Large-scale association analysis identifies new risk loci for
  coronary artery disease.
\newblock \emph{Nat. Genet.}, 45\penalty0 (1):\penalty0 25--33, January 2013.

\bibitem[Caruana et~al.(2015)Caruana, Lou, Gehrke, Koch, Sturm, and
  Elhadad]{Caruana2015}
Rich Caruana, Yin Lou, Johannes Gehrke, Paul Koch, Marc Sturm, and Noemie
  Elhadad.
\newblock {Intelligible Models for HealthCare}.
\newblock In \emph{Proceedings of the 21th ACM SIGKDD International Conference
  on Knowledge Discovery and Data Mining - KDD '15}, pages 1721--1730, 2015.
\newblock ISBN 9781450336642.
\newblock \doi{10.1145/2783258.2788613}.
\newblock URL \url{http://dx.doi.org/10.1145/2783258.2788613
  http://dl.acm.org/citation.cfm?doid=2783258.2788613}.

\bibitem[Cer et~al.(2017)Cer, Diab, Agirre, Lopez-Gazpio, and
  Specia]{cer-etal-2017-semeval}
Daniel Cer, Mona Diab, Eneko Agirre, I{\~n}igo Lopez-Gazpio, and Lucia Specia.
\newblock {S}em{E}val-2017 task 1: Semantic textual similarity multilingual and
  crosslingual focused evaluation.
\newblock In \emph{Proceedings of the 11th International Workshop on Semantic
  Evaluation ({S}em{E}val-2017)}, pages 1--14, Vancouver, Canada, August 2017.
  Association for Computational Linguistics.
\newblock \doi{10.18653/v1/S17-2001}.
\newblock URL \url{https://www.aclweb.org/anthology/S17-2001}.

\bibitem[Chaudhari et~al.(2019)Chaudhari, Choromanska, Soatto, LeCun, Baldassi,
  Borgs, Chayes, Sagun, and Zecchina]{chaudhari2019entropy}
Pratik Chaudhari, Anna Choromanska, Stefano Soatto, Yann LeCun, Carlo Baldassi,
  Christian Borgs, Jennifer Chayes, Levent Sagun, and Riccardo Zecchina.
\newblock Entropy-sgd: Biasing gradient descent into wide valleys.
\newblock \emph{Journal of Statistical Mechanics: Theory and Experiment},
  2019\penalty0 (12):\penalty0 124018, 2019.

\bibitem[Chizat et~al.(2019)Chizat, Oyallon, and Bach]{chizat2019lazy}
Lenaic Chizat, Edouard Oyallon, and Francis Bach.
\newblock On lazy training in differentiable programming.
\newblock 2019.

\bibitem[Choi et~al.(2017)Choi, {Taha Bahadori}, Song, Stewart, and
  Sun]{Choi2017}
Edward Choi, Mohammad {Taha Bahadori}, Le~Song, Walter~F Stewart, and Jimeng
  Sun.
\newblock {GRAM: Graph-based Attention Model for Healthcare Representation
  Learning}.
\newblock 2017.
\newblock \doi{10.1145/3097983.3098126}.
\newblock URL \url{http://dx.doi.org/10.1145/3097983.3098126}.

\bibitem[Collins et~al.(2015)Collins, Reitsma, Altman, and
  Moons]{collins2015transparent}
Gary~S Collins, Johannes~B Reitsma, Douglas~G Altman, and Karel~GM Moons.
\newblock Transparent reporting of a multivariable prediction model for
  individual prognosis or diagnosis (tripod): the tripod statement.
\newblock \emph{British Journal of Surgery}, 102\penalty0 (3):\penalty0
  148--158, 2015.

\bibitem[Collins et~al.(2017)Collins, Sohl-Dickstein, and
  Sussillo]{Collins2017}
Jasmine Collins, Jascha Sohl-Dickstein, and David Sussillo.
\newblock {Capacity and trainability in recurrent neural networks}.
\newblock In \emph{5th International Conference on Learning Representations,
  ICLR 2017 - Conference Track Proceedings}, 2017.

\bibitem[De-Arteaga et~al.(2019)De-Arteaga, Romanov, Wallach, Chayes, Borgs,
  Chouldechova, Geyik, Kenthapadi, and Kalai]{de2019bias}
Maria De-Arteaga, Alexey Romanov, Hanna Wallach, Jennifer Chayes, Christian
  Borgs, Alexandra Chouldechova, Sahin Geyik, Krishnaram Kenthapadi, and
  Adam~Tauman Kalai.
\newblock Bias in bios: A case study of semantic representation bias in a
  high-stakes setting.
\newblock In \emph{Proceedings of the Conference on Fairness, Accountability,
  and Transparency}, pages 120--128, 2019.

\bibitem[Deng et~al.(2009)Deng, Dong, Socher, Li, Li, and
  Fei-Fei]{imagenet_cvpr09}
J.~Deng, W.~Dong, R.~Socher, L.-J. Li, K.~Li, and L.~Fei-Fei.
\newblock {ImageNet: A Large-Scale Hierarchical Image Database}.
\newblock In \emph{CVPR09}, 2009.

\bibitem[Devlin et~al.(2019)Devlin, Chang, Lee, and Toutanova]{devlin2019bert}
Jacob Devlin, Ming-Wei Chang, Kenton Lee, and Kristina Toutanova.
\newblock Bert: Pre-training of deep bidirectional transformers for language
  understanding.
\newblock In \emph{Proceedings of the 2019 Conference of the North American
  Chapter of the Association for Computational Linguistics: Human Language
  Technologies, Volume 1 (Long and Short Papers)}, pages 4171--4186, 2019.

\bibitem[Djolonga et~al.(2020)Djolonga, Yung, Tschannen, Romijnders, Beyer,
  Kolesnikov, Puigcerver, Minderer, D'Amour, Moldovan,
  et~al.]{djolonga2020robustness}
Josip Djolonga, Jessica Yung, Michael Tschannen, Rob Romijnders, Lucas Beyer,
  Alexander Kolesnikov, Joan Puigcerver, Matthias Minderer, Alexander D'Amour,
  Dan Moldovan, et~al.
\newblock On robustness and transferability of convolutional neural networks.
\newblock \emph{arXiv preprint arXiv:2007.08558}, 2020.

\bibitem[Dodge et~al.(2020)Dodge, Ilharco, Schwartz, Farhadi, Hajishirzi, and
  Smith]{dodge2020fine}
Jesse Dodge, Gabriel Ilharco, Roy Schwartz, Ali Farhadi, Hannaneh Hajishirzi,
  and Noah Smith.
\newblock Fine-tuning pretrained language models: Weight initializations, data
  orders, and early stopping.
\newblock \emph{arXiv preprint arXiv:2002.06305}, 2020.

\bibitem[Duncan et~al.(2019)Duncan, Shen, Gelaye, Meijsen, Ressler, Feldman,
  Peterson, and Domingue]{Duncan2019-prs-ancestry}
L~Duncan, H~Shen, B~Gelaye, J~Meijsen, K~Ressler, M~Feldman, R~Peterson, and
  B~Domingue.
\newblock Analysis of polygenic risk score usage and performance in diverse
  human populations.
\newblock \emph{Nat. Commun.}, 10\penalty0 (1):\penalty0 3328, July 2019.

\bibitem[Dusenberry et~al.(2020)Dusenberry, Tran, Choi, Kemp, Nixon, Jerfel,
  Heller, and Dai]{Dusenberry2020}
Michael~W Dusenberry, Dustin Tran, Edward Choi, Jonas Kemp, Jeremy Nixon,
  Ghassen Jerfel, Katherine Heller, and Andrew~M Dai.
\newblock Analyzing the role of model uncertainty for electronic health
  records.
\newblock In \emph{Proceedings of the ACM Conference on Health, Inference, and
  Learning}, pages 204--213, 2020.

\bibitem[Esteva et~al.(2017)Esteva, Kuprel, Novoa, Ko, Swetter, Blau, and
  Thrun]{esteva2017dermatologist}
Andre Esteva, Brett Kuprel, Roberto~A Novoa, Justin Ko, Susan~M Swetter,
  Helen~M Blau, and Sebastian Thrun.
\newblock Dermatologist-level classification of skin cancer with deep neural
  networks.
\newblock \emph{Nature}, 542\penalty0 (7639):\penalty0 115--118, 2017.

\bibitem[Feng et~al.(2019)Feng, Le, and McCoy]{Feng2019}
Chenchen Feng, David Le, and Allison~B. McCoy.
\newblock {Using Electronic Health Records to Identify Adverse Drug Events in
  Ambulatory Care: A Systematic Review}.
\newblock \emph{Applied Clinical Informatics}, 10\penalty0 (1):\penalty0
  123--128, 2019.
\newblock ISSN 18690327.
\newblock \doi{10.1055/s-0039-1677738}.

\bibitem[Fisher et~al.(2019)Fisher, Rudin, and Dominici]{fisher2019all}
Aaron Fisher, Cynthia Rudin, and Francesca Dominici.
\newblock All models are wrong, but many are useful: Learning a variable's
  importance by studying an entire class of prediction models simultaneously.
\newblock \emph{Journal of Machine Learning Research}, 20\penalty0
  (177):\penalty0 1--81, 2019.

\bibitem[Fitzpatrick(1975)]{fitzpatrick1975sun}
TB~Fitzpatrick.
\newblock Sun and skin.
\newblock \emph{Journal de Medecine Esthetique}, 2:\penalty0 33--34, 1975.

\bibitem[Flaxman et~al.(2020)Flaxman, Mishra, Gandy, Unwin, Mellan, Coupland,
  Whittaker, Zhu, Berah, Eaton, et~al.]{flaxman2020estimating}
Seth Flaxman, Swapnil Mishra, Axel Gandy, H~Juliette~T Unwin, Thomas~A Mellan,
  Helen Coupland, Charles Whittaker, Harrison Zhu, Tresnia Berah, Jeffrey~W
  Eaton, et~al.
\newblock Estimating the effects of non-pharmaceutical interventions on
  covid-19 in europe.
\newblock \emph{Nature}, 584\penalty0 (7820):\penalty0 257--261, 2020.

\bibitem[Fort et~al.(2019)Fort, Hu, and Lakshminarayanan]{fort2019deep}
Stanislav Fort, Huiyi Hu, and Balaji Lakshminarayanan.
\newblock Deep ensembles: A loss landscape perspective.
\newblock \emph{arXiv preprint arXiv:1912.02757}, 2019.

\bibitem[Frankle et~al.(2020)Frankle, Dziugaite, Roy, and
  Carbin]{frankle2020linear}
Jonathan Frankle, Gintare~Karolina Dziugaite, Daniel~M Roy, and Michael Carbin.
\newblock Linear mode connectivity and the lottery ticket hypothesis.
\newblock In \emph{Proceedings of the 37th International Conference on Machine
  Learning}, 2020.

\bibitem[Futoma et~al.(2020)Futoma, Simons, Panch, Doshi-Velez, and
  Celi]{futoma2020myth}
Joseph Futoma, Morgan Simons, Trishan Panch, Finale Doshi-Velez, and
  Leo~Anthony Celi.
\newblock The myth of generalisability in clinical research and machine
  learning in health care.
\newblock \emph{The Lancet Digital Health}, 2\penalty0 (9):\penalty0 e489 --
  e492, 2020.
\newblock ISSN 2589-7500.
\newblock \doi{https://doi.org/10.1016/S2589-7500(20)30186-2}.
\newblock URL
  \url{http://www.sciencedirect.com/science/article/pii/S2589750020301862}.

\bibitem[Garg et~al.(2019)Garg, Perot, Limtiaco, Taly, Chi, and
  Beutel]{garg2019counterfactual}
Sahaj Garg, Vincent Perot, Nicole Limtiaco, Ankur Taly, Ed~H Chi, and Alex
  Beutel.
\newblock Counterfactual fairness in text classification through robustness.
\newblock In \emph{Proceedings of the 2019 AAAI/ACM Conference on AI, Ethics,
  and Society}, pages 219--226, 2019.

\bibitem[Garipov et~al.(2018)Garipov, Izmailov, Podoprikhin, Vetrov, and
  Wilson]{garipov2018loss}
Timur Garipov, Pavel Izmailov, Dmitrii Podoprikhin, Dmitry~P Vetrov, and
  Andrew~G Wilson.
\newblock Loss surfaces, mode connectivity, and fast ensembling of dnns.
\newblock In \emph{Advances in Neural Information Processing Systems}, pages
  8789--8798, 2018.

\bibitem[Geirhos et~al.(2019)Geirhos, Rubisch, Michaelis, Bethge, Wichmann, and
  Brendel]{geirhos2018imagenettrained}
Robert Geirhos, Patricia Rubisch, Claudio Michaelis, Matthias Bethge, Felix~A.
  Wichmann, and Wieland Brendel.
\newblock Imagenet-trained {CNN}s are biased towards texture; increasing shape
  bias improves accuracy and robustness.
\newblock In \emph{International Conference on Learning Representations}, 2019.
\newblock URL \url{https://openreview.net/forum?id=Bygh9j09KX}.

\bibitem[Geirhos et~al.(2020)Geirhos, Jacobsen, Michaelis, Zemel, Brendel,
  Bethge, and Wichmann]{geirhos2020shortcut}
Robert Geirhos, J{\"o}rn-Henrik Jacobsen, Claudio Michaelis, Richard Zemel,
  Wieland Brendel, Matthias Bethge, and Felix~A Wichmann.
\newblock Shortcut learning in deep neural networks.
\newblock \emph{arXiv preprint arXiv:2004.07780}, 2020.

\bibitem[Goodfellow et~al.(2016)Goodfellow, Bengio, Courville, and
  Bengio]{goodfellow2016deep}
Ian Goodfellow, Yoshua Bengio, Aaron Courville, and Yoshua Bengio.
\newblock \emph{Deep learning}, volume~1.
\newblock MIT press Cambridge, 2016.

\bibitem[Gulshan et~al.(2016)Gulshan, Peng, Coram, Stumpe, Wu, Narayanaswamy,
  Venugopalan, Widner, Madams, Cuadros, et~al.]{gulshan2016development}
Varun Gulshan, Lily Peng, Marc Coram, Martin~C Stumpe, Derek Wu, Arunachalam
  Narayanaswamy, Subhashini Venugopalan, Kasumi Widner, Tom Madams, Jorge
  Cuadros, et~al.
\newblock Development and validation of a deep learning algorithm for detection
  of diabetic retinopathy in retinal fundus photographs.
\newblock \emph{Jama}, 316\penalty0 (22):\penalty0 2402--2410, 2016.

\bibitem[He et~al.(2016)He, Zhang, Ren, and Sun]{he2016deep}
Kaiming He, Xiangyu Zhang, Shaoqing Ren, and Jian Sun.
\newblock Deep residual learning for image recognition.
\newblock In \emph{Proceedings of the IEEE conference on computer vision and
  pattern recognition}, pages 770--778, 2016.

\bibitem[Heinze-Deml et~al.(2018)Heinze-Deml, Peters, and
  Meinshausen]{heinze2018invariant}
Christina Heinze-Deml, Jonas Peters, and Nicolai Meinshausen.
\newblock Invariant causal prediction for nonlinear models.
\newblock \emph{Journal of Causal Inference}, 6\penalty0 (2), 2018.

\bibitem[Hendrycks and Dietterich(2019)]{hendrycks2018benchmarking}
Dan Hendrycks and Thomas Dietterich.
\newblock Benchmarking neural network robustness to common corruptions and
  perturbations.
\newblock In \emph{International Conference on Learning Representations}, 2019.
\newblock URL \url{https://openreview.net/forum?id=HJz6tiCqYm}.

\bibitem[Hendrycks et~al.(2019)Hendrycks, Zhao, Basart, Steinhardt, and
  Song]{hendrycks2019natural}
Dan Hendrycks, Kevin Zhao, Steven Basart, Jacob Steinhardt, and Dawn Song.
\newblock Natural adversarial examples.
\newblock \emph{arXiv preprint arXiv:1907.07174}, 2019.

\bibitem[Hendrycks et~al.(2020)Hendrycks, Basart, Mu, Kadavath, Wang, Dorundo,
  Desai, Zhu, Parajuli, Guo, et~al.]{hendrycks2020many}
Dan Hendrycks, Steven Basart, Norman Mu, Saurav Kadavath, Frank Wang, Evan
  Dorundo, Rahul Desai, Tyler Zhu, Samyak Parajuli, Mike Guo, et~al.
\newblock The many faces of robustness: A critical analysis of
  out-of-distribution generalization.
\newblock \emph{arXiv preprint arXiv:2006.16241}, 2020.

\bibitem[Hochreiter and Schmidhuber(1997)]{Hochreiter1997}
Sepp Hochreiter and J\"urgen Schmidhuber.
\newblock {Long Short-Term Memory}.
\newblock \emph{Neural Computation}, 9\penalty0 (8):\penalty0 1735--1780, 1997.
\newblock URL
  \url{http://www7.informatik.tu-muenchen.de/{~}hochreithttp://www.idsia.ch/{~}juergen}.

\bibitem[Hoffmann et~al.(2019)Hoffmann, Latza, Baumeister, Br{\"u}nger,
  Buttmann-Schweiger, Hardt, Hoffmann, Karch, Richter, Schmidt,
  et~al.]{hoffmann2019guidelines}
Wolfgang Hoffmann, Ute Latza, Sebastian~E Baumeister, Martin Br{\"u}nger, Nina
  Buttmann-Schweiger, Juliane Hardt, Verena Hoffmann, Andr{\'e} Karch, Adrian
  Richter, Carsten~Oliver Schmidt, et~al.
\newblock Guidelines and recommendations for ensuring good epidemiological
  practice (gep): a guideline developed by the german society for epidemiology.
\newblock \emph{European journal of epidemiology}, 34\penalty0 (3):\penalty0
  301--317, 2019.

\bibitem[Hooker(2020)]{hooker2020hardware}
Sara Hooker.
\newblock The hardware lottery.
\newblock \emph{arXiv preprint arXiv:2009.06489}, 2020.

\bibitem[Hovy et~al.(2006)Hovy, Marcus, Palmer, Ramshaw, and
  Weischedel]{hovy-etal-2006-ontonotes}
Eduard Hovy, Mitchell Marcus, Martha Palmer, Lance Ramshaw, and Ralph
  Weischedel.
\newblock {O}nto{N}otes: The 90{\%} solution.
\newblock In \emph{Proceedings of the Human Language Technology Conference of
  the {NAACL}, Companion Volume: Short Papers}, pages 57--60, New York City,
  USA, June 2006. Association for Computational Linguistics.
\newblock URL \url{https://www.aclweb.org/anthology/N06-2015}.

\bibitem[Howard and Ruder(2018)]{howard2018universal}
Jeremy Howard and Sebastian Ruder.
\newblock Universal language model fine-tuning for text classification.
\newblock In \emph{Proceedings of the 56th Annual Meeting of the Association
  for Computational Linguistics (Volume 1: Long Papers)}, pages 328--339, 2018.

\bibitem[Ilyas et~al.(2019)Ilyas, Santurkar, Tsipras, Engstrom, Tran, and
  Madry]{ilyas2019adversarial}
Andrew Ilyas, Shibani Santurkar, Dimitris Tsipras, Logan Engstrom, Brandon
  Tran, and Aleksander Madry.
\newblock Adversarial examples are not bugs, they are features.
\newblock In \emph{Advances in Neural Information Processing Systems}, pages
  125--136, 2019.

\bibitem[{International Schizophrenia Consortium} et~al.(2009){International
  Schizophrenia Consortium}, Purcell, Wray, Stone, Visscher, O'Donovan,
  Sullivan, and Sklar]{ISC2009-prs}
{International Schizophrenia Consortium}, Shaun~M Purcell, Naomi~R Wray,
  Jennifer~L Stone, Peter~M Visscher, Michael~C O'Donovan, Patrick~F Sullivan,
  and Pamela Sklar.
\newblock Common polygenic variation contributes to risk of schizophrenia and
  bipolar disorder.
\newblock \emph{Nature}, 460\penalty0 (7256):\penalty0 748--752, August 2009.

\bibitem[Izmailov et~al.(2018)Izmailov, Podoprikhin, Garipov, Vetrov, and
  Wilson]{izmailov2018averaging}
Pavel Izmailov, Dmitrii Podoprikhin, Timur Garipov, Dmitry Vetrov, and
  Andrew~Gordon Wilson.
\newblock Averaging weights leads to wider optima and better generalization.
\newblock \emph{arXiv preprint arXiv:1803.05407}, 2018.

\bibitem[Jacovi et~al.(2020)Jacovi, Marasović, Miller, and
  Goldberg]{jacovi2020formalizing}
Alon Jacovi, Ana Marasović, Tim Miller, and Yoav Goldberg.
\newblock Formalizing trust in artificial intelligence: Prerequisites, causes
  and goals of human trust in ai.
\newblock \emph{arXiv preprint arXiv:2010.07487}, 2020.

\bibitem[Kaushik et~al.(2020)Kaushik, Hovy, and Lipton]{Kaushik2020Learning}
Divyansh Kaushik, Eduard Hovy, and Zachary Lipton.
\newblock Learning the difference that makes a difference with
  counterfactually-augmented data.
\newblock In \emph{International Conference on Learning Representations}, 2020.
\newblock URL \url{https://openreview.net/forum?id=Sklgs0NFvr}.

\bibitem[Kellum and Bihorac(2019)]{kellum2019artificial}
John~A Kellum and Azra Bihorac.
\newblock Artificial intelligence to predict aki: is it a breakthrough?
\newblock \emph{Nature Reviews Nephrology}, pages 1--2, 2019.

\bibitem[Kelly et~al.(2019)Kelly, Karthikesalingam, Suleyman, Corrado, and
  King]{kelly2019key}
Christopher~J Kelly, Alan Karthikesalingam, Mustafa Suleyman, Greg Corrado, and
  Dominic King.
\newblock Key challenges for delivering clinical impact with artificial
  intelligence.
\newblock \emph{BMC medicine}, 17\penalty0 (1):\penalty0 195, 2019.

\bibitem[Khera et~al.(2018)Khera, Chaffin, Aragam, Haas, Roselli, Choi,
  Natarajan, Lander, Lubitz, Ellinor, and Kathiresan]{Khera2018-prs-diseases}
Amit~V Khera, Mark Chaffin, Krishna~G Aragam, Mary~E Haas, Carolina Roselli,
  Seung~Hoan Choi, Pradeep Natarajan, Eric~S Lander, Steven~A Lubitz, Patrick~T
  Ellinor, and Sekar Kathiresan.
\newblock Genome-wide polygenic scores for common diseases identify individuals
  with risk equivalent to monogenic mutations.
\newblock \emph{Nat. Genet.}, 50\penalty0 (9):\penalty0 1219--1224, September
  2018.

\bibitem[Khwaja(2012)]{Khwaja2012}
Arif Khwaja.
\newblock {KDIGO clinical practice guidelines for acute kidney injury}.
\newblock \emph{Nephron - Clinical Practice}, 120\penalty0 (4), oct 2012.
\newblock ISSN 16602110.
\newblock \doi{10.1159/000339789}.

\bibitem[Kleinberg et~al.(2015)Kleinberg, Ludwig, Mullainathan, and
  Obermeyer]{kleinberg2015prediction}
Jon Kleinberg, Jens Ludwig, Sendhil Mullainathan, and Ziad Obermeyer.
\newblock Prediction policy problems.
\newblock \emph{American Economic Review}, 105\penalty0 (5):\penalty0 491--95,
  2015.

\bibitem[Kolesnikov et~al.(2019)Kolesnikov, Beyer, Zhai, Puigcerver, Yung,
  Gelly, and Houlsby]{kolesnikov2019large}
Alexander Kolesnikov, Lucas Beyer, Xiaohua Zhai, Joan Puigcerver, Jessica Yung,
  Sylvain Gelly, and Neil Houlsby.
\newblock Large scale learning of general visual representations for transfer.
\newblock \emph{arXiv preprint arXiv:1912.11370}, 2019.

\bibitem[Krause et~al.(2018)Krause, Gulshan, Rahimy, Karth, Widner, Corrado,
  Peng, and Webster]{krause2018grader}
Jonathan Krause, Varun Gulshan, Ehsan Rahimy, Peter Karth, Kasumi Widner,
  Greg~S Corrado, Lily Peng, and Dale~R Webster.
\newblock Grader variability and the importance of reference standards for
  evaluating machine learning models for diabetic retinopathy.
\newblock \emph{Ophthalmology}, 125\penalty0 (8):\penalty0 1264--1272, 2018.

\bibitem[Kusner et~al.(2017)Kusner, Loftus, Russell, and
  Silva]{kusner2017counterfactual}
Matt~J Kusner, Joshua Loftus, Chris Russell, and Ricardo Silva.
\newblock Counterfactual fairness.
\newblock In \emph{Advances in neural information processing systems}, pages
  4066--4076, 2017.

\bibitem[Lakshminarayanan et~al.(2017)Lakshminarayanan, Pritzel, and
  Blundell]{Lakshminarayanan2017deepensembles}
Balaji Lakshminarayanan, Alexander Pritzel, and Charles Blundell.
\newblock Simple and scalable predictive uncertainty estimation using deep
  ensembles.
\newblock In I.~Guyon, U.~V. Luxburg, S.~Bengio, H.~Wallach, R.~Fergus,
  S.~Vishwanathan, and R.~Garnett, editors, \emph{Advances in Neural
  Information Processing Systems 30}, pages 6402--6413. Curran Associates,
  Inc., 2017.
\newblock URL
  \url{http://papers.nips.cc/paper/7219-simple-and-scalable-predictive-uncertainty-estimation-using-deep-ensembles.pdf}.

\bibitem[Lan et~al.(2019)Lan, Chen, Goodman, Gimpel, Sharma, and
  Soricut]{lan2019albert}
Zhenzhong Lan, Mingda Chen, Sebastian Goodman, Kevin Gimpel, Piyush Sharma, and
  Radu Soricut.
\newblock Albert: A lite bert for self-supervised learning of language
  representations.
\newblock \emph{arXiv preprint arXiv:1909.11942}, 2019.

\bibitem[Ledoit and P{\'e}ch{\'e}(2011)]{ledoit2011eigenvectors}
Olivier Ledoit and Sandrine P{\'e}ch{\'e}.
\newblock Eigenvectors of some large sample covariance matrix ensembles.
\newblock \emph{Probability Theory and Related Fields}, 151\penalty0
  (1-2):\penalty0 233--264, 2011.

\bibitem[Lei et~al.(2018)Lei, Zhang, Wang, Dai, and Artzi]{Lei2018}
Tao Lei, Yu~Zhang, Sida~I. Wang, Hui Dai, and Yoav Artzi.
\newblock {Simple recurrent units for highly parallelizable recurrence}.
\newblock In \emph{Proceedings of the 2018 Conference on Empirical Methods in
  Natural Language Processing, EMzhou 2018}, pages 4470--4481. Association for
  Computational Linguistics, sep 2018.
\newblock ISBN 9781948087841.
\newblock \doi{10.18653/v1/d18-1477}.
\newblock URL \url{http://arxiv.org/abs/1709.02755}.

\bibitem[Linzen(2020)]{linzen-2020-accelerate}
Tal Linzen.
\newblock How can we accelerate progress towards human-like linguistic
  generalization?
\newblock In \emph{Proceedings of the 58th Annual Meeting of the Association
  for Computational Linguistics}, pages 5210--5217, Online, July 2020.
  Association for Computational Linguistics.
\newblock \doi{10.18653/v1/2020.acl-main.465}.
\newblock URL \url{https://www.aclweb.org/anthology/2020.acl-main.465}.

\bibitem[Liu et~al.(2020{\natexlab{a}})Liu, Rivera, Moher, Calvert, and
  Denniston]{liu2020reporting}
Xiaoxuan Liu, Samantha~Cruz Rivera, David Moher, Melanie~J Calvert, and
  Alastair~K Denniston.
\newblock Reporting guidelines for clinical trial reports for interventions
  involving artificial intelligence: the consort-ai extension.
\newblock \emph{bmj}, 370, 2020{\natexlab{a}}.

\bibitem[Liu et~al.(2019)Liu, Ott, Goyal, Du, Joshi, Chen, Levy, Lewis,
  Zettlemoyer, and Stoyanov]{liu2019roberta}
Yinhan Liu, Myle Ott, Naman Goyal, Jingfei Du, Mandar Joshi, Danqi Chen, Omer
  Levy, Mike Lewis, Luke Zettlemoyer, and Veselin Stoyanov.
\newblock Roberta: A robustly optimized bert pretraining approach.
\newblock \emph{arXiv preprint arXiv:1907.11692}, 2019.

\bibitem[Liu et~al.(2020{\natexlab{b}})Liu, Jain, Eng, Way, Lee, Bui, Kanada,
  de~Oliveira~Marinho, Gallegos, Gabriele, et~al.]{liu2020deep}
Yuan Liu, Ayush Jain, Clara Eng, David~H Way, Kang Lee, Peggy Bui, Kimberly
  Kanada, Guilherme de~Oliveira~Marinho, Jessica Gallegos, Sara Gabriele,
  et~al.
\newblock A deep learning system for differential diagnosis of skin diseases.
\newblock \emph{Nature Medicine}, pages 1--9, 2020{\natexlab{b}}.

\bibitem[Magliacane et~al.(2018)Magliacane, van Ommen, Claassen, Bongers,
  Versteeg, and Mooij]{Magliacane++_NeurIPS_18}
Sara Magliacane, Thijs van Ommen, Tom Claassen, Stephan Bongers, Philip
  Versteeg, and Joris~M Mooij.
\newblock Domain adaptation by using causal inference to predict invariant
  conditional distributions.
\newblock In S.~Bengio, H.~Wallach, H.~Larochelle, K.~Grauman, N.~Cesa-Bianchi,
  and R.~Garnett, editors, \emph{Advances in Neural Information Processing
  Systems 31 ({N}eur{IPS}2018)}, pages 10869--10879. Curran Associates, Inc.,
  2018.
\newblock URL
  \url{http://papers.nips.cc/paper/8282-domain-adaptation-by-using-causal-inference-to-predict-invariant-conditional-distributions.pdf}.

\bibitem[Martin et~al.(2017)Martin, Gignoux, Walters, Wojcik, Neale, Gravel,
  Daly, Bustamante, and Kenny]{Martin2017-prs-ancestry}
Alicia~R Martin, Christopher~R Gignoux, Raymond~K Walters, Genevieve~L Wojcik,
  Benjamin~M Neale, Simon Gravel, Mark~J Daly, Carlos~D Bustamante, and
  Eimear~E Kenny.
\newblock Human demographic history impacts genetic risk prediction across
  diverse populations.
\newblock \emph{Am. J. Hum. Genet.}, 100\penalty0 (4):\penalty0 635--649, April
  2017.

\bibitem[Martin et~al.(2019)Martin, Kanai, Kamatani, Okada, Neale, and
  Daly]{Martin2019-prs-fairness}
Alicia~R Martin, Masahiro Kanai, Yoichiro Kamatani, Yukinori Okada, Benjamin~M
  Neale, and Mark~J Daly.
\newblock Clinical use of current polygenic risk scores may exacerbate health
  disparities.
\newblock \emph{Nat. Genet.}, 51\penalty0 (4):\penalty0 584--591, April 2019.

\bibitem[Marx et~al.(2019)Marx, Calmon, and Ustun]{marx2019predictive}
Charles~T Marx, Flavio du~Pin Calmon, and Berk Ustun.
\newblock Predictive multiplicity in classification.
\newblock \emph{arXiv preprint arXiv:1909.06677}, 2019.

\bibitem[McCoy et~al.(2019{\natexlab{a}})McCoy, Min, and
  Linzen]{mccoy2019berts}
R~Thomas McCoy, Junghyun Min, and Tal Linzen.
\newblock Berts of a feather do not generalize together: Large variability in
  generalization across models with similar test set performance.
\newblock \emph{arXiv preprint arXiv:1911.02969}, 2019{\natexlab{a}}.

\bibitem[McCoy et~al.(2019{\natexlab{b}})McCoy, Pavlick, and
  Linzen]{mccoy2019right}
R~Thomas McCoy, Ellie Pavlick, and Tal Linzen.
\newblock Right for the wrong reasons: Diagnosing syntactic heuristics in
  natural language inference.
\newblock \emph{arXiv preprint arXiv:1902.01007}, 2019{\natexlab{b}}.

\bibitem[Mei and Montanari(2019)]{mei2019generalization}
Song Mei and Andrea Montanari.
\newblock The generalization error of random features regression: Precise
  asymptotics and double descent curve.
\newblock \emph{arXiv:1908.05355}, 2019.

\bibitem[Mikolov et~al.(2013)Mikolov, Chen, Corrado, and
  Dean]{mikolov2013efficient}
Tomas Mikolov, Kai Chen, Greg Corrado, and Jeffrey Dean.
\newblock Efficient estimation of word representations in vector space.
\newblock \emph{arXiv preprint arXiv:1301.3781}, 2013.

\bibitem[Morales et~al.(2018)Morales, Welter, Bowler, Cerezo, Harris, McMahon,
  Hall, Junkins, Milano, Hastings, Malangone, Buniello, Burdett, Flicek,
  Parkinson, Cunningham, Hindorff, and MacArthur]{Morales2018-gwas-ancestry}
Joannella Morales, Danielle Welter, Emily~H Bowler, Maria Cerezo, Laura~W
  Harris, Aoife~C McMahon, Peggy Hall, Heather~A Junkins, Annalisa Milano, Emma
  Hastings, Cinzia Malangone, Annalisa Buniello, Tony Burdett, Paul Flicek,
  Helen Parkinson, Fiona Cunningham, Lucia~A Hindorff, and Jacqueline A~L
  MacArthur.
\newblock A standardized framework for representation of ancestry data in
  genomics studies, with application to the {NHGRI-EBI} {GWAS} catalog.
\newblock \emph{Genome Biol.}, 19\penalty0 (1):\penalty0 21, February 2018.

\bibitem[Mullainathan and Spiess(2017)]{mullainathan2017machine}
Sendhil Mullainathan and Jann Spiess.
\newblock Machine learning: an applied econometric approach.
\newblock \emph{Journal of Economic Perspectives}, 31\penalty0 (2):\penalty0
  87--106, 2017.

\bibitem[Nadeem et~al.(2020)Nadeem, Bethke, and Reddy]{nadeem2020stereoset}
Moin Nadeem, Anna Bethke, and Siva Reddy.
\newblock Stereoset: Measuring stereotypical bias in pretrained language
  models.
\newblock \emph{arXiv preprint arXiv:2004.09456}, 2020.

\bibitem[Naik et~al.(2018)Naik, Ravichander, Sadeh, Rose, and
  Neubig]{naik2018stress}
Aakanksha Naik, Abhilasha Ravichander, Norman Sadeh, Carolyn Rose, and Graham
  Neubig.
\newblock Stress test evaluation for natural language inference.
\newblock \emph{arXiv preprint arXiv:1806.00692}, 2018.

\bibitem[Nakkiran et~al.(2020)Nakkiran, Kaplun, Bansal, Yang, Barak, and
  Sutskever]{Nakkiran2020Deep}
Preetum Nakkiran, Gal Kaplun, Yamini Bansal, Tristan Yang, Boaz Barak, and Ilya
  Sutskever.
\newblock Deep double descent: Where bigger models and more data hurt.
\newblock In \emph{International Conference on Learning Representations}, 2020.
\newblock URL \url{https://openreview.net/forum?id=B1g5sA4twr}.

\bibitem[{National Institute for Health and Care Excellence
  (NICE)}(2019)]{nice2019}
{National Institute for Health and Care Excellence (NICE)}.
\newblock Acute kidney injury: prevention, detection and management.
\newblock \emph{NICE Guideline NG148}, 2019.

\bibitem[Neal(1996)]{neal1996priors}
Radford~M Neal.
\newblock Priors for infinite networks.
\newblock In \emph{Bayesian Learning for Neural Networks}, pages 29--53.
  Springer, 1996.

\bibitem[Need and Goldstein(2009)]{Need2009-gwas-fairness}
Anna~C Need and David~B Goldstein.
\newblock Next generation disparities in human genomics: concerns and remedies.
\newblock \emph{Trends Genet.}, 25\penalty0 (11):\penalty0 489--494, November
  2009.

\bibitem[Nestor et~al.(2019)Nestor, McDermott, Boag, Berner, Naumann, Hughes,
  Goldenberg, and Ghassemi]{Nestor2019}
Bret Nestor, Matthew B.~A. McDermott, Willie Boag, Gabriela Berner, Tristan
  Naumann, Michael~C Hughes, Anna Goldenberg, and Marzyeh Ghassemi.
\newblock {Feature Robustness in Non-stationary Health Records: Caveats to
  Deployable Model Performance in Common Clinical Machine Learning Tasks}.
\newblock \emph{Proceedings of Machine Learning Research}, 106:\penalty0 1--23,
  2019.
\newblock URL \url{https://mimic.physionet.org/mimicdata/carevue/
  http://arxiv.org/abs/1908.00690}.

\bibitem[Oakden-Rayner et~al.(2020)Oakden-Rayner, Dunnmon, Carneiro, and
  R{\'e}]{oakden2020hidden}
Luke Oakden-Rayner, Jared Dunnmon, Gustavo Carneiro, and Christopher R{\'e}.
\newblock Hidden stratification causes clinically meaningful failures in
  machine learning for medical imaging.
\newblock In \emph{Proceedings of the ACM Conference on Health, Inference, and
  Learning}, pages 151--159, 2020.

\bibitem[Obermeyer et~al.(2019)Obermeyer, Powers, Vogeli, and
  Mullainathan]{Obermeyer2019}
Ziad Obermeyer, Brian Powers, Christine Vogeli, and Sendhil Mullainathan.
\newblock {Dissecting racial bias in an algorithm used to manage the health of
  populations}.
\newblock \emph{Science}, 366\penalty0 (6464):\penalty0 447--453, oct 2019.
\newblock ISSN 10959203.
\newblock \doi{10.1126/science.aax2342}.

\bibitem[Panigutti et~al.(2020)Panigutti, Perotti, and
  Pedreschi]{Panigutti2020}
Cecilia Panigutti, Alan Perotti, and Dino Pedreschi.
\newblock {Doctor XAI An ontology-based approach to black-box sequential data
  classification explanations}.
\newblock In \emph{FAT* 2020 - Proceedings of the 2020 Conference on Fairness,
  Accountability, and Transparency}, pages 629--639, 2020.
\newblock ISBN 9781450369367.
\newblock \doi{10.1145/3351095.3372855}.
\newblock URL \url{https://doi.org/10.1145/3351095.3372855}.

\bibitem[Peters et~al.(2016)Peters, B\"uhlmann, and
  Meinshausen]{peters2016causal}
Jonas Peters, Peter B\"uhlmann, and Nicolai Meinshausen.
\newblock Causal inference by using invariant prediction: identification and
  confidence intervals.
\newblock \emph{Journal of the Royal Statistical Society: Series B (Statistical
  Methodology)}, 78\penalty0 (5):\penalty0 947--1012, 2016.
\newblock \doi{10.1111/rssb.12167}.
\newblock URL
  \url{https://rss.onlinelibrary.wiley.com/doi/abs/10.1111/rssb.12167}.

\bibitem[Peters et~al.(2018)Peters, Neumann, Iyyer, Gardner, Clark, Lee, and
  Zettlemoyer]{peters2018deep}
Matthew Peters, Mark Neumann, Mohit Iyyer, Matt Gardner, Christopher Clark,
  Kenton Lee, and Luke Zettlemoyer.
\newblock Deep contextualized word representations.
\newblock In \emph{Proceedings of the 2018 Conference of the North American
  Chapter of the Association for Computational Linguistics: Human Language
  Technologies, Volume 1 (Long Papers)}, pages 2227--2237, 2018.

\bibitem[Popejoy and Fullerton(2016)]{Popejoy2016-genomics-fairness}
Alice~B Popejoy and Stephanie~M Fullerton.
\newblock Genomics is failing on diversity.
\newblock \emph{Nature}, 538\penalty0 (7624):\penalty0 161--164, October 2016.

\bibitem[Popescu and Khalilia(2011)]{Popescu2011}
Mihail Popescu and Mohammad Khalilia.
\newblock {Improving disease prediction using ICD-9 ontological features}.
\newblock In \emph{IEEE International Conference on Fuzzy Systems}, pages
  1805--1809, 2011.
\newblock ISBN 9781424473175.
\newblock \doi{10.1109/FUZZY.2011.6007410}.

\bibitem[Price et~al.(2006)Price, Patterson, Plenge, Weinblatt, Shadick, and
  Reich]{Price2006-pca-gwas}
Alkes~L Price, Nick~J Patterson, Robert~M Plenge, Michael~E Weinblatt, Nancy~A
  Shadick, and David Reich.
\newblock Principal components analysis corrects for stratification in
  genome-wide association studies.
\newblock \emph{Nat. Genet.}, 38\penalty0 (8):\penalty0 904--909, August 2006.

\bibitem[Purcell et~al.(2007)Purcell, Neale, Todd-Brown, Thomas, Ferreira,
  Bender, Maller, Sklar, de~Bakker, Daly, and Sham]{Purcell2007-plink}
Shaun Purcell, Benjamin Neale, Kathe Todd-Brown, Lori Thomas, Manuel A~R
  Ferreira, David Bender, Julian Maller, Pamela Sklar, Paul I~W de~Bakker,
  Mark~J Daly, and Pak~C Sham.
\newblock {PLINK}: a tool set for whole-genome association and population-based
  linkage analyses.
\newblock \emph{Am. J. Hum. Genet.}, 81\penalty0 (3):\penalty0 559--575,
  September 2007.

\bibitem[Raghunathan et~al.(2020)Raghunathan, Xie, Yang, Duchi, and
  Liang]{raghunathan2020understanding}
Aditi Raghunathan, Sang~Michael Xie, Fanny Yang, John Duchi, and Percy Liang.
\newblock Understanding and mitigating the tradeoff between robustness and
  accuracy.
\newblock \emph{arXiv preprint arXiv:2002.10716}, 2020.

\bibitem[Rahimi and Recht(2008)]{rahimi2008random}
Ali Rahimi and Benjamin Recht.
\newblock Random features for large-scale kernel machines.
\newblock In \emph{Advances in neural information processing systems}, pages
  1177--1184, 2008.

\bibitem[Ribeiro et~al.(2020)Ribeiro, Wu, Guestrin, and
  Singh]{ribeiro2020beyond}
Marco~Tulio Ribeiro, Tongshuang Wu, Carlos Guestrin, and Sameer Singh.
\newblock Beyond accuracy: Behavioral testing of {NLP} models with
  {C}heck{L}ist.
\newblock In \emph{Proceedings of the 58th Annual Meeting of the Association
  for Computational Linguistics}, pages 4902--4912, Online, July 2020.
  Association for Computational Linguistics.
\newblock \doi{10.18653/v1/2020.acl-main.442}.
\newblock URL \url{https://www.aclweb.org/anthology/2020.acl-main.442}.

\bibitem[Rivera et~al.(2020)Rivera, Liu, Chan, Denniston, and
  Calvert]{rivera2020guidelines}
Samantha~Cruz Rivera, Xiaoxuan Liu, An-Wen Chan, Alastair~K Denniston, and
  Melanie~J Calvert.
\newblock Guidelines for clinical trial protocols for interventions involving
  artificial intelligence: the spirit-ai extension.
\newblock \emph{bmj}, 370, 2020.

\bibitem[Ross et~al.(2017)Ross, Hughes, and Doshi-Velez]{ross2017right}
Andrew~Slavin Ross, Michael~C Hughes, and Finale Doshi-Velez.
\newblock Right for the right reasons: training differentiable models by
  constraining their explanations.
\newblock In \emph{Proceedings of the 26th International Joint Conference on
  Artificial Intelligence}, pages 2662--2670. AAAI Press, 2017.

\bibitem[Rudinger et~al.(2018)Rudinger, Naradowsky, Leonard, and
  Van~Durme]{rudinger2018gender}
Rachel Rudinger, Jason Naradowsky, Brian Leonard, and Benjamin Van~Durme.
\newblock Gender bias in coreference resolution.
\newblock In \emph{Proceedings of the 2018 Conference of the North American
  Chapter of the Association for Computational Linguistics: Human Language
  Technologies, Volume 2 (Short Papers)}, pages 8--14, 2018.

\bibitem[Sch{\"o}lkopf(2019)]{scholkopf2019causality}
Bernhard Sch{\"o}lkopf.
\newblock Causality for machine learning.
\newblock \emph{arXiv preprint arXiv:1911.10500}, 2019.

\bibitem[Semenova et~al.(2019)Semenova, Rudin, and Parr]{semenova2019study}
Lesia Semenova, Cynthia Rudin, and Ronald Parr.
\newblock A study in rashomon curves and volumes: A new perspective on
  generalization and model simplicity in machine learning.
\newblock \emph{arXiv preprint arXiv:1908.01755}, 2019.

\bibitem[Slatkin(2008)]{Slatkin-LD}
Montgomery Slatkin.
\newblock Linkage disequilibrium — understanding the evolutionary past and
  mapping the medical future.
\newblock \emph{Nature Reviews Genetics}, 9:\penalty0 477--485, 2008.

\bibitem[Snoek et~al.(2019)Snoek, Ovadia, Fertig, Lakshminarayanan, Nowozin,
  Sculley, Dillon, Ren, and Nado]{snoek2019can}
Jasper Snoek, Yaniv Ovadia, Emily Fertig, Balaji Lakshminarayanan, Sebastian
  Nowozin, D~Sculley, Joshua Dillon, Jie Ren, and Zachary Nado.
\newblock Can you trust your model's uncertainty? evaluating predictive
  uncertainty under dataset shift.
\newblock In \emph{Advances in Neural Information Processing Systems}, pages
  13969--13980, 2019.

\bibitem[Sudlow et~al.(2015)Sudlow, Gallacher, Allen, Beral, Burton, Danesh,
  Downey, Elliott, Green, Landray, Liu, Matthews, Ong, Pell, Silman, Young,
  Sprosen, Peakman, and Collins]{Sudlow2015-ukb}
Cathie Sudlow, John Gallacher, Naomi Allen, Valerie Beral, Paul Burton, John
  Danesh, Paul Downey, Paul Elliott, Jane Green, Martin Landray, Bette Liu,
  Paul Matthews, Giok Ong, Jill Pell, Alan Silman, Alan Young, Tim Sprosen, Tim
  Peakman, and Rory Collins.
\newblock {UK} biobank: an open access resource for identifying the causes of a
  wide range of complex diseases of middle and old age.
\newblock \emph{PLoS Med.}, 12\penalty0 (3):\penalty0 e1001779, March 2015.

\bibitem[Sun et~al.(2017)Sun, Shrivastava, Singh, and Gupta]{sun2017revisiting}
Chen Sun, Abhinav Shrivastava, Saurabh Singh, and Abhinav Gupta.
\newblock Revisiting unreasonable effectiveness of data in deep learning era.
\newblock In \emph{Proceedings of the IEEE international conference on computer
  vision}, pages 843--852, 2017.

\bibitem[Szegedy et~al.(2017)Szegedy, Ioffe, Vanhoucke, and
  Alemi]{szegedy2017inception}
Christian Szegedy, Sergey Ioffe, Vincent Vanhoucke, and Alexander~A Alemi.
\newblock Inception-v4, inception-resnet and the impact of residual connections
  on learning.
\newblock In \emph{Thirty-first AAAI conference on artificial intelligence},
  2017.

\bibitem[Taori et~al.(2020)Taori, Dave, Shankar, Carlini, Recht, and
  Schmidt]{taori2020measuring}
Rohan Taori, Achal Dave, Vaishaal Shankar, Nicholas Carlini, Benjamin Recht,
  and Ludwig Schmidt.
\newblock Measuring robustness to natural distribution shifts in image
  classification.
\newblock \emph{arXiv preprint arXiv:2007.00644}, 2020.

\bibitem[Ting et~al.(2017)Ting, Cheung, Lim, Tan, Quang, Gan, Hamzah,
  Garcia-Franco, San~Yeo, Lee, et~al.]{ting2017development}
Daniel Shu~Wei Ting, Carol Yim-Lui Cheung, Gilbert Lim, Gavin Siew~Wei Tan,
  Nguyen~D Quang, Alfred Gan, Haslina Hamzah, Renata Garcia-Franco, Ian~Yew
  San~Yeo, Shu~Yen Lee, et~al.
\newblock Development and validation of a deep learning system for diabetic
  retinopathy and related eye diseases using retinal images from multiethnic
  populations with diabetes.
\newblock \emph{Jama}, 318\penalty0 (22):\penalty0 2211--2223, 2017.

\bibitem[Toma{\v{s}}ev et~al.(2019{\natexlab{a}})Toma{\v{s}}ev, Glorot, Rae,
  Zielinski, Askham, Saraiva, Mottram, Meyer, Ravuri, Protsyuk, Connell,
  Hughes, Karthikesalingam, Cornebise, Montgomery, Rees, Laing, Baker,
  Peterson, Reeves, Hassabis, King, Suleyman, Back, Nielson, Ledsam, and
  Mohamed]{Tomasev2019}
Nenad Toma{\v{s}}ev, Xavier Glorot, Jack~W Rae, Michal Zielinski, Harry Askham,
  Andre Saraiva, Anne Mottram, Clemens Meyer, Suman Ravuri, Ivan Protsyuk,
  Alistair Connell, C{\'{i}}an~O Hughes, Alan Karthikesalingam, Julien
  Cornebise, Hugh Montgomery, Geraint Rees, Chris Laing, Clifton~R Baker, Kelly
  Peterson, Ruth Reeves, Demis Hassabis, Dominic King, Mustafa Suleyman, Trevor
  Back, Christopher Nielson, Joseph~R Ledsam, and Shakir Mohamed.
\newblock {A clinically applicable approach to continuous prediction of future
  acute kidney injury}.
\newblock \emph{Nature}, 572\penalty0 (7767):\penalty0 116--119, aug
  2019{\natexlab{a}}.
\newblock ISSN 0028-0836.
\newblock \doi{10.1038/s41586-019-1390-1}.

\bibitem[Toma{\v{s}}ev et~al.(2019{\natexlab{b}})Toma{\v{s}}ev, Glorot, Rae,
  Zielinski, Askham, Saraiva, Mottram, Meyer, Ravuri, Protsyuk, Connell,
  Hugues, Kathikesalingam, Cornebise, Montgomery, Rees, Laing, Baker, Peterson,
  Reeves, Hassabis, King, Suleyman, Back, Nielson, Ledsam, and
  Mohamed]{Tomasev2019a}
Nenad Toma{\v{s}}ev, Xavier Glorot, Jack~W. Rae, Michal Zielinski, Harry
  Askham, Andre Saraiva, Anne Mottram, Clemens Meyer, Suman Ravuri, Ivan
  Protsyuk, Alistair Connell, Cian~O. Hugues, Alan Kathikesalingam, Julien
  Cornebise, Hugh Montgomery, Geraint Rees, Chris Laing, Clifton~R. Baker,
  Kelly Peterson, Ruth Reeves, Demis Hassabis, Dominic King, Mustafa Suleyman,
  Trevor Back, Christopher Nielson, Joseph~R. Ledsam, and Shakir Mohamed.
\newblock {Developing Deep Learning Continuous Risk Models for Early Adverse
  Event Prediction in Electronic Health Records: an AKI Case Study}.
\newblock \emph{PROTOCOL available at Protocol Exchange}, version 1, jul
  2019{\natexlab{b}}.
\newblock \doi{10.21203/RS.2.10083/V1}.

\bibitem[Vaswani et~al.(2017)Vaswani, Shazeer, Parmar, Uszkoreit, Jones, Gomez,
  Kaiser, and Polosukhin]{vaswani2017attention}
Ashish Vaswani, Noam Shazeer, Niki Parmar, Jakob Uszkoreit, Llion Jones,
  Aidan~N Gomez, {\L}ukasz Kaiser, and Illia Polosukhin.
\newblock Attention is all you need.
\newblock In \emph{Advances in neural information processing systems}, pages
  5998--6008, 2017.

\bibitem[Vilhj{\'a}lmsson et~al.(2015)Vilhj{\'a}lmsson, Yang, Finucane, Gusev,
  Lindstr{\"o}m, Ripke, Genovese, Loh, Bhatia, Do, Hayeck, Won, {Schizophrenia
  Working Group of the Psychiatric Genomics Consortium, Discovery, Biology, and
  Risk of Inherited Variants in Breast Cancer (DRIVE) study}, Kathiresan, Pato,
  Pato, Tamimi, Stahl, Zaitlen, Pasaniuc, Belbin, Kenny, Schierup, De~Jager,
  Patsopoulos, McCarroll, Daly, Purcell, Chasman, Neale, Goddard, Visscher,
  Kraft, Patterson, and Price]{Vilhjalmsson2015-ldpred}
Bjarni~J Vilhj{\'a}lmsson, Jian Yang, Hilary~K Finucane, Alexander Gusev, Sara
  Lindstr{\"o}m, Stephan Ripke, Giulio Genovese, Po-Ru Loh, Gaurav Bhatia, Ron
  Do, Tristan Hayeck, Hong-Hee Won, {Schizophrenia Working Group of the
  Psychiatric Genomics Consortium, Discovery, Biology, and Risk of Inherited
  Variants in Breast Cancer (DRIVE) study}, Sekar Kathiresan, Michele Pato,
  Carlos Pato, Rulla Tamimi, Eli Stahl, Noah Zaitlen, Bogdan Pasaniuc, Gillian
  Belbin, Eimear~E Kenny, Mikkel~H Schierup, Philip De~Jager, Nikolaos~A
  Patsopoulos, Steve McCarroll, Mark Daly, Shaun Purcell, Daniel Chasman,
  Benjamin Neale, Michael Goddard, Peter~M Visscher, Peter Kraft, Nick
  Patterson, and Alkes~L Price.
\newblock Modeling linkage disequilibrium increases accuracy of polygenic risk
  scores.
\newblock \emph{Am. J. Hum. Genet.}, 97\penalty0 (4):\penalty0 576--592,
  October 2015.

\bibitem[Wang et~al.(2018)Wang, Singh, Michael, Hill, Levy, and
  Bowman]{wang2018glue}
Alex Wang, Amanpreet Singh, Julian Michael, Felix Hill, Omer Levy, and Samuel
  Bowman.
\newblock Glue: A multi-task benchmark and analysis platform for natural
  language understanding.
\newblock In \emph{Proceedings of the 2018 EMNLP Workshop BlackboxNLP:
  Analyzing and Interpreting Neural Networks for NLP}, pages 353--355, 2018.

\bibitem[Wang et~al.(2019)Wang, Ge, Lipton, and Xing]{wang2019learning}
Haohan Wang, Songwei Ge, Zachary Lipton, and Eric~P Xing.
\newblock Learning robust global representations by penalizing local predictive
  power.
\newblock In \emph{Advances in Neural Information Processing Systems}, pages
  10506--10518, 2019.

\bibitem[Wang et~al.(2020)Wang, Wu, Huang, and Xing]{wang2020high}
Haohan Wang, Xindi Wu, Zeyi Huang, and Eric~P Xing.
\newblock High-frequency component helps explain the generalization of
  convolutional neural networks.
\newblock In \emph{Proceedings of the IEEE/CVF Conference on Computer Vision
  and Pattern Recognition}, pages 8684--8694, 2020.

\bibitem[Webster et~al.(2020)Webster, Wang, Tenney, Beutel, Pitler, Pavlick,
  Chen, and Petrov]{webster2020measuring}
Kellie Webster, Xuezhi Wang, Ian Tenney, Alex Beutel, Emily Pitler, Ellie
  Pavlick, Jilin Chen, and Slav Petrov.
\newblock Measuring and reducing gendered correlations in pre-trained models.
\newblock \emph{arXiv preprint arXiv:2010.06032}, 2020.

\bibitem[Wenzel et~al.(2020)Wenzel, Snoek, Tran, and
  Jenatton]{wenzel2020hyperparameter}
Florian Wenzel, Jasper Snoek, Dustin Tran, and Rodolphe Jenatton.
\newblock Hyperparameter ensembles for robustness and uncertainty
  quantification.
\newblock \emph{arXiv preprint arXiv:2006.13570}, 2020.

\bibitem[Williams et~al.(2018)Williams, Nangia, and Bowman]{williams2018broad}
Adina Williams, Nikita Nangia, and Samuel Bowman.
\newblock A broad-coverage challenge corpus for sentence understanding through
  inference.
\newblock In \emph{Proceedings of the 2018 Conference of the North American
  Chapter of the Association for Computational Linguistics: Human Language
  Technologies, Volume 1 (Long Papers)}, pages 1112--1122, 2018.

\bibitem[Wilson and Izmailov(2020)]{wilson2020bayesian}
Andrew~Gordon Wilson and Pavel Izmailov.
\newblock Bayesian deep learning and a probabilistic perspective of
  generalization.
\newblock \emph{arXiv preprint arXiv:2002.08791}, 2020.

\bibitem[Winkler et~al.(2019)Winkler, Fink, Toberer, Enk, Deinlein,
  Hofmann-Wellenhof, Thomas, Lallas, Blum, Stolz,
  et~al.]{winkler2019association}
Julia~K Winkler, Christine Fink, Ferdinand Toberer, Alexander Enk, Teresa
  Deinlein, Rainer Hofmann-Wellenhof, Luc Thomas, Aimilios Lallas, Andreas
  Blum, Wilhelm Stolz, et~al.
\newblock Association between surgical skin markings in dermoscopic images and
  diagnostic performance of a deep learning convolutional neural network for
  melanoma recognition.
\newblock \emph{JAMA dermatology}, 155\penalty0 (10):\penalty0 1135--1141,
  2019.

\bibitem[Wray et~al.(2007)Wray, Goddard, and Visscher]{Wray2007-prs}
Naomi~R Wray, Michael~E Goddard, and Peter~M Visscher.
\newblock Prediction of individual genetic risk to disease from genome-wide
  association studies.
\newblock \emph{Genome Res.}, 17\penalty0 (10):\penalty0 1520--1528, October
  2007.

\bibitem[Yang et~al.(2019)Yang, Dai, Yang, Carbonell, Salakhutdinov, and
  Le]{yang2019xlnet}
Zhilin Yang, Zihang Dai, Yiming Yang, Jaime Carbonell, Russ~R Salakhutdinov,
  and Quoc~V Le.
\newblock Xlnet: Generalized autoregressive pretraining for language
  understanding.
\newblock In \emph{Advances in neural information processing systems}, pages
  5753--5763, 2019.

\bibitem[Yin et~al.(2019)Yin, Lopes, Shlens, Cubuk, and Gilmer]{yin2019fourier}
Dong Yin, Raphael~Gontijo Lopes, Jon Shlens, Ekin~Dogus Cubuk, and Justin
  Gilmer.
\newblock A fourier perspective on model robustness in computer vision.
\newblock In \emph{Advances in Neural Information Processing Systems}, pages
  13255--13265, 2019.

\bibitem[Yu et~al.(2013)]{yu2013stability}
Bin Yu et~al.
\newblock Stability.
\newblock \emph{Bernoulli}, 19\penalty0 (4):\penalty0 1484--1500, 2013.

\bibitem[Zellers et~al.(2018)Zellers, Bisk, Schwartz, and
  Choi]{zellers2018swag}
Rowan Zellers, Yonatan Bisk, Roy Schwartz, and Yejin Choi.
\newblock Swag: A large-scale adversarial dataset for grounded commonsense
  inference.
\newblock \emph{arXiv preprint arXiv:1808.05326}, 2018.

\bibitem[Zellers et~al.(2019)Zellers, Holtzman, Bisk, Farhadi, and
  Choi]{zellers2019hellaswag}
Rowan Zellers, Ari Holtzman, Yonatan Bisk, Ali Farhadi, and Yejin Choi.
\newblock Hellaswag: Can a machine really finish your sentence?
\newblock In \emph{Proceedings of the 57th Annual Meeting of the Association
  for Computational Linguistics}, pages 4791--4800, 2019.

\bibitem[Zhao et~al.(2018)Zhao, Wang, Yatskar, Ordonez, and
  Chang]{zhao2018gender}
Jieyu Zhao, Tianlu Wang, Mark Yatskar, Vicente Ordonez, and Kai-Wei Chang.
\newblock Gender bias in coreference resolution: Evaluation and debiasing
  methods.
\newblock In \emph{Proceedings of the 2018 Conference of the North American
  Chapter of the Association for Computational Linguistics: Human Language
  Technologies, Volume 2 (Short Papers)}, pages 15--20, 2018.

\bibitem[Zhou et~al.(2020)Zhou, Nie, Tan, and Bansal]{zhou2020curse}
Xiang Zhou, Yixin Nie, Hao Tan, and Mohit Bansal.
\newblock The curse of performance instability in analysis datasets:
  Consequences, source, and suggestions.
\newblock \emph{arXiv preprint arXiv:2004.13606}, 2020.

\end{thebibliography}

\newpage
\appendix

\section{Computer Vision: Marginaliztaion versus Model Selection}
\label{sec:vision appendix}
\begin{figure}[ht]
    \centering
    \includegraphics[width=\textwidth]{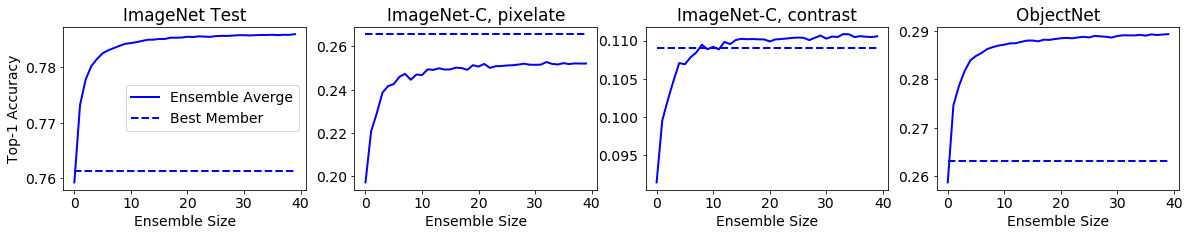}
    \caption{\textbf{Comparison of performance of the ``best'' ensemble member from an ensemble of 50 ResNet-50 predictor (dashed) against the average performance from averaging the predictions of differently-sized subsets of the ensemble (solid).} ImageNet Test is an iid evaluation; the other three panels show stress tests. See main text for full description.
    Iid performance on ImageNet improves as ensemble size increases, and this is associated with correlated improvements in stress tests.
    The larger the variability in stress test performance within the ensemble, the larger the ensemble needs to be to out-perform the best single ensemble member.
    In some cases, the ensemble average never out-performs the best single model.}
    \label{fig:ensemble}
\end{figure}

In the discussion in the main text, we argue that marginalization may not be the best response to underspecification when the goal is to obtain predictors that encode the ``right'' structure for a given application.
We suggest instead that model selection may be a more fruitful approaach here.
This is because, by the nature of underspecification, some predictors returned by the pipeline will exhibit worse behavior in deployment domains than others, so averaging them together does not guarantee that the ensemble average will out-perform the best member.
Notably, this represents a departure from the argument made in favor of marginalization for improving iid performance: in the training domain, all of the models in the near-optimal equivalence class $\mathcal F^*$ (recall this definition from Section 2 of the main text) contain a ``right'' answer for iid generalization, so one would expect that averaging them could only lead to improvements.

In this section, we provide some empirical support for this argument.
Broadly, there is an interplay between iid performance and performance on stress tests that can make marginalization beneficial, but when there is large variability in stress test performance across an ensemble, selecting the best single model can out-perform large ensemble averages.

In Figure~\ref{fig:ensemble}, we show a comparison between performance of individual ensemble members and ensemble averages on several test sets.
We calculate these metrics with respect to the ensemble of 50 ResNet-50 models used to produce the result in the main text.
The dashed line shows the performance of the best model from this ensemble, while the solid line shows the average performance from marginalizing across differently-sized subsets of models in this ensemble.
The ImageNet test set is the iid evaluation, while the other test sets are from the ImageNet-C and ObjectNet benchmarks.
As expected, performance on the ImageNet test set improves substantially as more ensemble members are averaged together.
This translates to correlated performance improvements on stress test benchmarks, which is a well-known phenomenon in the image robustness literature \citep[see, e.g.][]{taori2020measuring,djolonga2020robustness}.
Interestingly, however, it takes marginalizing across a larger subset of models to surpass the performance of the best predictor on stress tests compared to the iid evaluation.
In particular, the higher the variance of performance across the ensemble, the more predictors need to be averaged to beat the surpass single model.
In the case of the pixealate task, the full ensemble of 50 models is never able to surpass the best single model.

\section{Natural Language Processing: Analysis of Static Embeddings}
\label{sec:nlp appendix}
In the main text, we showed that underspecification plays a key role in shortcut learning in BERT-based NLP models.
However, highly parameterized models pre-date this approach, and here we provide a supplementary analysis suggesting that underspecification is also present in static word embeddings such as word2vec~\cite{mikolov2013efficient}.
Here, we examine stereotypical associations with respect to demographic attributes like race, gender, and age, which have been studied in the past \citep{bolukbasi2016man}.

We train twenty different 500-dimensional word2vec models~\cite{mikolov2013efficient} on large news and wikipedia datasets using the \texttt{demo-train-big-model-v1.sh} script from the canonical word2vec repository,\footnote{Canonical codebase is \url{https://code.google.com/archive/p/word2vec/}; a GitHub export of this repository is available at \url{https://github.com/tmikolov/word2vec}.} varying only the random seeds. These models obtain very consistent performance on a word analogy task, scoring between 76.2\% and 76.7\%.

As a stress test, we apply the Word Embedding Association Test, which quantifies the extent to which these associations are encoded by a given set of embeddings~\cite{caliskan2017semantics}.
Specifically, the WEAT score measures the relative similarity of two sets of \textit{target} words (e.g., types of flowers, types of insects) to two sets of \textit{attribute} words (e.g., pleasant words, unpleansant words).  
Let $X$ and $Y$ be the sets of target words, and $A$ and $B$ the sets of attribute words.  The test statistic is then: 
\begin{align*}
s(X, Y, A, B) = \sum_{x\in X} s(x, A, B) - \sum_{y\in Y} s(y, A, B) 
\intertext{where}
s(w, A, B) = \text{mean}_{a\in A}\text{cos}({w\vphantom{b}},{a\vphantom{b}}) - \text{mean}_{b\in B}\text{cos}({w\vphantom{b}}, {b})
\end{align*}
and $\cos({a\vphantom{b}},b)$ is the cosine distance between two word vectors.
This score is then normalized by the standard deviation of $s(w, A, B)$ for all $w$ in $X \cup Y$.
If the score is closer to zero, the relative similarity difference (i.e., bias) is considered to be smaller. 
\citet{caliskan2017semantics} provide a number of wordsets to probe biases along socially salient axes of gender, race, disability, and age. 

\begin{figure}
    \centering
    \includegraphics[width=.8\textwidth]{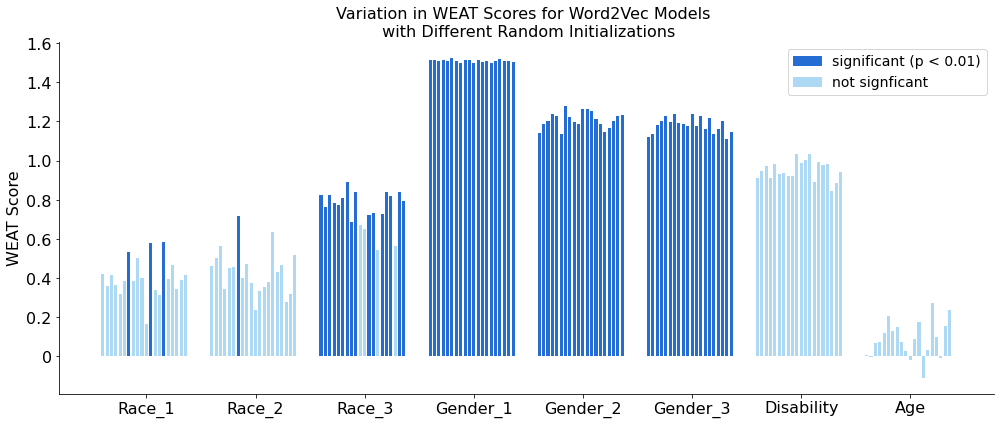}
    \caption{\textbf{Static word embeddings also show evidence of stereotype-aligned underspecification.}
    Word Embedding Association Test (WEAT) scores across twenty word2vec models. Each group corresponds to a specific association test, and each bar corresponds to the score on the test for a specific word2vec model.}
    \label{fig:nlp_weat}
\end{figure}

For each model and test, we compute statistical significance using a permutation test that compares the observed score against the score obtained under a random shuffling of target words.
As shown in \autoref{fig:nlp_weat}, we find strong and consistent gender associations, but observe substantial variation on the three tests related to associations with race: in many cases, whether an association is statistically significant depends on the random seed.
Finally, we note that the particular axes along which we observe sensitivity depends on the dataset used to train the embeddings, and could vary on embeddings trained on other corpora \cite{babaeianjelodar2020quantifying}.

\section{Clinical Prediction with EHR: Additional Details and Supplementary Ablation Experiment}
\label{sec:ehr appendix}

This section provides additional details and results for the analysis of the model in \cite{Tomasev2019} performed in the main text.
In particular, we provide some descriptive statistics regarding AKI prevalence, and additional summaries of model performance across different time slices and dataset shifts.

\subsection{Lab Order Patterns and Time of Day}
In the main text, we investigate how reliant predictors can be on signals related to the timing and composition of lab tests.
Here, we show some descriptive statistics for how these tests tend to be distributed in time, and some patterns that emerge as a result.

Table~\ref{tab:ehr_time_of_day_prevalence} shows patterns of AKI prevalence and creatinine sampling.
Even though AKI prevalence is largely constant across time windows, creatinine is sampled far more frequently in between 12am and 6am.
Thus, when creatinine samples are taken in other time windows, the prevalence of AKI conditional on that sample being taken is higher.

\begin{figure}[ht]
    \centering
    \includegraphics[width=0.5\textwidth]{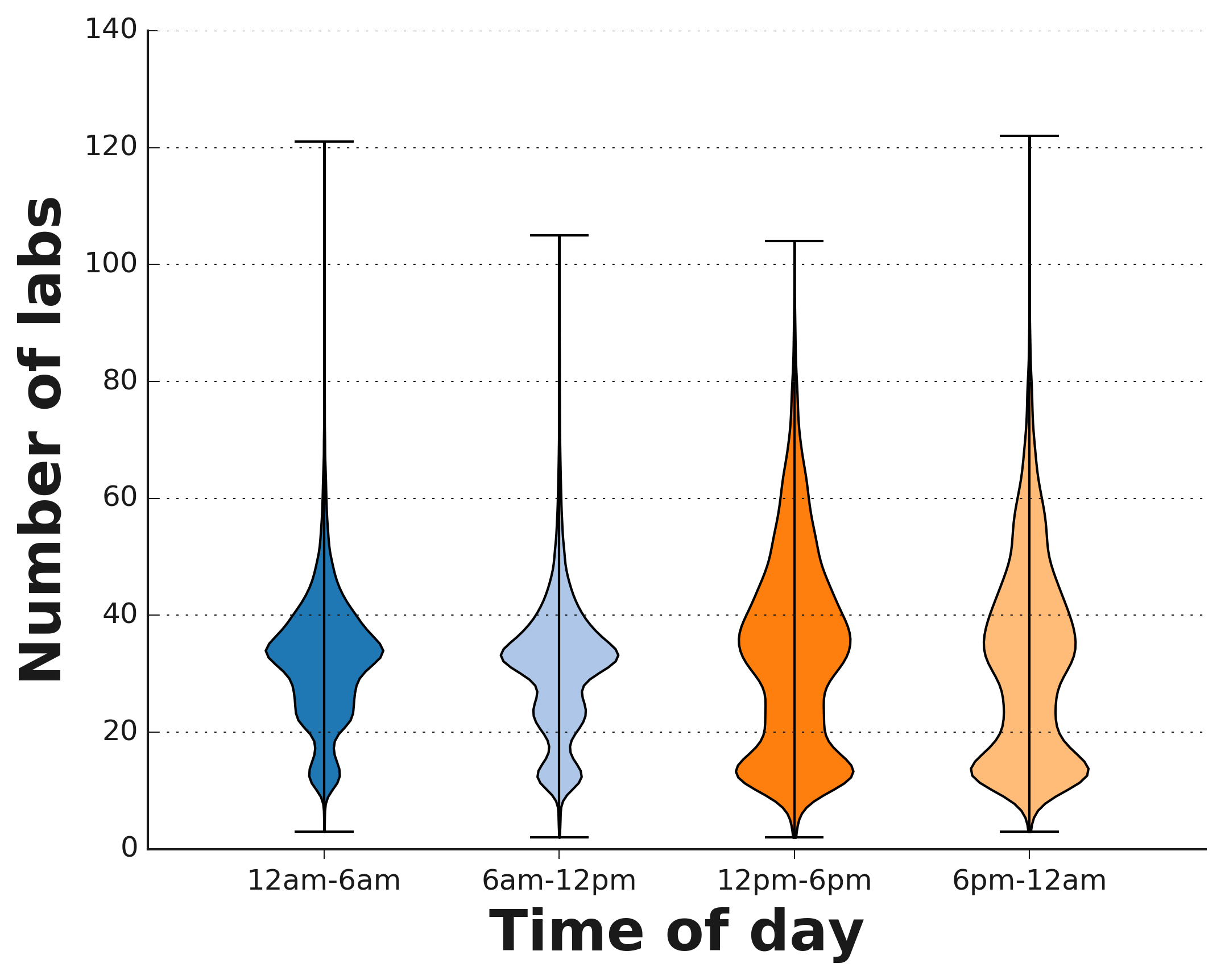}
    \caption{Distribution of number of lab values observed on average across 100000 time steps (random sample in the test set), per time of day. Each violin plot represent a time of day while the y-axis represents the number of lab values observed.}
    \label{fig:ehr_num_labs_tod}
\end{figure}

Figure~\ref{fig:ehr_num_labs_tod} shows the distributions of number of labs taken at different times of day.
The distribution of labs in the first two time buckets is clearly distinct from the distributions in the second two time buckets.


\begin{table}[ht]
\caption{\textbf{Patterns of creatinine sampling can induce a spurious relationship between time of day and AKI.} Prevalence of AKI is stable across times of day in the test set (test set prevalence is $2.269\%$), but creatinine samples are taken more frequently in the first two time buckets.
    As a result, \emph{conditional on a sample being taken}, AKI prevalence is higher in the latter two time buckets.}
\centering
\begin{tabular}{p{5cm}|r r r r}
\textbf{Metric} & \textbf{12am-6am} & \textbf{6am-12pm} & \textbf{12pm-6pm} & \textbf{6pm-12am} \\
\hline
\textbf{Prevalence of AKI (\%)} & 2.242 & 2.153 & 2.287 & 2.396\\
\textbf{Creatinine samples} & 332743 & 320068 & 89379 & 67765 \\
\textbf{Creatinine samples (\%)} & 3.570 & 3.433 & 0.959 & 0.727 \\
\textbf{Prevalence of AKI (\%) in creatinine samples} & 6.236 & 5.425 & 8.824 & 9.154\\
\end{tabular}
\label{tab:ehr_time_of_day_prevalence}
\end{table}

\subsection{Details of Predictor Performance on Intervened Data}

Here, we perform a stratified evaluation of model performance across different time buckets in the standard and intervened test data.
This analysis is repeated for each of the 15 models trained, and acoss the Shift and Shift+Labs intevened datasets described in the main text.
Table \ref{table:ehr_tod_sensitivity} displays model performance on each model instance for the test set as well as for time windows where creatinine samples were taken.

\begin{landscape}
\begin{table}[!ht]
    \caption{\textbf{Model sensitivity per time of day:} Normalized PRAUC on the test set, per time of day, and per time of day for creatinine samples for each model instance when the data is not perturbed (`Test'), when time of day is shifted (`Shift') and when time of day is shifted and only CHEM-7 labs are considered for creatinine samples (`Shift+Labs'). `Diff.' refers to the maximum difference in value between instances.}
\begin{small}
\begin{center}
    \begin{tabular}{l|r r r r r| r r r r r| r r r r r| r}
cell &  \multicolumn{5}{|c|}{\textbf{SRU}} &  \multicolumn{5}{|c|}{\textbf{UGRNN}}  & \multicolumn{5}{|c|}{\textbf{LSTM}} &  \\
 \hline
seed	&\textbf{1}	&\textbf{2}	&\textbf{3}	&\textbf{4}	&\textbf{5}	& \textbf{1}	&\textbf{2}	&\textbf{3}	&\textbf{4}	&\textbf{5} &\textbf{1}	&\textbf{2}	&\textbf{3}	&\textbf{4}	&\textbf{5} & \textbf{Diff.}	\\ \hline
\textbf{Test}	&	&	&	&	&	&	&	&	&	&	&	&	&	&	&	&	\\ \hline
12am - 6am	&34.69	&34.42	&34.64	&34.99	&34.74	&34.61	&34.88	&34.46	&34.86	&34.62	&35.43	&35.55	&35.42	&35.63	&35.51	&1.21	\\
6am - 12pm	&34.26	&34	&34.14	&34.32	&34.26	&33.98	&34.39	&34.05	&34.3	&34.01	&34.86	&35.06	&34.91	&35.12	&34.94	&1.14	\\
12pm - 6pm	&36.51	&36.26	&36.43	&36.67	&36.55	&36.13	&36.57	&36.24	&36.57	&36.25	&37.05	&37.28	&37.07	&37.35	&37.11	&1.22	\\
6pm - 12am	&37.22	&37	&37.17	&37.41	&37.24	&36.97	&37.34	&36.93	&37.32	&36.99	&37.78	&38.03	&37.82	&38.07	&37.89	&1.14	\\ \hline
\textbf{Creatinine samples}	&	&	&	&	&	&	&	&	&	&	&	&	&	&	&	&	\\
12am - 6am	&36.51	&36.32	&36.66	&37.06	&36.57	&36.52	&36.92	&36.31	&36.82	&36.55	&37.43	&37.64	&37.5	&37.57	&37.48	&1.33	\\
6am - 12pm	&34.4	&34.53	&34.24	&34.75	&34.55	&34.61	&34.85	&34.37	&34.59	&34.58	&35.42	&35.52	&35.28	&35.48	&35.41	&1.27	\\
12pm - 6pm	&42.68	&42.5	&42.68	&42.34	&42.23	&42.17	&42.49	&41.64	&42.44	&42.25	&43	&43.2	&42.93	&43.77	&43.38	&2.13	\\
6pm - 12am	&41.81	&41.86	&41.72	&42.17	&41.91	&41.87	&42.52	&41.05	&41.7	&41.75	&42.58	&42.79	&42.86	&42.84	&43.19	&2.14	\\ \hline
\textbf{Shift}	&	&	&	&	&	&	&	&	&	&	&	&	&	&	&	&	\\ \hline
\textbf{Population}	&34.93	&34.9	&34.74	&35.08	&34.93	&35.08	&35.46	&35.11	&35.53	&35.13	&35.97	&36.16	&35.93	&36.21	&35.82	&1.47	\\ \hline
\textbf{Test}	&	&	&	&	&	&	&	&	&	&	&	&	&	&	&	&	\\
12am - 6am	&33.75	&33.63	&33.67	&33.79	&33.52	&34.06	&34.51	&34.05	&34.55	&34.09	&35.03	&35.11	&34.82	&35.16	&34.85	&1.64	\\
6am - 12pm	&33.55	&33.54	&33.32	&33.74	&33.62	&33.7	&33.89	&33.7	&34	&33.62	&34.44	&34.62	&34.51	&34.73	&34.3	&1.4	\\
12pm - 6pm	&35.72	&35.78	&35.57	&35.94	&35.88	&35.79	&36.22	&35.92	&36.28	&35.97	&36.74	&36.97	&36.71	&37.03	&36.56	&1.45	\\
6pm - 12am	&36.49	&36.45	&36.28	&36.71	&36.45	&36.52	&36.99	&36.54	&37.04	&36.61	&37.45	&37.68	&37.41	&37.69	&37.32	&1.41	\\ \hline
\textbf{Creatinine samples}	&	&	&	&	&	&	&	&	&	&	&	&	&	&	&	&	\\
12am - 6am	&36.15	&36.1	&36.22	&36.66	&36.12	&36.18	&36.72	&36.12	&36.63	&36.24	&37.2	&37.43	&37.12	&37.41	&37.2	&1.33	\\
6am - 12pm	&33.72	&33.97	&33.64	&34.06	&33.97	&34.44	&34.5	&34.05	&34.39	&34.23	&35.03	&35.14	&34.97	&35.15	&34.97	&1.51	\\
12pm - 6pm	&41.81	&41.98	&41.99	&41.68	&41.55	&41.73	&42.2	&41.03	&42.27	&41.95	&42.77	&42.87	&42.67	&43.51	&42.95	&2.48	\\
6pm - 12am	&41	&41.27	&41.1	&41.49	&41.13	&41.27	&42.11	&40.6	&41.31	&41.36	&42.11	&42.33	&42.43	&42.51	&42.7	&2.1	\\ \hline
\textbf{Shift+Labs}	&	&	&	&	&	&	&	&	&	&	&	&	&	&	&	&	\\ \hline
\textbf{Population}	&33.66	&33.86	&33.46	&33.99	&33.76	&34.08	&34.77	&34.22	&34.67	&34.29	&35.16	&35.18	&34.93	&35.21	&34.76	&1.74	\\ \hline
\textbf{Test}	&	&	&	&	&	&	&	&	&	&	&	&	&	&	&	&	\\
12am - 6am	&32.5	&32.62	&32.44	&32.74	&32.29	&32.94	&33.79	&33.19	&33.72	&33.18	&34.25	&34.05	&33.81	&34.2	&33.72	&1.96	\\
6am - 12pm	&32.22	&32.44	&32.09	&32.55	&32.43	&32.76	&33.23	&32.84	&33.07	&32.77	&33.68	&33.72	&33.52	&33.73	&33.34	&1.64	\\
12pm - 6pm	&34.45	&34.71	&34.24	&34.85	&34.78	&34.82	&35.5	&35	&35.39	&35.13	&35.89	&35.98	&35.69	&36.01	&35.49	&1.77	\\
6pm - 12am	&35.27	&35.43	&34.96	&35.68	&35.34	&35.57	&36.33	&35.64	&36.26	&35.86	&36.62	&36.69	&36.42	&36.66	&36.27	&1.73	\\ \hline
\textbf{Creatinine samples}	&	&	&	&	&	&	&	&	&	&	&	&	&	&	&	&	\\
12am - 6am	&34.32	&34.55	&34.58	&35.03	&34.43	&34.56	&35.7	&35.03	&35.45	&34.64	&36.05	&35.99	&35.69	&36.08	&35.73	&1.76	\\
6am - 12pm	&32.58	&33.03	&32.65	&33.1	&32.91	&33.5	&33.85	&33.37	&33.65	&33.45	&34.29	&34.26	&34.03	&34.16	&34.28	&1.71	\\
12pm - 6pm	&40.05	&40.37	&40.1	&40.13	&39.95	&39.97	&40.91	&39.59	&40.85	&40.35	&41.1	&40.85	&40.83	&41.81	&41.03	&2.23	\\
6pm - 12am	&38.91	&39.43	&38.87	&39.53	&39.18	&39.33	&40.63	&39.13	&39.68	&39.7	&40.42	&40.23	&40.35	&40.79	&40.68	&1.92	\\
\end{tabular}
\label{table:ehr_tod_sensitivity}
\end{center}
\end{small}
\end{table}
\end{landscape}

\begin{figure}[!ht]
    \centering
    \includegraphics[width=\textwidth]{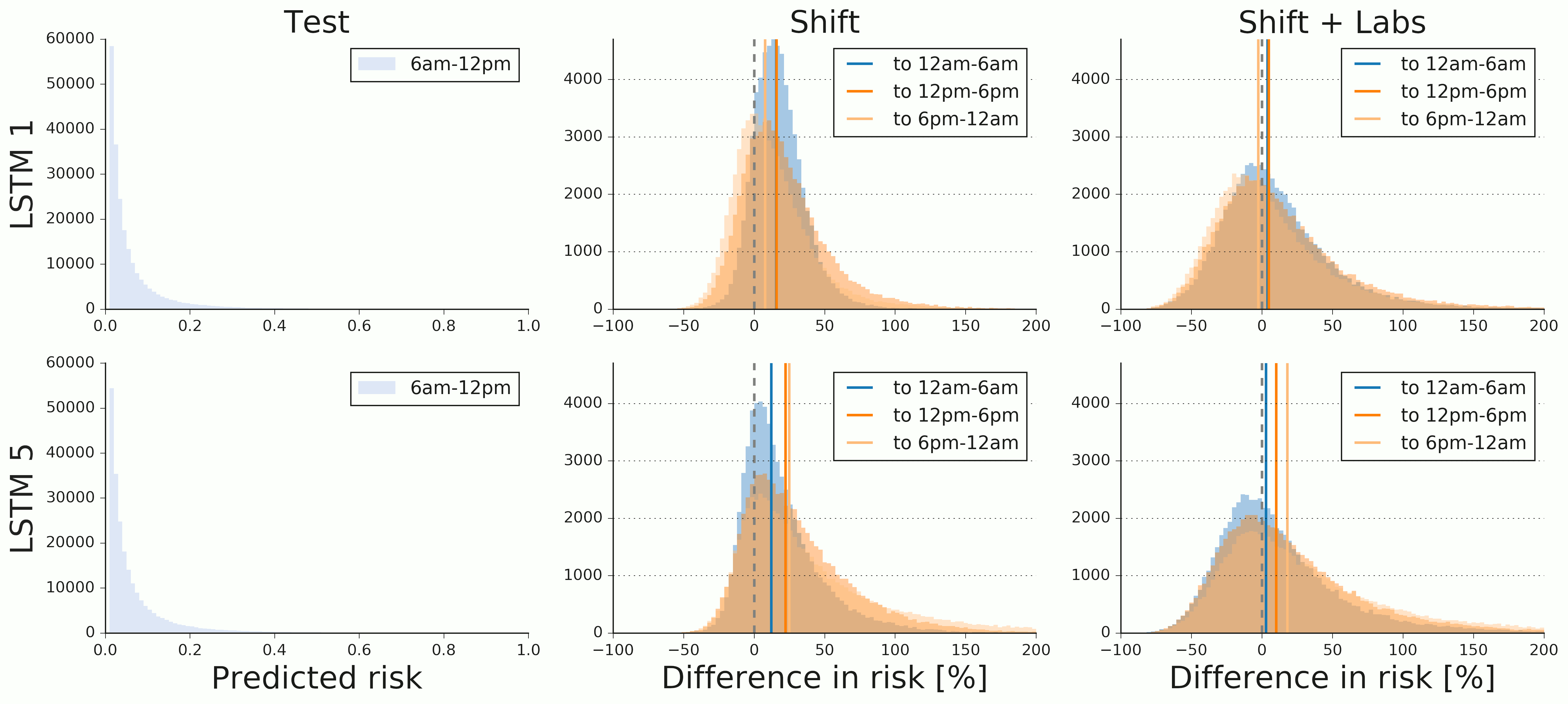}
    \caption{\textbf{Variability in AKI risk predictions from ensemble of RNN models processing electronic health records (EHR)}. Histograms showing showing risk predictions from two models, and changes induced by time of day and lab perturbations.
    Histograms show counts of patient-timepoints where creatinine measurements were taken in the morning (6am-12pm). LSTM 1 and 5 differ only in random seed.
    ``Test'' shows histogram of risk predicted in original test data.
    ``Shift'' and ``Shift + Labs'' show histograms of proportional changes (in \%) $\frac{\text{Perturbed} - \text{Baseline}}{\text{Baseline}}$ induced by the time-shift perturbation and the combined time-shift and lab perturbation, respectively.}
    
    \label{fig:ehr_shift_hist_tod2}
\end{figure}

\begin{figure}[!ht]
    \centering
    \includegraphics[width=\textwidth]{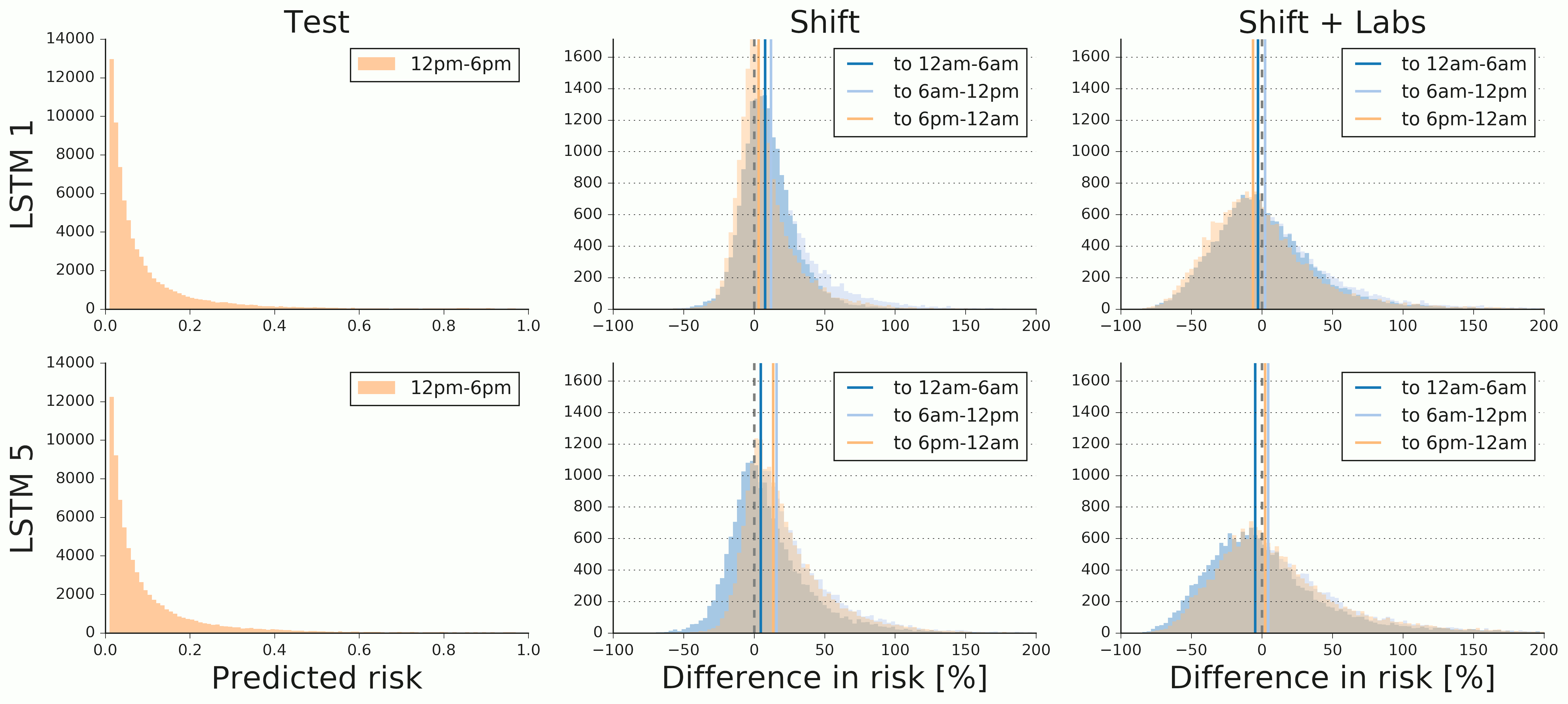}
    \caption{Same as Figure~\ref{fig:ehr_shift_hist_tod2} for the afternoon (12pm-6pm).}
    
    \label{fig:ehr_shift_hist_tod3}
\end{figure}

\begin{figure}[!ht]
    \centering
    \includegraphics[width=\textwidth]{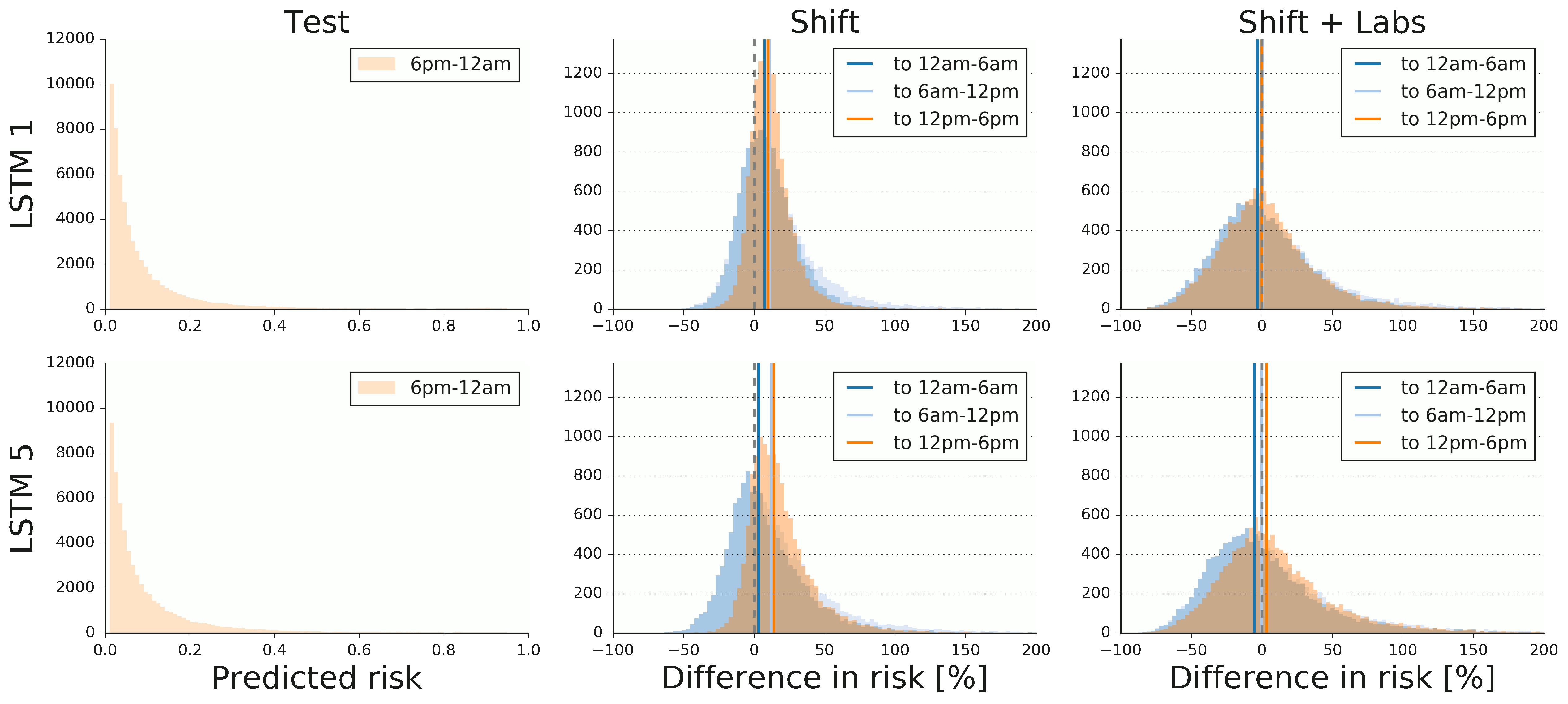}
    \caption{Same as Figure~\ref{fig:ehr_shift_hist_tod2} for the evening (6pm-12am).}
    
    \label{fig:ehr_shift_hist_tod4}
\end{figure}

When perturbing the correlation between AKI label, time of day and number of labs on a per patient basis, we observe a decrease in model performance, as well as a widening of the performance bounds (reported in the main text). This widening of performance bounds displays a differential effect of the shifts on the different models of the ensemble, both on the individual patient timepoints risk (Figures~\ref{fig:ehr_shift_hist_tod2},~\ref{fig:ehr_shift_hist_tod3},~\ref{fig:ehr_shift_hist_tod4}) and on the model's decisions (Table~\ref{tab:flips}). 

\begin{table}[t]
\centering
\caption{\textbf{Flipped decisions under time-shift and lab order composition interventions depend on random seed.} Each cell is number of patient-timepoints at which decisions changed when the time range feature and lab order composition were changed, for patient timepoints with creatinine measured.
``+ to -'' indicates a change from the ``at risk of AKI in next 48 hrs" to ``not at risk''; ``- to +'' indicates the opposite change.
Model 1 and model 2 are LSTM models that differ only in random seed. 
Overlap indicates the number of patient-timepoint flips shared between the two models.
The number of flips in each direction changes as a function of random seed, and the patient-timepoints that flip are largely disjoint between random seeds.
\label{tab:flips}}
\begin{tabular}{@{}rlrrrrrrrr@{}}
\toprule
\multicolumn{1}{l}{} &
  \textbf{Shifted} &
  \multicolumn{2}{c}{\begin{tabular}[c]{@{}c@{}}E. Morning\\ (12am-6am)\end{tabular}} &
  \multicolumn{2}{c}{\begin{tabular}[c]{@{}c@{}}Morning\\ (6am-12pm)\end{tabular}} &
  \multicolumn{2}{c}{\begin{tabular}[c]{@{}c@{}}Afternoon\\ (12pm-6pm)\end{tabular}} &
  \multicolumn{2}{c}{\begin{tabular}[c]{@{}c@{}}Night\\ (6pm-12am)\end{tabular}} \\ \midrule
 \textbf{Original} &
   &
  \textbf{+ to -} &
  \textbf{- to +} &
  \textbf{+ to -} &
  \textbf{- to +} &
  \textbf{+ to -} &
  \textbf{- to +} &
  \textbf{+ to -} &
  \textbf{- to +} \\
  \cmidrule{3-10}
&
  model 1 &
  1417 &
  251 &
  1378 &
  456 &
  1400 &
  394 &
  1700 &
  396 \\
  &
  model 2 &
   1560 &
   285 &
   1425 &
   553 &
   1206 &
   764 &
   1663 &
   499 \\
\multirow{-3}{*}{ \begin{tabular}[c]{@{}r@{}}E. Morning\\ (12am-6am)\end{tabular}} &
  overlap &
   570 &
   31 &
   459 &
   70 &
   423 &
   103 &
   646 &
   76 \\ \cmidrule(l){2-10}
  &
  model 1 &
   1202 &
   321 &
   1279 &
   188 &
   1616 &
   278 &
   1523 &
   215 \\
  &
  model 2 &
   1368 &
   281 &
   1437 &
   193 &
   1694 &
   297 &
   1395 &
   366 \\
\multirow{-3}{*}{ \begin{tabular}[c]{@{}r@{}}Morning\\ (6am-12pm)\end{tabular}} &
  overlap &
   483 &
   42 &
   508 &
   26 &
   700 &
   49 &
   585 &
   50 \\ \cmidrule(l){2-10}
  &
  model 1 &
   588 &
   221 &
   509 &
   318 &
   601 &
   161 &
   767 &
   150 \\
  &
  model 2 &
   738 &
   191 &
   555 &
   271 &
   607 &
   109 &
   708 &
   189 \\
\multirow{-3}{*}{ \begin{tabular}[c]{@{}r@{}}Afternoon\\ (12pm-6pm)\end{tabular}} &
  overlap &
   255 &
   32 &
   209 &
   67 &
   258 &
   23 &
   294 &
   33 \\ \cmidrule(l){2-10}
  &
  model 1 &
   422 &
   199 &
   423 &
   349 &
   410 &
   223 &
   441 &
   124 \\
  &
  model 2 &
   520 &
   124 &
   542 &
   233 &
   417 &
   181 &
   468 &
   95 \\
\multirow{-3}{*}{ \begin{tabular}[c]{@{}r@{}}Night\\ (6pm-12am)\end{tabular}} &
  overlap &
   189 &
   35 &
   188 &
   73 &
   159 &
   41 &
   186 &
   19 \\
   \bottomrule
\end{tabular}
\end{table}

\subsection{Preliminary Ablation Experiment}
\label{sec:ehr ablation}
Finally, to test the hypothesis that our results point to the possibility of modifying the signals a predictor uses to make its predictions \emph{without} affecting iid performance, we perform an experiment where we ablate the timestamp feature entirely while training a predictor.
In particular, we rerun the pipeline with an LSTM architecture.
This simple ablation leads to a test set population performance similar to the rest of our ensemble of predictors where that feature was included (normalized PRAUC of 0.368, compared to a range of 0.346 to 0.366).

In addition, there is evidence that underspecification here results from a collinearity between features, similar to that discussed in the Genomics example in the main text. In particular, this ablated model can predict time of day with an accuracy of 85\% using an auxiliary head (without backpropagation).
These results suggest that the signal related to time of day is present through different correlations that the training pipeline is unable to pull apart.

\section{Genomics: Full Experimental Details}
\label{sec:apd_genomics}

In this section we provide full details of the random featurization experiment using linear models in genomic medicine, along with a brief overview of the relevant research areas.

\subsection{Background}

In genetics research, a \emph{genome-wide association study} (\emph{GWAS}) is an observational study of a large group of individuals to identify genetic variants (or \emph{genotypes}) associated with a particular trait (\emph{phenotype}) of interest. One application of GWAS results is for construction of a \emph{polygenic risk score} (\emph{PRS}) \cite{Wray2007-prs, ISC2009-prs} for the phenotype for each individual, generally defined as a weighted sum of the associated genotypes, where the weights are derived from the GWAS. One crucial factor to consider in this construction is that genetic variants are not independent and may contain highly correlated pairs due to a phenomenon called \emph{linkage disequilibrium} (\emph{LD}) \cite{Slatkin-LD}. The most common way to correct for LD is to partition the associated variants into clusters of highly-correlated variants and to only include one representative of each cluster for the PRS (e.g. \cite{ISC2009-prs, CAD2013-prs}). There are other more advanced methods (e.g. Bayesian modeling of LD \cite{Vilhjalmsson2015-ldpred}) which will not be discussed here.

While PRS show potential for identifying high-risk individuals for certain common diseases (e.g. \cite{Khera2018-prs-diseases}) when derived and tested within one ancestry group (mostly European), recent work has shown that the prediction accuracy of PRS from one ancestry group does not necessarily generalize to other ancestry groups \cite{Martin2017-prs-ancestry, Duncan2019-prs-ancestry, Berg2019-prs-ancestry}. When combined with the fact that more than three quarters of individuals in widely-used GWAS are of European ancestry \cite{Morales2018-gwas-ancestry} (while representing less than a quarter of global population), this has raised scientific and ethical concerns about the clinical use of PRS and GWAS in the community \cite{Martin2019-prs-fairness, Need2009-gwas-fairness, Popejoy2016-genomics-fairness}.

\subsection{Methods}

In this work we investigate the issue of generalizability of PRS from a slightly different angle. Instead of focusing on the loss of predictive accuracy of a PRS when transferred to a different ancestry group, we investigate the \emph{sensitivity} of the PRS to the choice of genotypes used in the derivation of the score, when evaluated within the same ancestry group versus outside of the group. Our phenotype of interest is the \emph{intraocular pressure} (\emph{IOP}), a continuous phenotype representing the fluid pressure inside the eye. This metric is an important aspect in the evaluation of risk of eye diseases such as glaucoma. We aim to predict IOP of individuals with their demographic information (age, sex, and BMI) and their genomic variants only, using the \emph{UK Biobank} dataset \cite{Sudlow2015-ukb}, a large, de-identified biobank study in the United Kingdom.

\begin{table}[!ht]
\caption{\textbf{Distribution of IOP associated variants}. 129 variant clusters are distributed over 16 chromosomes.}
\begin{center}
\begin{tabular}{c|c||c|c}
    \textbf{chrom} & \textbf{clusters} & \textbf{chrom} & \textbf{clusters} \\
    \hline
    chr1 & 19 & chr9 & 10 \\
    chr2 & 10 & chr11 & 16 \\
    chr3 & 10 & chr13 & 2 \\
    chr4 & 13 & chr14 & 2 \\
    chr5 & 1 & chr16 & 2 \\
    chr6 & 9 & chr17 & 4 \\
    chr7 & 15 & chr20 & 2 \\
    chr8 & 11 & chr22 & 3 \\
\end{tabular}
\end{center}
\label{table:genomics_iop_variants}
\end{table}

We first performed a GWAS on IOP and identified 4,054 genetic variants significantly associated with IOP distributed over 16 human chromosomes. We partitioned the variants into 129 clusters, where variants in the same cluster are highly-correlated and the ones in different clusters are relatively less correlated, and constructed the set of ``index variants'', consisting of the best representative of each cluster. We identified and clustered the IOP-associated variants with PLINK v1.9 \cite{Purcell2007-plink}, a standard tool in population genetics, using the \texttt{-{}-clump} command. We used $5\times 10^{-8}$ for the index variant p-value threshold, $5\times 10^{-6}$ for the p-value threshold for the rest of the associated variants, $0.5$ for the $r^2$ threshold, and 250 kb for the clumping radius. Table \ref{table:genomics_iop_variants} summarizes the distribution of IOP-associated variant clusters over chromosomes.

After identifying the 129 clusters of variants, we created 1,000 \emph{sets} of variants, each set consisting of 129 variants, exactly one variant in each cluster. The first of those sets is the set of index variants identified by PLINK. We then sampled 999 sets of cluster representatives by sampling one variant in each cluster uniformly at random. Each of these 1,000 sets defines a set of 129 genomic features to be used in our regression models.

For training and evaluation we partitioned the UK Biobank population into British and ``non-British'' individuals, and then we randomly partitioned the British individuals into British training and evaluation set. We leave the ``non-British'' individuals out of training and use them solely for evaluation. We measured the ``British-ness'' of an individual by the distance from the coordinate-wise median of the \emph{self-reported} British individuals, in the 10-dimensional vector space of the top 10 principal components (PCs) of genetic variation \cite{Price2006-pca-gwas}. Individuals whose z-scored distance from the coordinate-wise median are no greater than 4 in this PC space, are considered British. We then randomly partitioned 91,971 British individuals defined as above into a British training set (82,309 individuals) and a British evaluation set (9,662 individuals). All remaining ``non-British'' set (14,898 individuals) was used for evaluation.

We trained linear regression models for predicting IOP with (a) demographics and a set of 129 genomic features (one of the 1,000 sets created above) and (b) demographic features only, using the British training set. We used $L_2$ regularization whose strength was determined by 10-fold cross validation in the training set.

We observed drastically increased sensitivity (Figure 3, left, in the main text) for the genomic models (blue dots) in the ``non-British'' evaluation set compared to the British evaluation set or the training set. Genomic models' margin of improvement from the baseline demographic model (gray line) is also decreased in the ``non-British'' evaluation set. In the ``non-British'' evaluation set we still see in general some improvement over the baseline, but the margins are highly variable. We also observed that the performance in British and ``non-British'' evaluation sets are mostly uncorrelated ($r=0.131$) given the same set of genomic features (Figure 3, middle, in the main text).

Another interesting point is that the model using the original index variants (red dot) outperforms models with other choices of cluster representative in the British evaluation set, but not in the "non-British" evaluation set. This implies the cluster representatives chosen in the training domain are not always the best representatives outside of the British ancestry group.

In summary, we investigated the sensitivity to the choice of features (variants) representing clusters of highly correlated features in the context of genomics in this section. We observed that the prediction errors become highly sensitive when the evaluation domain is distinct from the training domain, in addition to being higher in magnitude. As previously discussed the robust generalization of PRS in underrepresented ancestry groups is one of the major open questions for its real-world application in clinical settings.

\renewcommand{\P}{{\sf P}}
\newcommand{\Var}{\text{Var}}
\newcommand{\Cov}{\text{Cov}}

\newcommand{\Z}{\mathbb{Z}}
\newcommand{\R}{\mathbb{R}}
\newcommand{\C}{\mathbb{C}}
\newcommand{\N}{\mathbb{N}}
\renewcommand{\S}{\mathbb{S}}
\def\ball{{\mathsf B}}

\newcommand{\vphi}{\varphi}
\def\id{{\mathbf I}}

\renewcommand{\d}{\textup{d}}
\renewcommand{\l}{\vert}
\newcommand{\dl}{\Vert}
\renewcommand{\>}{\rangle}
\newcommand{\sign}{\text{sign}}
\newcommand{\diag}{\text{\rm diag}}
\newcommand{\tr}{\text{tr}}
\newcommand{\op}{{\rm op}}
\newcommand{\ones}{\bm{1}}
\newcommand{\what}{\widehat}
\newcommand{\grad}{\nabla}
\def\sT{{\mathsf T}}
\def\bzero{{\boldsymbol 0}}

\newcommand{\eqnd}{\, {\buildrel d \over =} \,} 
\newcommand{\eqndef}{\mathrel{\mathop:}=}
\def\doteq{{\stackrel{\cdot}{=}}}
\newcommand{\goto}{\longrightarrow}
\newcommand{\gotod}{\buildrel d \over \longrightarrow} 
\newcommand{\gotoas}{\buildrel a.s. \over \longrightarrow} 
\def\simiid{{\stackrel{i.i.d.}{\sim}}}

\newcommand{\notate}[1]{\textcolor{red}{\textbf{[#1]}}}
\newcommand{\mc}[1]{\mathcal{#1}}
\newcommand{\mb}[1]{\mathbf{#1}}

\newtheorem{question}{Question}
\newtheorem{assumption}{Assumption}
\newtheorem{property}{Property}
\newtheorem{objective}{Objective}
\newtheorem{claim}{Claim}

\def\bxi{{\boldsymbol \xi}}
\def\cA{{\mathcal A}}
\def\cO{{\mathcal O}}
\def\hE{{\hat{\mathbb E}}}
\def\bU{{\boldsymbol U}}
\def\bV{{\boldsymbol V}}
\def\bz{{\boldsymbol z}}
\def\bOmega{{\boldsymbol \Omega}}
\def\bfe{{\boldsymbol e}}
\def\bG{{\boldsymbol G}}
\def\bbm{{\boldsymbol m}}
\def\bmu{{\boldsymbol \mu}}
\def\bdelta{{\boldsymbol \delta}}
\def\bh{{\boldsymbol h}}
\def\blambda{{\boldsymbol \lambda}}
\def\bB{{\boldsymbol B}}
\def\beps{{\boldsymbol \varepsilon}}
\def\bH{{\boldsymbol H}}
\def\bK{{\boldsymbol K}}
\def\bQ{{\boldsymbol Q}}
\def\bv{{\boldsymbol v}}
\def\bDelta{{\boldsymbol \Delta}}
\def\bE{{\boldsymbol E}}
\def\bX{{\boldsymbol X}}
\def\bY{{\boldsymbol Y}}
\def\bw{{\boldsymbol w}}
\def\bx{{\boldsymbol x}}
\def\by{{\boldsymbol y}}
\def\bW{{\boldsymbol W}}
\def\hba{\hat{\boldsymbol a}}
\def\ba{{\boldsymbol a}}
\def\bt{{\boldsymbol t}}
\def\bT{{\boldsymbol T}}
\def\bDelta{{\boldsymbol \Delta}}
\def\be{{\boldsymbol e}}
\def\bu{{\boldsymbol u}}
\def\bg{{\boldsymbol g}}
\def\bA{{\boldsymbol A}}
\def\btheta{{\boldsymbol \theta}}
\def\bTheta{{\boldsymbol \Theta}}
\def\bLambda{{\boldsymbol \Lambda}}
\def\bbeta{{\boldsymbol \beta}}
\def\bJ{{\boldsymbol J}}
\def\bC{{\boldsymbol C}}
\def\boldf{{\boldsymbol f}}
\def\bM{{\boldsymbol M}}
\def\bP{{\boldsymbol P}}
\def\bS{{\boldsymbol S}}
\def\bO{{\boldsymbol O}}
\def\bD{{\boldsymbol D}}
\def\bPsi{{\boldsymbol \Psi}}
\def\bsh{{\boldsymbol h}}
\def\bF{{\boldsymbol F}}
\def\bZ{{\boldsymbol Z}}
\def\bsigma{{\boldsymbol \sigma}}
\def\bq{{\boldsymbol q}}
\def\bXi{{\boldsymbol \Xi}}
\def\bfeta{{\boldsymbol \eta}}
\def\bR{{\boldsymbol R}}
\def\bi{{\boldsymbol i}}
\def\projp{{\sf P}^{\perp}}

\def\hbtheta{\hat{\boldsymbol\theta}}
\def\btheta{{\boldsymbol\theta}}
\def\berr{\mbox{{\bf err}}}
\def\bErr{\mbox{{\bf Err}}}

\def\cQ{{\mathcal Q}}
\def\cF{{\mathcal F}}
\def\cE{{\mathcal E}}
\def\cM{{\mathcal M}}
\def\cV{{\mathcal V}}
\def\cK{{\mathcal K}}

\def\sF{{\mathsf F}}
\def\sH{{\sf H}}
\def\sF{{\sf F}}

\def\oc{{\overline c}}
\def\ok{{\overline k}}
\def\obA{{\overline \bA}}

\def\oR{{\overline R}}

\def\reals{{\mathbb R}}
\def\complex{{\mathbb C}}
\def\integers{{\mathbb N}}

\def\tbw{\tilde{\boldsymbol w}}

\def\tm{{\tilde m}}

\def\init{\mbox{\rm\tiny init}}
\def\stest{\mbox{\rm\tiny test}}

\def\sop{\mbox{\rm\tiny op}}
\def\sav{\mbox{\rm\tiny av}}
\def\sGD{\mbox{\rm\tiny GD}}

\def\tsigma{{\tilde \sigma}}
\def\err{{\rm err}}
\def\disk{{\mathbb D}}
\def\bsF{\textbf{\textsf{F}}}
\def\good{{\rm good}}
\def\Func{{\rm Func}}
\def\ratio{{\zeta}}
\def\tbtheta{{\tilde \btheta}}
\def\tbx{{\tilde \bx}}
\def\tbB{{\tilde \bB}}
\def\tbQ{{\tilde \bQ}}
\def\tbJ{{\tilde \bJ}}
\def\tbH{{\tilde \bH}}
\def\tQ{{\tilde Q}}
\def\tJ{{\tilde J}}
\def\tH{{\tilde H}}
\def\bfone{{\mathbf 1}}
\def\tz{{\tilde z}}
\def\tE{{\tilde E}}
\def\ob{\mu}
\def\Tr{{\rm Tr}}
\def\normal{{\mathsf N}}
\def\de{{\rm d}}
\def\Unif{{\rm Unif}}
\def\proj{{\mathsf P}}
\def\He{{\rm He}}
\def\sP{{\sf P}}
\def\sshift{\mbox{\tiny\rm shift}}

\def\balpha{{\boldsymbol \alpha}}

\def\prob{{\mathbb P}}

\def\K{{\mathbb K}}

\def\Poly{{\rm Poly}}
\def\Coeff{{\rm Coeff}}
\def\RF{{\rm RF}}
\def\NT{{\rm NT}}
\def\diag{{\rm diag}}
\def\rank{{\rm rank}}
\def\Log{{\rm Log}}

\def\normf{F}
\def\sigmalinear{\mu_1}
\def\sigmares{\mu_\star}
\def\hbbeta{\hat{\boldsymbol \beta}}
\def\bSigma{{\boldsymbol\Sigma}}

\def\lsamp{\mbox{\tiny\rm lsamp}}
\def\wide{\mbox{\tiny\rm wide}}
\def\rless{\mbox{\tiny\rm rless}}
\def\sALG{\mbox{\tiny\rm ALG}}
\def\ALG{\mbox{\rm ALG}}
\def\sNL{\mbox{\tiny\rm NL}}
\def\sL{\mbox{\tiny\rm L}}
\def\hf{\hat{f}}

\def\olambda{\overline{\lambda}}
\def\cuP{\mathscrsfs{P}}
\def\cuR{\mathscrsfs{R}}
\def\cuE{\mathscrsfs{E}}
\def\cuB{\mathscrsfs{B}}
\def\cuV{\mathscrsfs{V}}
\def\cuD{\mathscrsfs{D}}
\def\cuG{\mathscrsfs{G}}
\def\cuH{\mathscrsfs{H}}
\def\cuL{\mathscrsfs{L}}
\def\cuT{\mathscrsfs{T}}

\def\cuN{\mathscrsfs{N}}
\def\bzero{{\boldsymbol{0}}}
\def\bphi{{\boldsymbol{\phi}}}
\def\bPhi{{\boldsymbol{\Phi}}}

\def\hyp{{\tau}}

\def\ratio{\zeta}
\def\olambda{\overline{\lambda}}
\def\ba{{\boldsymbol a}}
\def\hba{\hat{\boldsymbol a}}

\section{Random Feature Model: Complete Theoretical Analysis}
\label{sec:theory appendix}

In this section we presents definitions and asymptotic formulas for the
random features model that is presented in Section 3 of the main text. Before focusing on the random features
model, we will describe the general setting and define some quantities of interest.

Our general focus is to investigate the validity of two main hypotheses
that arose from our empirical case studies:
$(i)$~The model learnt depends strongly on the arbitrary or random 
choices in the training procedure; $(ii)$~As a consequence, for most choices of the training procedure,
there exist test distributions that are close to the train
distribution and have much higher test error, while the test error on the same test distribution is unchanged for other choices of the training procedure.

\subsection{General Definitions}

We consider for simplicity a regression problem: we are given
data $\{(\bx_i,y_i)\}_{i\le n}$, with $\bx_i\in\reals^d$  
vector of covariates and $y_i\in\reals$ a response. We learn a model
$f_{\hyp}:\reals^d\to\reals$, where $\hyp$ captures the arbitrary choices in the training procedure,
such as initialization, stepsize schedule, and so on.
Also, to be concrete, we consider square loss and hence the test error reads
\begin{align}
  R(\hyp,\P_{\stest}) := \E_{\stest}\{(y-f_{\hyp}(\bx))^2\}\,.
\end{align}
Our definitions are easily generalized to other loss functions.
The notation emphasizes that the test error is computed with respect to 
a distribution $\P_{\stest}$ that is not necessarily the same as the training one
(which will be denoted simply by $\P$). The classical in-distribution test error reads $R(\hyp,\P)$.

As a first question, we want to investigate to what extent the model
$f_{\tau}(\bx)$ is dependent on the  arbitrary choice of $\tau$, in particular when this is random.
In order to explore this point, we define the model sensitivity as
\begin{align}
S(\hyp_1,\hyp_2;\P_{\stest}):= \E_{\stest}\big\{[f_{\hyp_1}(\bx)-f_{\hyp_2}(\bx)]^2\big\}\, .
\end{align}

We next want to explore the effect to this sensitivity on the out-of-distribution test error.
In particular, we want to understand whether the out-of-distribution error can increase
significantly, even when the in-distribution error does not change much.
Normally, the out-of-distribution risk is defined by
constructing a suitable neighborhood of the train distribution $\P$, call it $\cN(\P)$,
and letting
$R_{\sshift}(\hyp_0)  := \sup_{\P_{\stest}\in\cN(\P)}  R(\hyp_0;\P_{\stest})$.

Here we extend this classical definition, as to incorporate the constraint
that the distribution shift should not damage the model constructed with an average
choice of $\tau$:
\begin{align}
R_{\sshift}(\hyp_0;\delta)  := \sup_{\P_{\stest}\in\cN(\P)}  \big\{R(\hyp_0;\P_{\stest}): \; \E_{\hyp}R(\hyp;\P_{\stest})\le \delta\big\}\, .
\label{eq:Rshift}
\end{align}

\subsection{Random Featurization Maps}

A broad class of overparametrized models is obtained by constructing a 
featurization map $\bphi_{\hyp}:\reals^d\to\reals^N$.  We then fit a model that is linear
in $\bphi_{\hyp}(\bx)$, e.g. via min-norm interpolation
\begin{align}
    \mbox{minimize}&\;\;\;\;\|\btheta\|_2\, ,\\
    \mbox{subject to}&\;\;\;\; \by = \bPhi_{\hyp}(\bX)\btheta\, .
\end{align}
(Other procedures make sense as well.)
Here $\bPhi_{\hyp}(\bX)\in\reals^{n\times N}$ is the matrix whose $i$-th row is $\bphi_{\hyp}(\bx_i)$.
The corresponding estimator is denoted by $\hbtheta_{\hyp}$,
and the predictive model is $f_{\hyp}(\bx) = \<\hbtheta_{\hyp},\bphi_{\hyp}(\bx)\>$.
It is useful to consider a couple of examples.

\paragraph{Example 1.} Imagine training a highly overparametrized neural network
using SGD.  Let $F(\,\cdot\, ;\bw):\reals^d\to \reals$ be the input-output relation of the network.
In the lazy training regime, this is well approximated by 
its first-order Taylor expansion around the initialization $\bw_0$ \citep{chizat2019lazy}.
Namely $F(\bx;\bw)\approx F(\bx;\bw_0)+\<\nabla_{\bw} F(\bx;\bw_0),\bw-\bw_0\>$.
If the initialization is symmetric, we can further neglect the zero-th order term, 
and, by letting $\btheta=\bw-\bw_0$, we obtain $F(\bx;\bw)\approx\<\nabla_{\bw} F(\bx;\bw_0),\btheta\>$.
We can therefore identify 
$\hyp=(\bw_0)$ and $\bphi_{\hyp}(\bx) = \nabla_{\bw}F(\bx;\bw_0)$.

\paragraph{Example 2.} Imagine $\bx_i\in\reals^d$ represents the degree of activity of $d$ biological mechanism in patient $i$.
We do not have access to $\bx_i$, and instead we observe the expression levels
of $N_0\gg d$ genes, which are given by $\bu_{0,i} = \bW_{0}\bx_i+\bz_{0,i}$,
where $\bW_{0}\in\reals^{N_0\times d}$ and $\bz_{0,i}$ are unexplained effects.
We do not fit a model that uses the whole vector $\bu_{0,i}$,
and instead select by a clustering procedure $\hyp$ a subset of $N$ genes,
hence obtaining a vector of features $\bu_i = \bW_{\hyp}\bx_i+\bz_{i}\in\reals^N$. In this case we identify the featurization map with the random map $\bphi_{\hyp}(\bx_i) := \bW_{\hyp}\bx_i+\bz_{i}$.

\vspace{0.2cm}

As a mathematically rich and yet tractable model, we consider the random features model of \cite{rahimi2008random},
whereby
\begin{align}
  \bphi_{\tau}(\bx) = \sigma(\bW\bx)\, ,\;\;\;\;\; \bW\in\reals^{N\times d}\, .
\end{align}
Here $\sigma:\reals\to\reals$ is an activation function: it is understood that this applied to vectors entrywise.
Further, $\bW$ is a matrix of first-layer weights which are drawn randomly and are not optimized over.
We will draw the rows $(\bw_i)_{i\le N}$ independently with $\bw_i\sim\Unif(\S^{d-1}(1))$.
(Here and below $\S^{d-1}(r)$ is the sphere of radius $r$ in $d$ dimensions.) We identify the arbitrary training
choice with the choice of this first-layer weights $\hyp = \bW$. 

We assume an extremely simple data distribution,  namely $\bx_i\sim\Unif(\S^{d-1}(\sqrt{d}))$
and $y_i = f_*(\bx_i)= \<\bbeta_0,\bx_i\>$. Note that $\|f_*\|_{L^2}=\|\bbeta_0\|_2$.

We will derive exact characterizations of the sensitivity
and risk under shifted test distribution in the proportional asymptotics
$N,n,d\to\infty$, with 
\begin{align}
    \frac{N}{d}\to \psi_1\, ,\;\;\;\;
    \frac{n}{d}\to \psi_2\, ,\label{eq:Psi1Psi2}
\end{align}
for some $\psi_1,\psi_2\in (0,\infty)$. In what follows we will
assume to to be given sequences of triples $(N,n,d)$ which without
loss of generality we can think to be indexed by $d$.
When we write $d\to\infty$, it is understood that 
$N,n\to\infty$ as well ,  with Eq.~\eqref{eq:Psi1Psi2} holding.
Finally, we assume $\lim_{d\to\infty}\|\bbeta_0\|_2^2=r^2\in(0,\infty)$.

\subsection{Random features model: Risk}

We begin by recalling some results and notations from \cite{mei2019generalization}. We refer to the original paper
for formal statements.

In this section we consider the random features model  under a slightly more general setting than the one introduced above. Namely, we allow for Gaussian noise
$y_i = f_*(\bx_i)+\eps_i$, $\eps_i\sim\normal(0,s^2)$, and perform  ridge regression with a positive
regularization parameter $\lambda>0$:
\begin{align}
  \hbtheta(\bW,\bX,\by;\lambda) &:= \argmin_{\btheta\in\reals^N}\Big\{\frac{1}{n}\sum_{i=1}^n\big(y_i-
  \<\btheta,\sigma(\bW\bx_i)\>\big)^2+\frac{N\lambda}{d}\|\btheta\|_2^2\Big\}\, ,
\end{align}
In the following we will omit all arguments except $\bW$, and write $\hbtheta(\bW):=\hbtheta(\bW,\bX,\by;\lambda)$.
By specializing our general definition, the risk per realization of the first layer weights $\bW$
is given by 
\begin{align}
  R(\bW,\P_{\stest})=R(\hyp,\P_{\stest}) :=\E_{\stest}\big\{[y-\<\hbtheta(\bW),\sigma(\bW\bx)\>]^2\big\}\, .
\end{align}

 The activation function is assumed to be $\sigma\in L^2(\reals,\gamma)$, with $\gamma$
the Gaussian measure. Such an activation function is characterized via its projection onto constant and linear
functions
\begin{align}
  \mu_0 := \E\{\sigma(G)\}\, , \;\;\;\; \mu_1 :=\E\{G\sigma(G)\}\, ,\;\;\;\;
  \mu_*^2 := \E\{\sigma(G)^2\}-\mu_0^2-\mu_1^2\, .
\end{align}
In particular, we define the following ratio
\begin{equation}
  \ratio\equiv \frac{\mu_1}{\mu_*}\, .
\end{equation}
Let $\nu_1,\nu_2:\complex_+\to\complex$ be analytic functions such that, for $\Im(\xi)>C$ a large enough
constant $\nu_1(\xi), \nu_2(\xi)$  satisfy
\begin{equation}
\begin{aligned}
\nu_1 =&~ \psi_1\Big(-\xi -  \nu_2 - \frac{\ratio^2 \nu_2}{1- \ratio^2 \nu_1\nu_2}\Big)^{-1}\, ,\\
\nu_2 =&~ \psi_2\Big(-\xi - \nu_1 - \frac{\ratio^2 \nu_1}{1- \ratio^2 \nu_1\nu_2}\Big)^{-1}\, .
\end{aligned}
\end{equation}
We then let
\begin{align}\label{eqn:definition_chi_main_formula}
  \chi &\equiv \nu_1(\bi ( \psi_1 \psi_2 \olambda)^{1/2}) \cdot \nu_2(\bi ( \psi_1 \psi_2 \olambda)^{1/2}),\\
  \olambda &= \frac{\lambda}{\mu_*^2}\, .
\end{align}

Finally, define
\begin{equation}\label{eq:E012def}
\begin{aligned}
\cuE_0(\ratio, \psi_1, \psi_2, \olambda) \equiv&~  - \chi^5\ratio^6 + 3\chi^4 \ratio^4+ (\psi_1\psi_2 - \psi_2 - \psi_1 + 1)\chi^3\ratio^6 - 2\chi^3\ratio^4 - 3\chi^3\ratio^2 \\
&+ (\psi_1 + \psi_2 - 3\psi_1\psi_2 + 1)\chi^2\ratio^4 + 2\chi^2\ratio^2+ \chi^2+ 3\psi_1\psi_2\chi\ratio^2 - \psi_1\psi_2\, ,\\
\cuE_1(\ratio, \psi_1, \psi_2,\olambda)  \equiv&~ \psi_2\chi^3\ratio^4 - \psi_2\chi^2\ratio^2 + \psi_1\psi_2\chi\ratio^2 - \psi_1\psi_2\, , \\
\cuE_2(\ratio, \psi_1, \psi_2, \olambda) \equiv&~ \chi^5\ratio^6 - 3\chi^4\ratio^4+ (\psi_1 - 1)\chi^3\ratio^6 + 2\chi^3\ratio^4 + 3\chi^3\ratio^2 + (- \psi_1 - 1)\chi^2\ratio^4 - 2\chi^2\ratio^2 - \chi^2\, .
\end{aligned}
\end{equation}

We then have, from \cite[Theorem 1]{mei2019generalization}, the following characterization of the \emph{in-distribution} 
risk\footnote{Theorem 1 in \cite{mei2019generalization} holds for the risk, conditional on
  the realization of $\bX,\by$. The statement given here is obtained simply my taking expectation over
$\bX,\by$.}
\begin{align}\label{eq:RiskRF}
R(\bW, \P) = r^2\, \frac{\cuE_1(\ratio, \psi_1, \psi_2, \olambda)}{\cuE_0(\ratio, \psi_1, \psi_2, \olambda)}+ 
s^2  \frac{\cuE_2(\ratio, \psi_1, \psi_2, \olambda)}{\cuE_0(\ratio, \psi_1, \psi_2, \olambda)} + o_{\P}(1)\, .
\end{align}
Here the $o_P(1)$ term depends on the realization of $\bW$, and is such that $\E|o_{\P}(1)| \to 0$
as $N,n,d\to\infty$.

\begin{remark}
  Notice that the right-hand side of Eq.~\eqref{eq:RiskRF} is independent of $\bW$. Hence we see that the
  in-distribution error is (for large $N,n,d$) essentially the same for most choices of $\bW$.
\end{remark}

\subsection{Random features model: Sensitivity to random featurization}

\begin{figure}[t]
  \begin{center}
    \includegraphics[width=0.75\linewidth]{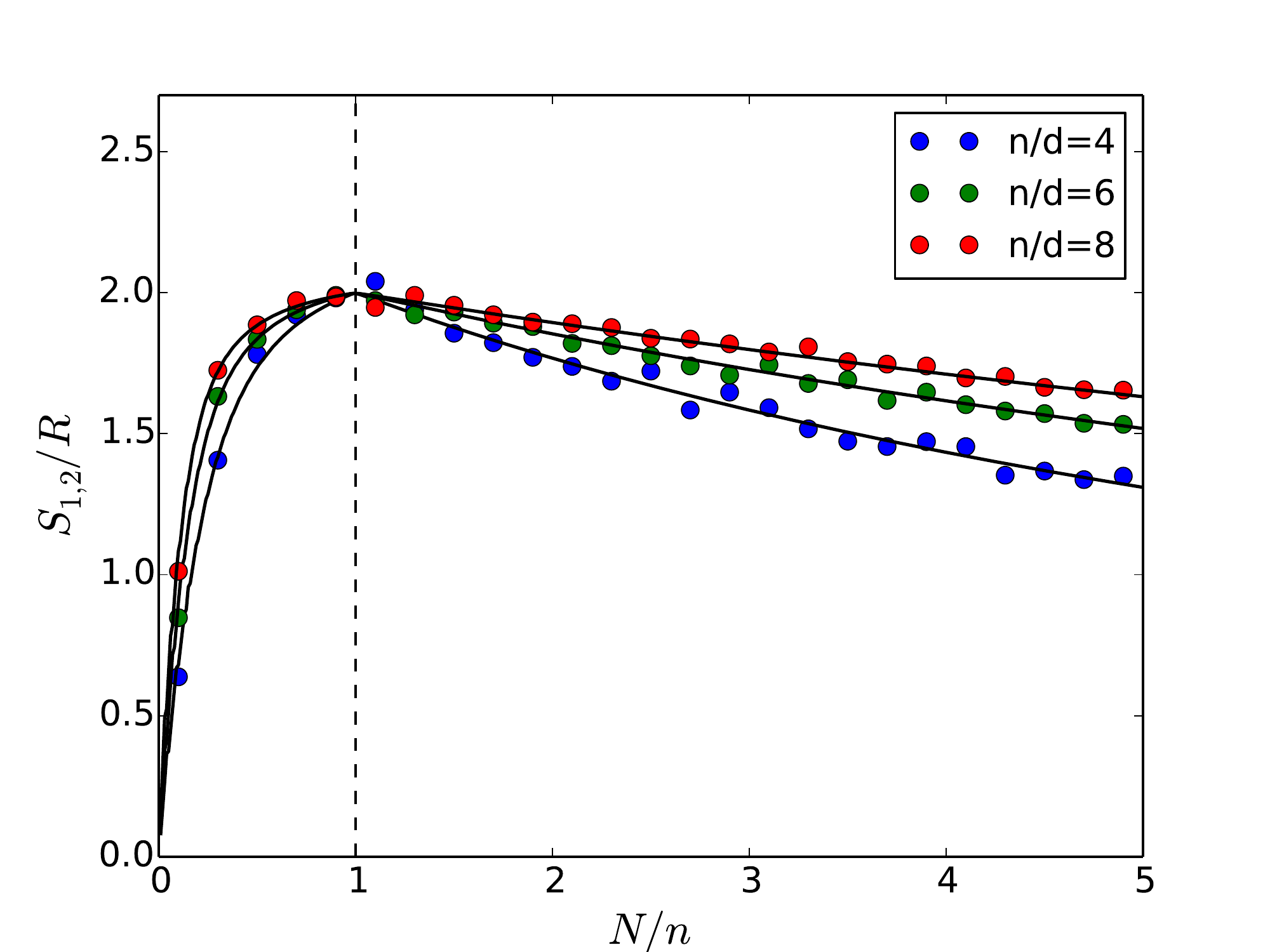}
    \end{center}
    \caption{
      Random features model trained via min-norm least squares:
      sensitivity to the initial condition, normalized by the risk. 
    Here the input dimension is $d=40$, $N$ is the number of neurons, and $n$ the number of samples. We use ReLU activations; the ground truth is linear with  $\|\bbeta_0\|_2=1$. 
    Circles are empirical results obtained by averaging over $50$  realizations. 
    Continuous lines correspond to the analytical prediction of Eq.~\eqref{eq:SavFormula}.}
    \label{fig:RF-Sensitivity-Theory}
\end{figure}
Let $\bW_1,\bW_2$ be two realizations of the first-layer weights.
We can decompose
\begin{align}
  \sigma(\bW\bx) = \mu_0+\mu_1\bW\bx +\mu_* \bz^{\perp}\, ,
\end{align}
where, under $\sP$, we have $\E\{\bz^{\perp}(\bz^{\perp})^{\sT}\}=\id+ c_n\bfone\bfone^{\sT}+\bDelta$,
$\|\bDelta\|_{\op} = o_{\P}(1)$, and $\E\{(\bW\bx)\bz^{\perp}\}=0$. We therefore have (writing $\hbtheta_i :=\hbtheta(\bW_i)$)
\begin{align*}
  S(\bW_1,\bW_2) &= \E\big\{(\<\hbtheta(\bW_1),\sigma(\bW_1\bx)\>- \<\hbtheta(\bW_2),\sigma(\bW_2\bx)\>)^2\big\}\\
  &= \mu_1^2\E\big\{(\<\hbtheta(\bW_1),\bW_1\bx\>- \<\hbtheta(\bW_2),\bW_2\bx\>)^2\big\}+
  \mu_*^2\E\big\{(\<\hbtheta(\bW_1)-\hbtheta(\bW_2),\bz^{\perp}\>)^2\big\}\\
  &=\mu_1^2\E_{\bX,\by}\big\{\big\|\bW_1^{\sT}\hbtheta_1-\bW_2^{\sT}\hbtheta_2\big\|_2^2\big\}+ (\mu_*^2+o_{\P}(1))
  \E_{\bX,\by}\{\|\hbtheta_1-\hbtheta_2\|^2\}\, .
\end{align*}

We consider two random independent choices of $\bW_1$, $\bW_2$, and define $S_{\sav}:=\E\{ S(\bW_1,\bW_2)\}$,
thus obtaining:
\begin{align}
  S_{\sav} = 2\mu_1^2\Big\{\E\big[\|\bW^{\sT}\hbtheta(\bW)\|_2^2\big]- \E\big[
    \big\|\E[\bW^{\sT}\hbtheta(\bW)|\bX,\bbeta_0]\big\|_2^2\big]\big]\Big\}+2\mu_{*}^2\E\big\{\|\hbtheta(\lambda)\|_2^2
    \big\}+o(1)\, .\label{eq:Sav}
\end{align}
In order to evaluate the asymptotics of this expression, we recall
some formulas that follow from \cite{mei2019generalization}:
\begin{align}
  \|\bW^{\sT}\hbtheta(\bW)\|^2_2 &= \frac{r^2}{\mu_*^2}\cdot\frac{\cuD_1(\ratio,\psi_1,\psi_2,\olambda)}{(\chi\ratio^2-1)\cuD_0(\ratio,\psi_1,\psi_2,\olambda)}+\frac{s^2}{\mu_*^2}\cdot\frac{\cuD_2(\ratio,\psi_1,\psi_2,\olambda)}{
    \cuD_0(\ratio,\psi_1,\psi_2,\olambda)}
  +o_{\P}(1)\, ,\label{eq:Wa2}\\
  \E(\bW^{\sT}\hbtheta(\bW)|\bbeta_0) & = \left\{\frac{1}{\mu_1}\cuH(\ratio,\psi_1,\psi_2,\olambda) +o(1)\right\}\bbeta_0\, ,\label{eq:eWa}\\
  \|\hbtheta(\bW)\|^2_2 & = \frac{r^2}{\mu_*^2}\frac{\cuG_1(\ratio,\psi_1,\psi_2,\olambda)}{\cuG_0(\ratio,\psi_1,\psi_2,\olambda)} +
  \frac{s^2}{\mu_*^2} \frac{\cuG_2(\ratio,\psi_1,\psi_2,\olambda)}{\cuG_0(\ratio,\psi_1,\psi_2,\olambda)}+o_{\P}(1)\, ,
  \label{eq:Theta2}
\end{align}
Here the terms $o_{\P}(1)$ converge to $0$ in $L^1$, and we used the following notations: 
\begin{align}
  \cuD_0(\ratio,\psi_1,\psi_2,\olambda) & = \chi^5\ratio^6 - 3\chi^4\ratio^4 +
  (\psi_1 + \psi_2 - \psi_1\psi_2 - 1)\chi^3\ratio^6+ 2\chi^3\ratio^4 + 3\chi^3\ratio^2 \\
  & +(3\psi_1\psi_2 - \psi_2 - \psi_1 - 1)\chi^2\ratio^4- 2\chi^2\ratio^2 - \chi^2 - 3\psi_1\psi_2\chi\ratio^2 + \psi_1\psi_2
  \, ,\nonumber\\
  \cuD_1(\ratio,\psi_1,\psi_2,\olambda) & = \chi^6\ratio^6-2\chi^5\ratio^4-(\psi_1\psi_2-\psi_1-\psi_2+1)\chi^4\ratio^6+\chi^4\ratio^4\\
  & +\chi^4\ratio^2-2(1-\psi_1\psi_2)\chi^3\ratio^4-(\psi_1+\psi_2+\psi_1\psi_2+1)\chi^2\ratio^2-\chi^2\, ,\nonumber\\
  \cuD_2(\ratio,\psi_1,\psi_2,\olambda) & = -(\psi_1-1)\chi^3\ratio^4-\chi^3\ratio^2+(\psi_1+1)\chi^2\ratio^2+\chi^2\, ,\\
  \cuG_0(\ratio,\psi_1,\psi_2,\olambda) & =  - \chi^5\ratio^6 + 3\chi^4\ratio^4 + (\psi_1\psi_2 - \psi_2 - \psi_1 + 1)\chi^3\ratio^6 - 2\chi^3\ratio^4 - 3\chi^3\ratio^2\\
  &~+ (\psi_1 + \psi_2 - 3\psi_1\psi_2 + 1)\chi^2\ratio^4 + 2\chi^2\ratio^2 + \chi^2 + 3\psi_1\psi_2\chi\ratio^2 - \psi_1\psi_2\, ,\nonumber\\
  \cuG_1(\ratio,\psi_1,\psi_2,\olambda) & =  - \chi^2 (\chi \ratio^4 - \chi \ratio^2 + \psi_2 \ratio^2 + \ratio^2 - \chi \psi_2 \ratio^4 + 1\, ,\\
  \cuG_2(\ratio,\psi_1,\psi_2,\olambda) & = \chi^2 (\chi \ratio^2 - 1) (\chi^2 \ratio^4 - 2 \chi \ratio^2 + \ratio^2 + 1)\, ,\\
  \cuH(\ratio,\psi_1,\psi_2,\olambda) & = \frac{\ratio^2\chi}{\ratio^2\chi-1}\, .
 \end{align}

We also claim that, for $s=0$ we have
\begin{align}
  \E(\bW^{\sT}\hbtheta(\bW)|\bX,\bbeta_0) & = \frac{\mu_1}{d}\bX^{\sT}\bX\Big(\frac{\mu_1^2}{d}\bX^{\sT}\bX+\mu_*^2(1+q)\id_d\Big)^{-1}\bbeta_0 + \berr(d,\lambda)\, ,\label{eq:Claim}\\
  q &:= -\frac{(\psi_2-\psi_1)_+}{\chi_0}\, ,\\
  \chi_0 & = \frac{1+\ratio^2-\psi_!\ratio^2-\sqrt{(1+\zeta^2-\psi_1\ratio^2)^2+4\psi_1\ratio^2}}{2\ratio^2}\, ,
\end{align}
where the error term $\berr(d,\lambda)$ satisfies
\begin{align}
  \lim_{\lambda\to 0}\lim_{d\to\infty}\E\big\{\|\berr(d,\lambda)\|_2^2\big\} = 0\, .
\end{align}
We refer to Section \ref{sec:SketchClaim} for a sketch of the proof of this claim.

Using Eq.~\eqref{eq:Claim} and  the asymptotics of the Stieltjes transform of the spectrum of
Wishart matrices, it follows that
\begin{align}
   \lim_{\lambda\to 0}\lim_{d\to\infty}\E\big[  \big\|\E[\bW^{\sT}\hba(\lambda)|\bX,\bbeta_0]\big\|_2^2\big]
  &= r^2\cuL(\zeta,\psi_1,\psi_2)\, ,\nonumber\\
   \cuL(\zeta,\psi_1,\psi_2)&:= 1-2\frac{1+q}{\ratio^2}\, g\Big(-\frac{1+q}{\ratio^2};\psi_2\Big)
   +\Big(\frac{1+q}{\ratio^2}\Big)^2 g'\Big(-\frac{1+q}{\ratio^2};\psi_2\Big)\, ,\label{eq:Ew2}\\
  g(z;\psi_2) &:= \frac{\psi_2-1-z-\Delta}{2z}\, ,\;\;\;\;\; \Delta:=-\sqrt{(\psi_2-1-z)^2-4z}\, ,\nonumber\\
  g'(z;\psi_2) & =\frac{-\psi_2+1+z+\Delta}{2z^2}-\frac{-\psi_2-1+z+\Delta}{2z\Delta}\, .\nonumber
\end{align}

Using Eqs.~\eqref{eq:Wa2}, \eqref{eq:Theta2}, \eqref{eq:Ew2} in Eq.~\eqref{eq:Sav}, we finally obtain:
\begin{align}
  \lim_{\lambda\to 0}  \lim_{d\to\infty}S_{\sav} =
  2r^2\Big\{\frac{\ratio^2\cuD_1}{(\chi\ratio^2-1)\cuD_0}-\cuL+\frac{\cuG_1}{\cuG_0}\Big\}\, .\label{eq:SavFormula}
\end{align}

In Figure 3 in the main text we compare this asymptotic prediction with numerical
simulations for $d=40$. We report the 
in-distribution sensitivity  $S_{\sav}$
 normalized by the risk $R(\bW,\P)$.
 In the classical underparametrized regime $N/n\to 0$,
 the sensitivity is small. However, as the number of neurons increases, $S/R$ grows rapidly,
 with $S/R$ not far from $2$
 over a large interval.
 Notice that $S/R=2$
has a special meaning.  Letting $h_i(\bx) = f_{\hyp_i}(\bx)-f_*(\bx)$, it corresponds to $\|h_1-h_2\|_{L^2}^2  = 2\|h_1\|_{L^2}^2=2\|h_2\|_{L^2}^2$,
i.e. $\<h_1,h_2\>_{L^2}=0$. In other words, two models generated with two random choices of $\tau$ as `as orthogonal as they can be'.

\subsection{Random features model: Distribution shift}

In order to explore the effect of a distribution shift in the random features model, 
we consider the case of a  mean shift. Namely, $\bx_{\stest} = \bx_0+\bx$ where $\bx\sim \Unif(\S^{d-1}(\sqrt{d}))$
is again uniform on the sphere, and $\bx_0$ is deterministic and adversarial
for a given choice $\bW_0$, under the constraint $\|\bx_0\|_2\le\Delta$. We denote this distribution
by $\P_{\bW_0,\Delta}$. We will construct a specific perturbation $\bx_0$ that produces a large
increase in  $R(\bW_0,\P_{\bW_0,\Delta})$ but a small  change on $R(\bW,\P_{\bW_0,\Delta})$ for a typical
random $\bW$ independent of $\bW_0$. We leave to future work the problem of determining the worst case
perturbation $\bx_0$.

We next consider the risk when the first layer weights are $\bW$, and the test distribution is
$\P_{\bW_0,\Delta}$. Using the fact that $\|\bx_0\|_{2}=\Delta\ll \|\bx\|_2$, we get
\begin{align*}
  R(\bW,\P_{\bW_0,\Delta}) &= \E\big\{(\<\bbeta_0,\bx_{\stest}\>-\<\hbtheta(\bW),\sigma(\bW\bx_{\stest})\>)^2\big\}\\
  &= \E\big\{(\<\bbeta_0,\bx\>+\<\bbeta_0,\bx_0\>-\<\hbtheta(\bW),\sigma(\bW\bx)\>-\<\hbtheta(\bW),\sigma'(\bW\bx)\odot
  \bW\bx_0\>)^2\big\} +o_{\P}(1)\\
  &\stackrel{(a)}{=} \E\big\{(\<\bbeta_0,\bx\>+\<\bbeta_0,\bx_0\>-\<\hbtheta(\bW),\sigma(\bW\bx)\>-\mu_1\<\hbtheta(\bW),\bW\bx_0\>)^2\big\} +o_{\P}(1)\\
  &\stackrel{(b)}{=}  R(\bW,\P_{\bW_0,\Delta})+ \<\bbeta_0-\mu_1\bW^{\sT}\hbtheta(\bW),\bx_0\>^2 +o_{\P}(1)\, ,
\end{align*}
where $(a)$ follows by replacing $\sigma'(\<\bw_i,\bx\>)$ by its expectation over $\bx$
$\E\sigma'(\<\bx_i,\bx\>)= \E\sigma'(G)+o(1) = \mu_1+o(1)$, and $(b)$ since $\E(\bx) = 0$.

The choice $\bx_0$ that maximizes the risk $R(\bW_0,\P_{\bW_0,\Delta})$ is
$\bx_0= \Delta (\bbeta_0-\mu_1\bW_0^{\sT}\hbtheta(\bW_0))/\|\bbeta_0-\mu_1\bW_0^{\sT}\hbtheta(\bW_0)\|_2$.
However this mean shift can have a significant component along $\bbeta_0$, which results in a large
increase of $R(\bW;\P_{\bW_0,\Delta})$ for other $\bW$ as well. To avoid this, we project this vector orthogonally to
$\bbeta_0$:
\begin{align}
  \bx_0 = -\Delta \frac{\projp_{\bbeta_0}\bW_0^{\sT}\hbtheta(\bW_0)}{\|\projp_{\bbeta_0}\bW_0^{\sT}\hbtheta(\bW_0)\|_2}\, .
\end{align}
This results in the following expression for the test error on the shifted distribution:
\begin{align}
  R(\bW,\P_{\bW_0,\Delta}) & =  R(\bW,\P)+ \Delta^2\mu_1^2 T(\bW,\bW_0) +o_{\P}(1)\, ,\\
 T(\bW,\bW_0) &:= \frac{\<\projp_{\bbeta_0}\bW^{\sT}\hbtheta(\bW),\projp_{\bbeta_0}\bW_0^{\sT}\hbtheta(\bW_0)\>}{\|\projp_{\bbeta_0}\bW_0^{\sT}\hbtheta(\bW_0)\|^2_2}\, .
\end{align}
We first consider the case $\bW=\bW_0$. We then have 
\begin{align}
  \E T(\bW_0,\bW_0) &:= \E\{\|\projp_{\bbeta_0}\bW_0^{\sT}\hbtheta(\bW_0)\|^2_2\} \\
  & = \E\{\|\bW_0^{\sT}\hbtheta(\bW_0)\|^2_2 \}-\frac{1}{r^2}\E\{\<\bbeta_0,\bW_0^{\sT}\hbtheta(\bW_0)\>^2\} \\
  & \stackrel{(a)}{=} r^2\left\{\frac{\ratio^2\cuD_1}{(\chi\ratio^2-1)\cuD_0}-\cuH^2\right\}+s^2\frac{\ratio^2\cuD_2}{\cuD_0}
  +o(1)\, ,
\end{align}
where $(a)$ follows by Eqs.~\eqref{eq:Wa2} and \eqref{eq:eWa}.

For $\bW$ independent of $\bW_0$, we have (for $s=0$)
\begin{align}
  \E T(\bW,\bW_0) &:= \E\left\{\frac{(\<\bW^{\sT}\hbtheta(\bW),\bW_0^{\sT}\hbtheta(\bW_0)\>-
    r^{-2}\<\bbeta_0,\bW_0^{\sT}\hbtheta(\bW_0)\>\<\bbeta_0,\bW_1^{\sT}\hbtheta(\bW_1)\>)^2}{\|\bW_0^{\sT}\hbtheta(\bW_0)\|^2_2 -r^{-2}\<\bbeta_0,\bW_0^{\sT}\hbtheta(\bW_0)\>^2}\right\}\nonumber\\
  & = 
  \frac{\big(\E_{\bX,\by}\{\|\E_{\bW}[\bW^{\sT}\hbtheta(\bW)|\bX,\by]\|^2\}
    -r^{-2}(\E_{\bW,\bX,\by}\<\bbeta_0,\bW^{\sT}\hbtheta(\bW)\>)^2\big)^2}
     {\E_{\bW,\bX,\by}\{\|\bW^{\sT}\hbtheta(\bW)\|^2_2\} -r^{-2}(\E_{\bW,\bX,\by}\<\bbeta_0,\bW^{\sT}\hbtheta(\bW)\>)^2}+o_{\P}(1)\nonumber\\
  & = \frac{\cuT_1(\zeta,\psi_1,\psi_2,\olambda)^2}{\cuT_0(\zeta,\psi_1,\psi_2,\olambda)}+o_{\P}(1)\, ,\label{eq:T1T0}
\end{align}
where
\begin{align}
  \cuT_0(\zeta,\psi_1,\psi_2,\olambda)&=\frac{\ratio^2\cuD_1(\zeta,\psi_1,\psi_2,\olambda)}{(\chi\ratio^2-1)\cuD_0(\zeta,\psi_1,\psi_2,\olambda)}-\cuH^2(\zeta,\psi_1,\psi_2,\olambda)\, ,\\
  \lim_{\olambda\to 0}\cuT_1(\zeta,\psi_1,\psi_2,\olambda)&=\cuL(\zeta,\psi_1,\psi_2)-\cuH^2(\zeta,\psi_1,\psi_2,0)\, .
\end{align}

In Figure~\ref{fig:RF-Shift-Theory} of the main text,
we plot the ratios
\begin{align}
  \frac{\E_{\bW_0} R(\bW_0;\P_{\bW_0,\Delta})}{\E_{\bW}R(\bW;\P)}\, ,
  \;\;\;\;\;\;\;
  \frac{\E_{\bW,\bW_0} R(\bW;\P_{\bW_0,\Delta})}{\E_{\bW}R(\bW;\P)}\,.
\end{align}
Note that the perturbation introduced here is extremely small in $\ell_{\infty}$ norm. Namely
$\|\bx_0\|_{\infty}\approx \Delta\sqrt{(2\log d)/d}$: as $d$
gets larger, this is much smaller
than the typical entry of $\bx$, which is of order $1$.

\subsection{Random features model: Derivation Eq.~\eqref{eq:Claim}}
\label{sec:SketchClaim}

In this section we outline the derivation of Eq.~\eqref{eq:Claim}. Let $\bU_{\sigma} = \sigma(\bX\bW^{\sT})\in\reals^{n\times N}$
(where $\sigma$ is applied entrywise to the matrix $\bX\bW^{\sT}$). Then
\begin{align}
  \E\{\bW^{\sT}\hbtheta(\bW) |\bX,\by\} &=  \E\{\bW^{\sT}(\bU_{\sigma}^{\sT}\bU_{\sigma}+du^2\id)^{-1}\bU_{\sigma}^{\sT}\by |\bX,\by\}\\
  & =  \E_{\bW}\{\bW^{\sT}(\bU_{\sigma}^{\sT}\bU_{\sigma}+du^2\id)^{-1}\bU_{\sigma}^{\sT}\bX\bbeta_0\}\, .
\end{align}
where $u^2= \lambda\psi_1\psi_2$.
We will work within the `noisy linear features model' of \cite{mei2019generalization} which replaces $\bU_{\sigma}$
with
\begin{align}
  \bU = \mu_1\bX\bW^{\sT}+\mu_*\bZ\, ,
\end{align}
where $(Z_{ij})_{i\le n,j\le N}\sim_{iid}\normal(0,1)$,
$(X_{ij})_{i\le n,j\le d}\sim_{iid}\normal(0,1)$, $(\bW_{ij})_{i\le N,j\le d}\sim_{iid}\normal(0,1/d)$.
The universality results of \cite{mei2019generalization} suggest that
this substitution produces an error that is asymptotically negligible, namely:
\begin{align}
  \E\{\bW^{\sT}\hbtheta(\bW) |\bX,\by\} &= \tbQ(\bX)\bbeta_0 +\berr_0(d,\lambda)\, ,\label{eq:EWW}\\
  \tbQ(\bX) &:= \E_{\bW,\bZ}\big\{\bW^{\sT}(\bU^{\sT}\bU_{\sigma}+du^2\id_N)^{-1}\bU^{\sT}\bX\big\}\, .
\end{align}
Note that, conditional on $\bX$ $\bU,\bW$ are jointly Gaussian, and therefore it is easy to compute the conditional expectation
\begin{align}
  \E\{\bW^{\sT}|\bU,\bX\} = \mu_1\Big(\mu_*^2d \id_d+\mu_1^2\bX^{\sT}\bX\Big)^{-1}\bX^{\sT}\bU\, .
\end{align}
We therefore get
\begin{align}
  \tbQ(\bX) &=\mu_1\Big(\mu_*^2d \id_d+\mu_1^2\bX^{\sT}\bX\Big)^{-1}\bX^{\sT}\bQ(\bX) \bX\, ,\label{eq:qQbQ}\\
  \bQ(\bX) & := \E_{\bW,\bZ}\{\bU(\bU^{\sT}\bU+du^2\id_N)^{-1}\bU^{\sT}\}\, .
\end{align}
At this point we claim that, for $\psi_1\neq\psi_2$,
\begin{align}\label{eq:ClaimQ}
  \bQ(\bX) =  \Big(\frac{\mu^2_1}{d}\bX\bX^{\sT}+\mu_*^2(1+q)\id_d\Big)
  \Big(\frac{\mu^2_1}{d}\bX\bX^{\sT}+\mu_*^2(1+q)\id_n\Big)^{-1}+\bErr(\lambda,d)\, ,
\end{align}
where the $\bErr(\lambda,d)$ is negligible for our purposes (namely
$\|\mu_1\Big(\mu_*^2d \id_d+\mu_1^2\bX^{\sT}\bX\Big)^{-1}\bX^{\sT}\bErr\bX\bbeta_0\|_2=o_{\P}(1)$
when $\lambda\to 0$ after $d\to\infty$).
The main claim of this section ---Eq.~\eqref{eq:Claim}--- follows by substituting \eqref{eq:ClaimQ}
in Eqs.~\eqref{eq:EWW}, \eqref{eq:qQbQ} after some algebraic manipulations.

In the overparametrized regime $\psi_1>\psi_2$ (which is the main focus of our analysis) Eq.~\eqref{eq:ClaimQ}
is straightforward. Notice that in this case $q=0$, and therefore this claim amounts to
$\bQ(\bX) = \id_n+\bErr(\lambda,d)$. Indeed this is straightforward since in that case $\bU$ has
full column rank, and minimum singular value bounded away from zero. Therefore
\begin{align}
 \left\|\bQ(\bX) - \E_{\bW,\bZ}\{\bU(\bU^{\sT}\bU)^{-1}\bU^{\sT}\} \right\|_{\sop}\le Cu^2= C'\lambda\, ,
\end{align}
and  $\bU(\bU^{\sT}\bU)^{-1}\bU^{\sT}=\id_n$.

In the underparametrized regime $\psi_1<\psi_2$ the result can be obtained
making use of \cite{ledoit2011eigenvectors}.

\end{document}